\documentclass{acmtrans2m}

\usepackage{latexsym} 
\usepackage{amssymb}		
\usepackage{amsmath}		
\usepackage{graphicx}

\newtheorem{theorem}{Theorem}[section]

\newtheorem{corollary}[theorem]{Corollary}

\newtheorem{lemma}[theorem]{Lemma}
\newdef{definition}[theorem]{Definition}
\newdef{remark}[theorem]{Remark}
\newdef{example}[theorem]{Example}
\newdef{claim}[theorem]{Claim}

\DeclareMathSymbol{\FORALL}   {\mathord}{symbols}{"38}
\DeclareMathSymbol{\EXISTS}   {\mathord}{symbols}{"39}
\DeclareMathSymbol{\SUCHTHAT} {\mathbin}{symbols}{"01}
\def\Forall#1#2{{\FORALL#1}\SUCHTHAT #2}
\def\Exists#1#2{{\EXISTS#1}\SUCHTHAT #2}
\DeclareSymbolFont{AMSb}{U}{msb}{m}{n}
\DeclareMathSymbol{\N}{\mathbin}{AMSb}{"4E}
\DeclareSymbolFontAlphabet{\mathbb}{AMSb}

\newcommand{\NOTE}[1]{\ifthenelse{\boolean{draft}}{\begin{quotation}\textbf{NOTE}:#1\end{quotation}}{}}
\newcommand{\set}[1]{\ensuremath{\{#1\}}}

\newcommand{\setmin}[2]{\ensuremath{#1\!\setminus\!#2}}

\newcommand{\clos}[1]{\ensuremath{\mathit{clos}(#1)}}

\newcommand{\body}[1]{\textrm{body}(#1)}

\newenvironment{pprogram}{\[\begin{array}{rll}}{\end{array}\]}

\newcommand{\tsrule}[2]{\ensuremath{\mathit{#1} &\gets& \mathit{#2}\\}}

\newcommand{\prule}[2]{\ensuremath{\mathit{#1}\gets\mathit{#2}}}

\newenvironment{pprogramn}{\[\begin{array}{rrll}}{\end{array}\]}
\newcommand{\nrule}[3]{\ensuremath{\mathit{#1}: & \mathit{#2} & \gets & \mathit{#3} \\}}

\newcommand{\naf}[1]{not~#1}
\newcommand{\NAF}{\ensuremath{\textit{not}}}

\newcommand{\HBase}[1]{\ensuremath{\mathcal{B}_{#1}}}

\newcommand{\ground}[2]{{#1}_{#2}}

\newcommand{\true}{\textbf{true}}
\newcommand{\false}{\textbf{false}}

\newcommand{\preds}[1]{\ensuremath{\mathit{preds}(#1)}}

\newcommand{\edb}[1]{\ensuremath{\mathit{edb}(#1)}}

\newcommand{\vars}[1]{\ensuremath{\mathit{vars}(#1)}}

\newcommand{\exptime}{\textsc{exptime}}

\newcommand{\np}{\ensuremath{\textsc{np}}}
\newcommand{\p}{\ensuremath{\textsc{p}}}

\newcommand{\exptimex}[1]{{#1}-\textsc{exptime}}

\newcommand{\nexptime}{\ensuremath{\textsc{nexptime}}}

\newcommand{\xnexptime}[1]{\ensuremath{{#1}\mbox{-}\textsc{nexptime}}}

\newcommand{\exptimetwonp}[1]{\ensuremath{2\mbox{-}\textsc{exptime}^{\textsc{np}}}}
\newcommand{\exptimetwonexptime}[1]{\ensuremath{2\mbox{-}\textsc{exptime}^{\textsc{nexptime}}}}
\newcommand{\exptimetwosigma}[1]{\ensuremath{2\mbox{-}\textsc{exptime}^{\SIGMA{2}}}}
\newcommand{\exptimetwonexptimetwo}[1]{\ensuremath{2\mbox{-}\textsc{exptime}^{2\mbox{-}\textsc{nexptime}}}}

\newcommand{\card}[1]{\ensuremath{\mbox{\ensuremath{\vert #1 \vert}}}}

\newcommand{\posi}[1]{\ensuremath{{#1}^{+}}}
\newcommand{\nega}[1]{\ensuremath{{#1}^{-}}}

\newcommand{\lit}[1]{\ensuremath{\mathit{#1}}}

\newcommand{\cts}[1]{\ensuremath{\mathit{cts}{(#1)}}}

\newcommand{\Comp}[1]{\ensuremath{\mathtt{comp}(#1)}}
\newcommand{\Compg}[1]{\ensuremath{\mathtt{compgl}(#1)}}
\newcommand{\CComp}[1]{\ensuremath{\mathtt{ccomp}(#1)}}
\newcommand{\GComp}[1]{\ensuremath{\mathtt{gcomp}(#1)}}
\newcommand{\GCompg}[1]{\ensuremath{\mathtt{gcompgl}(#1)}}

\newcommand{\impl}[2]{\ensuremath{\mathit{#1} \Rightarrow \mathit{#2}}}
\newcommand{\equiva}[2]{\ensuremath{\mathit{#1} \Leftrightarrow \mathit{#2}}}

\newcommand{\pprog}[2]{\ensuremath{{#1}_{#2}}}

\newcommand{\lfp}[3]{\ensuremath{[\mathrm{LFP}~{#1}{#2}.{#3}]}}

\newcommand{\oper}[1]{\ensuremath{\psi^{#1}}}
\newcommand{\operphi}[1]{\ensuremath{\phi^{#1}}}

\newcommand{\lfpoint}[1]{\ensuremath{\mathrm{LFP}({#1})}}

\newcommand{\fix}[1]{\ensuremath{\mathtt{fix}(#1)}}

\newcommand{\sat}[1]{\ensuremath{\mathtt{sat}(#1)}}

\newcommand{\gl}[1]{\ensuremath{\mathtt{gl}(#1)}}

\newcommand{\fpf}[1]{\ensuremath{\mathtt{fpf}(#1)}}

\newcommand{\mugf}{\ensuremath{\mu\mbox{GF}}}
\newcommand{\mulgf}{\ensuremath{\mu\mbox{LGF}}}
\newcommand{\mulgfbrack}{\ensuremath{\mu\mbox{(L)GF}}}

\newcommand{\hbg}[1]{\ensuremath{{#1}^{\mathrm{f}}}}

\newcommand{\gua}[1]{\ensuremath{{#1}^{\mathrm{g}}}}

\newcommand{\dlite}{Datalog \textsc{lite}}
\newcommand{\dliter}{Datalog \textsc{liter}}

\newcommand{\rel}[1]{\ensuremath{\mathcal{#1}}}

\newcommand{\double}[1]{\ensuremath{\neg \neg {#1}}}

\newcommand{\id}[1]{\ensuremath{\mathit{id}{(#1)}}}

\newcommand{\SIGMA}[1]{\ensuremath{\Sigma^\p_{#1}}}

\newcommand{\strule}[2]{\ensuremath{\mathit{#1} & \gets \mathit{#2}}}

\newcommand{\Next}[1]{\ensuremath{\mathsf{X}#1}}

\newcommand{\eventually}[1]{\ensuremath{\mathsf{F}#1}}
\newcommand{\always}[1]{\ensuremath{\mathsf{G}#1}}
\newcommand{\until}[2]{\ensuremath{#1~\mathsf{U}~#2}}
\newcommand{\some}[1]{\ensuremath{\mathsf{E}#1}}
\newcommand{\all}[1]{\ensuremath{\mathsf{A}#1}}

\newcommand{\pr}[1]{\ensuremath{[#1]}}

\newcommand{\gfp}[3]{\ensuremath{[\mathrm{GFP}~{#1}{#2}.{#3}]}}

\newcommand{\gfpoint}[1]{\ensuremath{\mathrm{GFP}({#1})}}

\newcommand{\gli}[1]{\ensuremath{{#1}^{\mbox{\footnotesize x}}}}
\newcommand{\fx}{\mbox{\footnotesize x}}
\newcommand{\glireduct}[2]{\ensuremath{{#1}^{\mbox{\footnotesize x}(#2)}}}

\newcommand{\HBaseU}[2]{\ensuremath{\mathcal{B}_{#1}^{#2}}}
\newcommand{\glit}[1]{\ensuremath{\mathtt{gli}(#1)}}
\newcommand{\dlitem}{Datalog \textsc{litem}}

\title{Open Answer Set Programming with Guarded Programs}
            
\author{STIJN HEYMANS \\ Digital Enterprise Research Institute (DERI) \\ Leopold-Franzens-Universit\"at\\ Innsbruck, Austria
	\\ stijn.heymans@deri.org
	\and DAVY VAN NIEUWENBORGH, and DIRK VERMEIR \\ Dept. of Computer Science\\
	Vrije Universiteit Brussel, VUB\\
	Pleinlaan 2, B1050 Brussels, Belgium\\
	\{dvnieuwe,dvermeir\}@vub.ac.be}
 
\begin{abstract} 
Open answer set programming (OASP) is an extension of answer set
programming where one may ground a program with an arbitrary superset
of the program's constants.  We define a fixed point logic (FPL)
extension of Clark's completion such that open answer sets correspond
to models of FPL formulas and identify a syntactic subclass of
programs, called (loosely) guarded programs. Whereas reasoning with
general programs in OASP is undecidable, the FPL translation of
(loosely) guarded  programs falls in the decidable (loosely) guarded
fixed point logic (\mulgfbrack{}).  Moreover, we reduce normal closed
ASP to loosely guarded OASP, enabling 
for the first time, a characterization of an answer set semantics by
\mulgf{} formulas.
\par
We further extend the open answer set semantics for programs with
generalized literals. Such \emph{generalized programs (gPs)} have
interesting properties, e.g., the ability to express infinity axioms.
We restrict the syntax of gPs such that both rules and generalized
literals are \emph{guarded}. Via a translation to guarded fixed point
logic, we deduce \exptimex{2}-completeness of satisfiability
checking in such \emph{guarded gPs} (GgPs).  \emph{Bound GgPs} are
restricted GgPs with \exptime{}-complete satisfiability checking, but
still sufficiently expressive to optimally simulate \emph{computation
tree logic} (CTL).  We translate \dlite{} programs to GgPs,
establishing equivalence of GgPs under an open answer set semantics,
alternation-free $\mugf$, and \dlite.  \end{abstract}
 
\category{I.2.3}{Artificial Intelligence}{Deduction and Theorem Proving}[Logic Programming]
\category{I.2.4}{Artificial Intelligence}{Knowledge Representation Formalisms and Methods}
\terms{Theory}
\keywords{Answer Set Programming, Open Domains, Fixed Point Logic}

\begin{document}

\begin{bottomstuff} 
This is a revised and extended version of \cite{hvnv-lpnmr2005} and
\cite{hvnv-foiks2006}. \\ Stijn Heymans is supported by the European
Commission under the projects Knowledge Web and SUPER; by the FFG
(\"Osterreichische Forschungsf\"orderungsgeselleschaft mbH) under the
projects RW$^2$, SemNetMan, and SEnSE.  Davy Van Nieuwenborgh is
supported by the Flemish Fund for Scientific Research
(FWO-Vlaanderen).
\end{bottomstuff}
            
\maketitle

\section{Introduction}

In closed answer set programming (ASP) \cite{gelfond88stable}, a
program consisting of a rule $\prule{p(X)}{\naf{q(X)}}$ and a fact
$q(a)$ is grounded with the program's constant $a$, yielding
$\prule{p(a)}{\naf{q(a)}}$ and $q(a)$. This program has one answer set
$\set{q(a)}$ such that one concludes that the predicate $p$ is not
satisfiable, i.e., there is no answer set of the program that contains
a literal with predicate $p$.  Adding more constants to the program
could make $p$ satisfiable, e.g., in the absence of a deducible
$q(b)$, one has $p(b)$. However, in the context of conceptual modeling,
such as designing database schema constraints,
 this implicit dependence on constants in
the program in order to reach sensible conclusions
 is infeasible.  One wants to be able to test satisfiability
of a predicate $p$ in a schema independent of any associated data, see, e.g., conceptual modeling as 
in the \emph{Object-role Modeling} paradigm \cite{halpin}. 
\par
For answer set programming, this problem was solved in
\cite{Gelfond:93}, where $k$-belief sets are the answer
sets of a program that is extended with $k$ extra constants (reasoning
with $k$-belief sets was shown to be undecidable in \cite{schlipf93}).
We extended this idea, e.g., in
\cite{hvnv-eswc2005}, by allowing for arbitrary, thus possibly
infinite, universes\footnote{
	Note that answer sets for programs with (infinite) universes were also considered in \cite{schlipf95}.
}.  \emph{Open answer sets} are pairs $(U,M)$
with $M$ an answer set of the program grounded with $U$. The above
program has an open answer set $(\set{x,a},\set{q(a),p(x)})$ where $p$
is satisfiable.  Open Answer Set Programming solves the above conceptual
modeling problem, confirmed by the ability of Open Answer Set Programming to
simulate several expressive Description Logics \cite{hvnv-eswc2005}. Moreover,
as it is a rule-based formalism Open Answer Set Programming is thus very
suitable to function as an integrating formalism of Description Logics and
Logic Programming.

Characteristic about (O)ASP is its treatment of negation as failure
(naf): one guesses an interpretation for a program, computes the
program without naf (the GL-reduct \cite{gelfond88stable}), calculates
the iterated fixed point of this reduct, and checks whether this fixed
point equals the initial interpretation. We compile these external
manipulations, i.e., not expressible in the language of programs
itself, into fixed point logic (FPL) \cite{gradel} formulas.
First, we rewrite
an arbitrary program as a program containing only one designated
predicate $p$ and (in)equality; this makes sure that when calculating
a fixed point of the predicate variable $p$, it constitutes a fixed
point of the whole program.  In the next phase, such a
\emph{$p$-program} $P$ is translated to FPL formulas \Comp{P}.
\Comp{P} ensures satisfiability of program rules by formulas
comparable to those in Clark's completion.  The specific answer set
semantics is encoded by formulas indicating that for each atom
$p(\textbf{x})$ in the model there must be a true rule body that
motivates the atom, and this in a minimal way, i.e., using a fixed
point predicate. Negation as failure is correctly handled by making
sure that only those rules that would be present in the GL-reduct can
be used to motivate atoms.
\par
In \cite{chandra}, Horn clauses were translated to FPL formulas and in
\cite{dataloglite} reasoning with an extension of stratified Datalog
is reduced to FPL, but, to the best of our knowledge, this is the
first encoding of an answer set semantics in FPL.
\par
In \cite{lin:assat,lee:loop}, ASP with (finite) propositional programs
is reduced to propositional satisfiability checking.  The translation
makes the loops in a program explicit and ensures that atoms
$p(\textbf{x})$ are motivated by bodies outside of these loops.  Although
this is an elegant characterization of answer sets in the
propositional case, the approach does not seem to hold for OASP, where
programs are not propositional but possibly ungrounded and with
infinite universes.  Instead, we directly use the built-in ``loop
detection" mechanism of FPL, which enables us to go beyond
propositional programs.
\par
Translating OASP to FPL is thus interesting in its own right, but 
it also enables the analysis of decidability of OASP via
decidability results of fragments of FPL.  Satisfiability checking of
a predicate $p$ w.r.t. a program, i.e., checking whether there exists
an open answer set containing some $p(\textbf{x})$, is undecidable.
It is well-known that satisfiability
checking in FOL is undecidable, and thus the extension to FPL is too\cite{moschovakis}.
However, expressive decidable fragments of FPL have been
identified \cite{gradel}: \emph{(loosely) guarded fixed point logic}
(\mulgfbrack) extends the \emph{(loosely) guarded fragment}\index{guarded fragment} (L)GF of
FOL with fixed point predicates.
\par
GF is identified in \cite{andreka98} as a fragment of FOL satisfying
properties such as decidability of reasoning and the tree model
property, i.e., every model can be rewritten as a tree model.  The
restriction of quantified variables by a \emph{guard}, an atom
containing the variables in the formula, ensures decidability in GF.
Guards are responsible for the tree model property of GF (where the
concept of tree is adapted for predicates with arity larger than $2$),
which in turn enables tree-automata techniques for showing
decidability of satisfiability checking.  In \cite{benthem}, GF is
extended to LGF where guards can be conjunctions of atoms and,
roughly, every pair of variables must be together in some atom in the
guard.  Satisfiability checking in both GF and LGF is
\exptimex{2}-complete \cite{gradelrestraining}, as are their extensions
with fixed point predicates \mugf{} and \mulgf{} \cite{gradel}. 
\par
We identify a syntactically restricted class of programs,
\emph{(loosely) guarded programs ((L)GPs)}, for which the FPL
translation falls in (alternation-free\footnote{$\mulgfbrack$ without
nested fixed point variables in alternating least and greatest fixed
point formulas.}) \mulgfbrack{}, making satisfiability checking
w.r.t. (L)GPs decidable and in \exptimex{2}.  In LGPs, rules have a
set of atoms, the guard, in the positive body, such that every pair of
variables in the rule appears together in an atom in that guard.  GPs
are the restriction of LGPs where guards must consist of exactly one
atom. 
\par
Programs under the normal answer set semantics can be rewritten as
LGPs under the open answer set semantics by guarding all variables
with atoms that can only introduce constants from the original program.
Besides the desirable property that OASP with LGPs is thus a proper
decidable extension of normal ASP, this yields that satisfiability
checking w.r.t. LGPs is, at least, \nexptime{}-hard.
\par
\dlite{} \cite{dataloglite} is a language based on stratified Datalog
with input predicates where rules are monadic or guarded and may have
generalized literals in the body, i.e., literals of the form
$\Forall{\textbf{Y}}{\impl{a}{b}}$ for atoms $a$ and $b$.  It has an
appropriately adapted bottom-up fixed point semantics.  \dlite{} is 
devised to ensure linear time model checking while being expressive
enough to capture \emph{computational tree logic} \cite{emerson:1982}
and alternation-free $\mu$-calculus \cite{kozen83}.  Moreover, it is
shown to be equivalent to alternation-free \mugf{}.
Our reduction of GPs to alternation-free \mugf{} ensures that we have a reduction from
GPs to \dlite{}, and thus couples the answer set semantics to a fixed
point semantics based on stratified programs.  Intuitively, the guess
for an interpretation in the answer set semantics corresponds to the
input structure one feeds to the stratified Datalog program.  The
translation from GPs to \dlite{} needs only one stratum to
subsequently perform the minimality check of answer set programming.
\par
The other way around, we reduce satisfiability checking in
recursion-free \dlite{} to satisfiability checking w.r.t. GPs.
Recursion-free \dlite{} is equivalent to GF \cite{dataloglite}, and,
since satisfiability checking of GF formulas is
\exptimex{2}-hard \cite{gradelrestraining}, we obtain
\exptimex{2}-completeness for satisfiability checking w.r.t. (L)GPs. 
\par
We next extend programs with generalized
literals, resulting in \emph{generalized programs
(gPs)}\index{generalized program}.  A
generalized literal is a first-order formula of the form
$\Forall{\textbf{Y}}{\phi\Rightarrow \psi}$ where $\textbf{Y}$ is a sequence
of variables, $\phi$ is a finite
boolean combination of atomic formula and $\psi$ is an atom. Intuitively, such a generalized
literal is true
in an open interpretation $(U,M)$ if for all substitutions\footnote{
	As usual, for a finite boolean combination of atomic formula $\xi$, we use
	$\xi[\textbf{Y}\mid \textbf{y}]$ to denote the formula $\xi$ where all occurrences of $Y$ are replaced by $y$.
} $[\textbf{Y}\mid \textbf{y}]$, $\textbf{y}$ in $U$, such that $\phi[\textbf{Y}\mid \textbf{y}]$ is true
in $M$, $\psi[\textbf{Y}\mid \textbf{y}]$ is true in $M$. 
\par
Generalized literals $\Forall{\textbf{Y}}{\phi\Rightarrow \psi}$, with
$\phi$ an atom instead of a finite boolean combination of atomic formula,  were
introduced in Datalog\footnote{
The extension of logic programming syntax with
first-order formulas dates back to \cite{lloydexpressive}.
}
with the language
\dlite{}.
In open answer set programming (OASP), we define 
a reduct that removes the generalized literals.
E.g., a rule
\[
r:\prule{ok}{\Forall{X}{critical(X)\Rightarrow work(X)}}
\]
expresses that a system is OK if all critical devices are 
functioning:  the \emph{GeLi-reduct (generalized literal reduct)} of such a rule for an open interpretation $(\set{x_0,
\ldots},M)$ where $M$ contains $\lit{critical(x_i)}$ for even $i$,
contains a rule 
\[
r':\prule{ok}{work(x_0), work(x_2), \ldots}
\]
indicating that the system is OK if the critical devices $x_0$, $x_2,
\ldots$ are working.  The GeLi-reduct does not contain generalized
literals and one can apply the normal answer set semantics, modified
to take into account the infinite body.
\par
Just as it is not feasible to introduce all relevant constants in a program to
ensure correct conceptual reasoning, 
it is not feasible, not even
possible, to write
knowledge directly as in $r'$ for it has an infinite body. Furthermore, even in the presence of
a finite universe, generalized
literals  allow for a more robust representation of knowledge than
would be possible without them.
E.g., with
 critical devices $y_1$ and $y_2$, a rule
$s:\prule{ok}{work(y_1), work(y_2)}$ does the job as well as $r$
(and in fact $s$ is the GeLi-reduct of $r$), but adding new critical devices, implies
revisiting $s$ and replacing it by a rule that reflects the updated
situation.  Not only is this cumbersome, it may well be impossible as
$s$ contains no explicit reference to critical devices, and
the knowledge engineer may not have a clue as to
which rules to modify.
\par
One can modify the aforementioned FPL translation of programs without
generalized literals to take into account generalized literals.  With
this FPL translation, we then have again a mapping from one
undecidable framework into another undecidable framework. 
We restrict gPs, resulting in \emph{guarded gPs (GgPs)}\index{guarded
gPs}, such that all
variables in a rule appear in an atom in the positive body and all
generalized literals are guarded, where a generalized literal is
guarded if it can be written as a guarded formula in \mugf.
 The FPL translation of GgPs then
falls into the $\mugf$ fragment, yielding a \exptimex{2} upper
complexity bound for satisfiability checking.  Together with the \exptimex{2}-completeness of
guarded programs without generalized literals
this establishes \exptimex{2}-completeness for
satisfiability checking w.r.t. GgPs.  As a consequence, adding
generalized literals to a guarded program does not increase the complexity of
reasoning.
\par
We further illustrate the expressiveness of (bound) GgPs by simulating
reasoning in \emph{computational tree logic (CTL)} \cite{emerson:1990}, a temporal logic.
\textit{Temporal logics} \cite{emerson:1990} are widely used
for expressing  properties of nonterminating programs.
Transformation semantics, such as \textit{Hoare's logic} are not
appropriate here since they depend on the program having a final state that
can be verified to satisfy certain properties.  Temporal logics on the
other hand have a notion of (infinite) time and may express properties
of a program along a time line, without the need for that program to
terminate.  E.g., formulas may express that from each state a program
should be able to reach its initial state:
$\all{\always{\some{\eventually{\mathit{initial}}}}}$.
\par
Two well-known temporal logics are \emph{linear temporal logic
(LTL)} \cite{emerson:1990,sistla:1985} and \emph{computation tree logic
(CTL)} \cite{emerson:1990,emerson:1982b,clarke:1986}, which 
differ in their interpretation of time: the former assumes that time
is linear, i.e., for every state of the program there is only one
successor state, while time is branching for the latter, i.e., every
state may have different successor states, corresponding to
nondeterministic choices for the program.
\par
Since CTL satisfiability checking is \exptime-complete and
satisfiability checking w.r.t.  GgPs is \exptimex{2}-complete, a
reduction from CTL to GgPs does not seem to be optimal.  However, we
can show that the particular translation has a special form, i.e., it
is \emph{bound}, for which reasoning is \exptime{}-complete and thus
optimal.
\par
Finally, we can reduce general \dlite{} reasoning, i.e., with recursion, to reasoning with GgPs.  In particular,
we prove a generalization of the well-known result from
\cite{gelfond88stable} that the unique answer set of a stratified program
coincides with its least fixed point model: for a universe $U$, the
unique open answer set $(U,M)$ of a stratified Datalog program with
generalized literals is
identical\footnote{Modulo equality atoms, which are implicit in OASP,
but explicit in \dlite{}.} to its least fixed point model
with input structure $\id{U}$, the identity relation on $U$.
Furthermore, the \dlite{}
simulation, together with the reduction of GgPs to alternation-free  $\mugf$, as well as
the equivalence of alternation-free \mugf{} and \dlite{}
\cite{dataloglite}, 
lead to the conclusion that alternation-free $\mugf$, \dlite{}, and
OASP with GgPs, are equivalent, i.e., their satisfiability checking
problems can be effectively polynomially reduced to one another.
\par
GgPs are thus just as expressive as \dlite{}, however, from a
knowledge representation viewpoint, GgPs allow for a compact
expression of circular knowledge.  E.g., the omni-present construction
with rules $\prule{a(X)}{\naf{b(X)}}$ and $\prule{b(X)}{\naf{a(X)}}$
is not stratified and cannot be (directly) expressed in \dlite{}.  The
reduction to \dlite{} does indicate that negation as failure under the
(open) answer set semantics is not that special, but can be regarded as convenient semantic sugar.
\par
The remainder of the paper starts with an introduction of the open answer
set semantics, fixed point logic, and computation tree logic.
In Section \ref{sec:fpl}, we reduce satisfiability checking w.r.t.
arbitrary logic programs to satisfiability checking of
alternation-free fixed point logic formulas.  We identify in Section
\ref{sec:goasp} syntactical classes of programs for which this FPL
translation falls into the decidable logic $\mugf$ or $\mulgf$, i.e.,
guarded or loosely guarded fixed point logic.
\par
In Section \ref{sec:answergl}, we introduce so-called generalized
literals and modify the translation to FPL in Section
\ref{sec:fplgl}.  Section \ref{sec:goaspg} mirrors Section
\ref{sec:goasp} and identifies classes of programs with
generalized literals that can be mapped to guarded FPL.  In
Section \ref{sec:related}, we relate the obtained languages under the
open answer set semantics to \dlite{} which has a least fixed point
model semantics.  Section \ref{sec:ctloasp} discusses a translation from CTL to bound guarded programs.
Finally, Section \ref{sec:conclusions} contains
conclusions and directions for further research.

\section{Preliminaries}

\subsection{Open Answer Set Programming}

We introduce open answer set programming (OASP) as in
\cite{hvnv-jal2006}.
\emph{Constants}, \emph{variables}, \emph{terms}, and \emph{atoms} are
defined as usual\footnote{
	Note that we do not allow function symbols.
}.  A \textit{literal} is an atom $p(\textbf{t})$ or a
\textit{naf-atom} $\naf{p(\textbf{t})}$.\footnote{We have no classical negation
$\neg$, however, programs with $\neg$ can be reduced to programs
without it, see e.g.  \cite{lifschitz:2001}.} The \textit{positive
part} of a set of literals $\alpha$ is $\posi{\alpha} =
\set{p(\textbf{t}) \mid p(\textbf{t}) \in \alpha}$ and the \textit{negative
part} of $\alpha$ is $\nega{\alpha} = \set{p(\textbf{t}) \mid
\naf{p(\textbf{t})} \in \alpha}$.  We assume the existence of binary
predicates $=$ and $\neq$, where $t = s$ is considered as an atom and
$t\neq s$ as $\naf{t =s}$.  E.g., for $\alpha = \set{X\neq Y, Y = Z}$,
we have $\posi{\alpha} = \set{Y = Z}$ and $\nega{\alpha} = \set{X =
Y}$.  A \emph{regular} atom is an atom that is not an equality atom.
For a set $X$ of atoms, $\naf{X} = \set{\naf{l} \mid l \in X}$.
\par
A \textit{program} is a countable set of rules \prule{\alpha}{\beta},
where $\alpha$ and $\beta$ are finite sets of literals,
$\card{\posi{\alpha}} \leq 1$, and $\Forall{t,s}{t = s \not\in
{\posi{\alpha}}}$, i.e., $\alpha$ contains at most one positive atom,
and this atom cannot be an equality atom.\footnote{The condition
$\card{\posi{\alpha}} \leq 1$ ensures that the GL-reduct is
non-disjunctive.} The set $\alpha$ is the \textit{head} of the rule
and represents a disjunction of literals, while $\beta$ is called the
\textit{body} and represents a conjunction of literals.  If $\alpha =
\emptyset$, the rule is called a \textit{constraint}.  \emph{Free
rules} are rules of the form \prule{q(\textbf{t})\lor\naf{q(\textbf{t})}}{}
for a tuple $\textbf{t}$ of terms; they enable a choice for the inclusion
of atoms.  We call a predicate $p$ free if there is a free rule 
\prule{p(\textbf{t})\lor\naf{p(\textbf{t})}}{}.
  Atoms, literals, rules, and programs that do not contain
variables are \textit{ground}.  
\par
For a program $P$, let $\cts{P}$ be the constants in $P$, $\vars{P}$
its variables, and \preds{P} its predicates.  A \emph{universe} $U$
for $P$ is a non-empty countable\footnote{Note that $U$ is countable, as later on, this is needed to be able to use a result from \cite{gradelrestraining}
that indicates that the fixed point can be reached at the first ordinal $\omega$.}  superset of the constants in $P$:
$\cts{P} \subseteq U$.  We call $\ground{P}{U}$ the ground program
obtained from $P$ by substituting every variable in $P$ by every
possible constant in $U$.  Let $\HBaseU{P}{U}$ be the set of ground regular atoms
that can be formed from a ground program $P$ and the elements in $U$. 
\par
Let $I$ be a subset of some $\HBaseU{P}{U}$.  For a ground regular atom $p(\textbf{t})$, we write $I
\models p(\textbf{t})$ if $p(\textbf{t}) \in I$; For an equality atom
$p(\textbf{t}) \equiv t = s$, we have $I \models p(\textbf{t})$ if $s$ and
$t$ are equal constants.  We have $I \models \naf{p(\textbf{t})}$ if $I
\not\models p(\textbf{t})$.  For a set of ground literals $X$, $I\models
X$ if $I \models l$ for every $l \in X$.  A ground rule $r:
\prule{\alpha}{\beta}$ is \textit{satisfied} w.r.t. $I$, denoted $I
\models r$, if $I \models l$ for some $l \in \alpha$ whenever $I
\models \beta$.  A ground constraint \prule{}{\beta} is satisfied
w.r.t. $I$ if $I \not\models \beta$.
For a ground program $P$ without \NAF, $I$ is
a \textit{model} of $P$ if $I$ satisfies every rule in $P$; it is an
\textit{answer set} of $P$ if it is a subset minimal model of $P$.
For ground programs $P$ containing \NAF, the
\textit{GL-reduct} \cite{gelfond88stable} w.r.t. $I$ is defined as
$P^I$, where $P^I$ contains \prule{\posi{\alpha}}{\posi{\beta}} for
\prule{\alpha}{\beta} in $P$, $I \models \naf{\nega{\beta}}$ and $I
\models \nega{\alpha}$.  $I$ is an \textit{answer set} of a ground $P$
if $I$ is an answer set of $P^I$.
\par
In the following, a program is assumed to be a finite set of rules;
infinite programs only appear as byproducts of grounding a finite
program with an infinite universe.  An \textit{open interpretation} of
a program  $P$ is a pair $(U, M)$ where $U$ is a universe for $P$ and
$M$ is a subset of $\HBaseU{P}{U}$.  An \textit{open answer set} of
$P$ is an open interpretation $(U,M)$ of $P$ with $M$ an answer set of
$\ground{P}{U}$.  An $n$-ary predicate $p$ in $P$ is \emph{satisfiable}
if there is an open answer set $(U,M)$ of $P$ and a $\textbf{x} \in U^n$
such that $p(\textbf{x}) \in M$.  We assume
 that
when satisfiability checking a predicate $p$, $p$ is always non-free, i.e., there
are no free rules with $p$ in the head.  Note that satisfiability
checking of a free $n$-ary predicate $p$ w.r.t. $P$ can always be
reduced to satisfiability checking of a new non-free $n$-ary predicate
$p'$ w.r.t. $P\cup \set{\prule{p'(\textbf{X})}{p(\textbf{X})}}$. Note that
this is a linear reduction.
\par
\begin{example}\label{ex:restore}
Take the program
\begin{pprogramn}
\nrule{r_1}{restore(X)}{crash(X),y(X,Y),backSucc(Y)}
\nrule{r_2}{backSucc(X)}{\neg crash(X),y(X,Y),\naf{backFail(Y)}}
\nrule{r_3}{backFail(X)}{\naf{backSucc(X)}}
\nrule{r_4}{}{{y}(Y_1,X),{y}(Y_2,X),Y_1 \neq Y_2}
\nrule{r_5}{y(X,Y) \lor \naf{y(X,Y)}}{}
\nrule{r_6}{crash(X) \lor \naf{crash(X)}}{}
\nrule{r_7}{\neg crash(X) \lor \naf{\neg crash(X)}}{}
\end{pprogramn}
Rule $r_1$ 
represents the knowledge that a system that has crashed on a
particular day $X$ (\lit{crash(X)}), can be restored on that day (\lit{restore(X)}) if a backup of the system
on the day $Y$ before (\lit{y(X,Y)} -- \lit{y} stands for
\emph{yesterday}) succeeded (\lit{backSucc(Y)}).
Backups succeed, if the system does not
crash and it cannot be established that the backups at previous dates
failed ($r_2$) and a backup fails if it does not succeed ($r_3$).  Rule $r_4$
 ensures that for a particular today there can be
only one tomorrow.
Rules $r_5$, $r_6$, and $r_7$ allow to freely introduce \lit{y},
\lit{crash}, and \lit{\neg crash} literals.  Indeed, take, e.g.,
$\lit{crash(x)}$ in an interpretation; the GL-reduct w.r.t. that
interpretation contains then the rule $\prule{crash(x)}{}$ which
motivates the presence of the \lit{crash} literal in an (open) answer
set.  If there is no \lit{crash(x)} in an interpretation then the
GL-reduct removes the rule $r_5$ (more correctly, its grounded version with $x$).
Below, we formally define rules of such a form as \emph{free
rules} in correspondence with the intuition that they allow for a free
introduction of literals.
\par
Every open answer set $(U, M)$ of this
program that makes \lit{restore} satisfiable, i.e., such that there is a
$\lit{restore(x) \in M}$ for some $x \in U$, must be infinite.  An
example of such an open answer set $M$ is (we omit $U$ if it is clear from $M$)
\begin{multline*}
\set{restore(x),crash(x),backFail(x),y(x,x_1),\\
backSucc(x_1),\lnot crash(x_1),y(x_1,x_2)\\
backSucc(x_2),\lnot crash(x_2),y(x_2,x_3),\ldots}
\end{multline*} 
One sees that every \lit{backSucc} literal with element $x_i$ enforces
a new \lit{y}-successor $x_{i+1}$ since none of the previously
introduced universe elements can be used without violating rule $r_4$,
thus enforcing an infinite open answer set.
\par
Indeed, assume $\lit{restore}$ is satisfiable w.r.t. $P$.  Then, there
must be a $x_0$ in the universe $U$ of some open answer set $(U,M)$ such
that $\lit{restore(x_0)}\in M$.  With $r_1$, we must have that
$\lit{crash(x_0)}\in M$, and there must be some $x_1\in U$ such that
$\lit{y(x_0,x_1)}\in M$ and $\lit{backSucc(x_1)}\in M$, and thus, with
rule $r_2$, $\lit{\neg crash(x_1)}\in M$, $\lit{y(x_1,x_2)}\in M$ and
$\lit{backFail(x_2)}\not\in M$. With $\lit{crash(x_0)}\in M$ and $\neg
\lit{crash(x_1)}\in M$, we are sure that $x_1\neq x_0$. With $r_3$,
one must have that $\lit{backSucc(x_2)}\in M$ such that $x_2\neq x_0$
for the same reason.   Furthermore, $x_2\neq x_1$, since otherwise
$\lit{y(x_0,x_1)}\in M$ and $\lit{y(x_1,x_1)}\in M$: with $x_0\neq
x_1$ this is a contradiction with $r_4$.  Thus, summarizing, $x_2\neq
x_1$ and $x_2\neq x_0$.  One can continue this way, and one will be
obliged to introduce new $x_i$'s ad infinitum.
\end{example}
\par
Rules
$\prule{\alpha}{\beta}$ are such that 
$\card{\posi{\alpha}}\leq 1$.
This restriction ensures that the GL-reduct contains no disjunction
in the head anymore, i.e., the head will be an atom or it will be
empty.  This property of the GL-reduct allows us to define an
\emph{immediate consequence operator} \cite{emden76} $T$ that
computes the closure of a set of literals w.r.t. a GL-reduct.
\par
For a program $P$ and an open interpretation $(U,M)$ of $P$,
$T_P^{(U,M)}: \HBase{P_U}\to \HBase{P_U}$ is defined as $T(B) = B
\cup \set{a \mid a \gets \beta \in P_U^M \land B\models \beta }$.
Additionally, we define $T^0(B) = B$, and $T^{n+1}(B) = T(T^{n}(B))$.\footnote{ We omit the sub- and
superscripts $(U,M)$ and $P$ from $T_P^{(U,M)}$ if they are clear from
the context and, furthermore, we will usually write $T$ instead of
$T(\emptyset)$.}
\par
Although we allow for infinite universes, we can motivate the presence
of atoms in open answer sets in a finite way, where the motivation
of an atom is formally expressed by the immediate consequence
operator.
\begin{theorem}\label{th:finitelyminimal} 
Let $P$ be a program and
$(U,M)$ an open answer set of $P$. Then, $\Forall{a \in M}{\Exists{n
< \infty}{a \in T^{n}}}$.  
\end{theorem}

For the relation of OASP with other logic programming paradigms that
allow for (some form of) opennes, we refer to \cite{hvnv-jal2006}.

\subsection{Fixed Point Logic}

Extensions of \emph{first-order logic (FOL)}\index{first-order
logic}\index{FOL} that allow for the expression of recursive
procedures are well-investigated in \emph{finite model
theory}\index{finite model theory}, see e.g., \cite{moschovakis,Immerman86}. Also
in the presence of infinite models, so-called \emph{fixed point logic
(FPL)}\index{fixed point logic}\index{FPL} proves to be an
interesting logic \cite{flum}.  E.g., a decidable subclass of FPL is
the \emph{guarded fixed point logic} \cite{gradel}, which lifts
propositional $\mu$-calculus \cite{kozen83} to a first-order setting.
\par
We assume
FOL interpretations are represented as pairs
 $(U,M)$ where $M$ is an interpretation
 over the domain $U$.  Furthermore, we consider FOL
with equality such that equality is always interpreted as the identity
relation over $U$.
\par
We define \emph{fixed point logic (FPL)}  along the lines of
\cite{gradel}, i.e., as an extension of first-order logic, where
formulas may additionally be \emph{fixed point formulas} of the form \index{$\lfp{W}{\textbf{X}}{\psi(W,\textbf{X})}(\textbf{X})$}
\index{$\gfp{W}{\textbf{X}}{\psi(W,\textbf{X})}(\textbf{X})$}
\begin{equation}\label{eq:fix}
\lfp{W}{\textbf{X}}{\psi(W,\textbf{X})}(\textbf{X}) \quad \mbox{ or }\quad \gfp{W}{\textbf{X}}{\psi(W,\textbf{X})}(\textbf{X}) \; ,
\end{equation}
where $W$ is an $n$-ary predicate variable, $\textbf{X}$ is an $n$-ary
sequence of distinct variables, $\psi(W,\textbf{X})$ is a (FPL) formula with all
free variables contained in $\textbf{X}$ and $W$ appears only positively
in $\psi(W,\textbf{X})$.\footnote{
	A formula $\psi$ is in \emph{negation-normal
	form}\index{negation-normal form} if the only
	used connectives are $\land$, $\lor$, and $\neg$, and $\neg$ only
	appears in front of atoms.  Let $\psi$ be a formula in
	negation-normal form. A predicate $p$ appears then only positively
	in $\psi$ if there is no $\neg p$ in $\psi$.
	}
\par
For an interpretation $(U,M)$ and a valuation $\chi$ of the free
predicate variables, except $W$, in $\psi$, we define the operator 
$\oper{(U,M),\chi}:2^{U^n} \to 2^{U^n}$\index{$\oper{(U,M)}$}
on sets $S$ of $n$-ary tuples
\begin{equation}\label{eq:oper}
\oper{(U,M),\chi}(S) \equiv \set{\textbf{x} \in U^n \mid
(U,M),\chi\cup\set{W\to S} \models
\psi(W,\textbf{x})}\; ,
\end{equation}
where $\chi\cup\set{W\to S}$ is the valuation $\chi$ extended such
that the extension of $W$ is assigned to $S$.  If $\psi(W,\textbf{X})$ contains only the
predicate variable $W$, we often omit the valuation $\chi$ and write
just $\psi^{(U,M)}$.
By definition, $W$ appears only positively in $\psi$ such that
$\oper{(U,M),\chi}$ is monotonic on sets of $n$-ary $U$-tuples and thus has
a least and greatest fixed point \cite{tarski55}, which we denote by
\lfpoint{\oper{(U,M),\chi}} and \gfpoint{\oper{(U,M),\chi}}
\index{\lfpoint{\oper{(U,M),\chi}}}\index{\gfpoint{\oper{(U,M),\chi}}}
respectively.  Finally, we have that 
\begin{equation}\label{eq:fixsem}
(U,M),\chi \models \lfp{W}{\textbf{X}}{\psi(W,\textbf{X})}(\textbf{x}) \iff \textbf{x} \in \lfpoint{\oper{(U,M),\chi}} \; ,
\end{equation}
and similarly for greatest fixed point formulas.
We call an FPL \emph{sentence}\index{sentence} (i.e., an FPL
formula without free variables)
\emph{alternation-free}\index{alternation-free} if it does not
contain subformulas $\psi \equiv \lfp{T}{\textbf{X}}{\varphi}(\textbf{X})$
and $\theta \equiv \gfp{S}{\textbf{Y}}{\eta}(\textbf{Y})$ such that
$T$ occurs in $\eta$ and $\theta$ is a subformula of $\varphi$, 
or $S$ occurs in $\varphi$ and $\psi$ is a subformula of $\eta$. 
We can eliminate greatest fixed point formulas from a formula, by the
equivalence:
\begin{equation}\label{eq:gfplfp}
\gfp{W}{\textbf{X}}{\psi} \equiv \neg \lfp{W}{\textbf{X}}{\neg \psi[W/\neg W]}\;,
\end{equation}
where $\neg \psi[W/
\neg W]$ is $\neg \psi$ with $W$ replaced by $\neg W$.  If we thus remove
greatest fixed point predicates, and if negations appear
only in front of atoms or least fixed point formulas, then
a formula is alternation-free iff no fixed point variable $W$ appears
in the scope of a negation.
\par
As in \cite{Gr02}, we define
\[
\begin{split}
\oper{(U,M)}\uparrow 0 &\equiv \emptyset \\
\oper{(U,M)}\uparrow \alpha+1 &\equiv \oper{(U,M)}(\oper{(U,M)}\uparrow \alpha) \text{ for ordinals $\alpha$} \\
\oper{(U,M)}\uparrow \beta &\equiv \bigcup_{\alpha< \beta}(\oper{(U,M)}\uparrow \alpha) \text{ for limit ordinals $\beta$}
\end{split}
\]
Furthermore, since $\oper{(U,M)}$ is monotone, we have that
$\oper{(U,M)}\uparrow 0 \subseteq \oper{(U,M)}\uparrow
1\subseteq \ldots$ and there exists a (limit) ordinal $\alpha$ such that
$\oper{(U,M)}\uparrow \alpha = \lfpoint{\oper{(U,M)}}$.

\begin{example}\label{ex:infinite}
Take the conjunction of the following formulas, i.e., the  
infinity axiom\footnote{An \emph{infinity axiom}\index{infinity axiom}
is a formula that has only infinite models (if it has models).} from \cite{gradel}:
\begin{gather}
\Exists{X,Y}{F(X,Y)}\label{for:1} \\
\Forall{X,Y}{\left(F(X,Y)\Rightarrow (\Exists{Z}{F(Y,Z)})\right)}\label{for:2} \\
\Forall{X,Y}{F(X,Y)\Rightarrow \lfp{W}{X}{\Forall{Y}{F(Y,X)\Rightarrow W(Y)}}(X)}\label{for:3}
\end{gather}
A model of these formulas contains at least one $F(x,y)$ (by formula
(\ref{for:1})), which then leads to a $F$-chain by formula (\ref{for:2}).
Formula (\ref{for:3}) ensures that each element $x$ is on a well-founded
chain (and thus formula (\ref{for:2}) actually generates an infinite
chain).
\end{example}


\subsection{Computation Tree Logic}

We introduce in this subsection the temporal logic \emph{computation
tree logic (CTL)} 
\cite{emerson:1990,emerson:1982b,clarke:1986}.
Let AP be the finite set of available
proposition symbols.
CTL formulas are defined as follows:
\begin{itemize}
\item every proposition symbol $P \in AP$ is a formula,
\item if $p$ and $q$ are formulas, so are $p\land q$ and $\neg p$,
\item if $p$ and $q$ are formulas, then \some{\Next{p}},
\some{(\until{p}{q})}, \all{\Next{p}},  and
\all{(\until{p}{q})} are formulas.
\end{itemize}
The semantics of a CTL formula is given by \textit{(temporal)
structures}\index{temporal structure}.  A  structure $K$ is a tuple $(S,R,L)$ with $S$ a
countable set of states, $R\subseteq S \times S$ a total relation in
$S$, i.e., $\Forall{s \in S}{\Exists{t \in S}{(s,t) \in R}}$, and $L: S
\to 2^{AP}$ a function labeling states with propositions.
Intuitively, $S$ is a set of states, $R$ indicates the permitted
transitions between states, and $L$ indicates which propositions are
true at certain states.
\par
A path\index{path} $\pi$ in $K$ is an infinite sequence of states $(s_0,s_1,
\ldots)$ such that $(s_{i-1}, s_i) \in R$ for each $i > 0$. For a path
$\pi = (s_0, s_1, \ldots)$, we denote the element $s_i$ with $\pi_i$.
For a  structure $K = (S,R,L)$, a state $s \in S$, and a 
formula $p$, we inductively define when $K$ is a \textit{model}\index{model} of $p$ at $s$,
denoted $K, s \models p$:
\begin{itemize}
\item $K,s\models P$ iff $P \in L(s)$ for $P \in AP$,
\item $K,s\models \neg p$ iff not $K,s \models p$,
\item $K,s\models p\land q$ iff $K,s \models p$ and $K,s \models q$,
\item $K,s\models \some{\Next{p}}$ iff there is a $(s,t) \in R$ and $K, t \models p$,
\item $K,s\models \all{\Next{p}}$ iff for all $(s,t) \in R$, $K, t \models p$,
\item $K, s \models \some{(\until{p}{q})}$ iff there exists a path $\pi$ in $K$ with
$\pi_0 = s$ and $\Exists{k\geq 0}{(K, \pi_k \models q \land \Forall{j < k}{K, \pi_j\models p})}$,
\item $K, s \models \all{(\until{p}{q})}$ iff for all paths $\pi$ in $K$ with
$\pi_0 = s$ we have $\Exists{k\geq 0}{(K, \pi_k \models q \land \Forall{j < k}{K, \pi_j\models p})}$.
\end{itemize}
Intuitively, $K,s\models \some{\Next{p}}$ ($K,s\models \all{\Next{p}}$) can be read
as ``there is some neXt state where $p$ holds" (``$p$ holds in all
next states"), and $K,s\models \some{(\until{p}{q})}$ ($K,s\models \all{(\until{p}{q})}$) 
as ``there is some path from $s$ along which $p$ holds Until $q$ holds (and
$q$ eventually holds)" (``for all paths from $s$, $p$ holds until $q$
holds (and $q$ eventually holds)").
\par
Some common abbreviations for CTL formulas are $\some{\eventually{p}} =
\some{(\until{true}{p})}$ (there is some path on which $p$ will
eventually hold), $\all{\eventually{p}} =
\all{(\until{true}{p})}$ ($p$ will eventually hold on all paths), $\some{\always{p}} =
\neg\all{\eventually{\neg p}}$ (there is some path on which $p$ holds
globally), and $\all{\always{p}} =
\neg \some{\eventually{\neg p}}$ ($p$ holds everywhere on all paths). 
Furthermore, we have the standard propositional abbreviations $p \lor q = \neg(\neg p \land
\neg q)$, $p \Rightarrow q = \neg p \lor q$, and
$p \Leftrightarrow q = (p \Rightarrow q) \land (q \Rightarrow p)$.
\par
A  structure $K = (S, R, L)$ \textit{satisfies} a CTL formula $p$ if there is
a state $s \in S$ such that $K, s \models p$; we also call $K$ a
\textit{model} of $p$.  A CTL formula $p$ is
\textit{satisfiable} iff there is a model of $p$.
\par
\begin{example}\label{ex:absenceofstarv}
Consider the expression of \textit{absence of
starvation}\index{starvation} $t
\Rightarrow \all{\eventually{c}}$ \cite{clarke:1986} for a process in a
mutual exclusion\index{mutual exclusion} problem\footnote{
In the mutual exclusion problem, we have two or more processes that want to
access a critical section of code, but cannot do this at the same
time.  The problem is then how to model the behavior of the processes
(or the concurrent program in general), such that this \emph{mutual
exclusion} is never violated. For more details, we refer to, e.g., 
\cite{emerson:1982,emerson:1990,clarke:1986,attie:2001,huth,manna:1984}.  
}
.  The formula demands that if a process
tries ($t$) to enter a critical region, it will eventually succeed in
doing so ($c$) for all possible future execution paths.
\par
We will usually represent structures by diagrams as in
Figure~\ref{fig:ex1}, where states are nodes, transitions between
nodes define $R$, and the labels of the nodes contain the propositions
true at the corresponding states.
E.g., take the structure $K=(S,R,L)$ with 
\begin{itemize}
\item $S=\set{s_0,s_1,s_2}$,
\item $R = \set{(s_0,s_0),(s_0,s_1),(s_1,s_2),(s_2,s_0)}$, and
\item $L(s_0) = L(s_1) =
t$, $L(s_2) = c$,
\end{itemize}
which
is represented by
Figure~\ref{fig:ex1}. This structure does not satisfy $t \Rightarrow
\all{\eventually{c}}$ at $s_0$ since on the path $(s_0, s_0,
\ldots)$ the proposition $c$ never holds.  We have, however, $K, s_1 \models t
\Rightarrow \all{\eventually{c}}$: $t$ holds at $s_1$ such that we
must have that on all paths from $s_1$ the proposition $c$ must eventually hold; since the only path
from $s_1$ leads to $s_2$ where $c$ holds, $t \Rightarrow
\all{\eventually{c}}$ holds at $s_1$. We also have $K, s_2 \models t
\Rightarrow \all{\eventually{c}}$, 
since $t\not\in L(s_2)$.
\begin{figure}
\centerline{\includegraphics{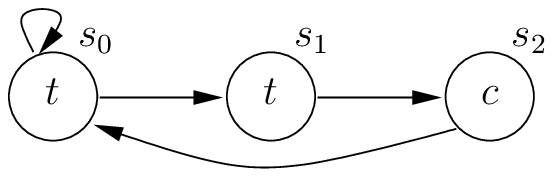}}
\caption{Example Structure $t
\Rightarrow \all{\eventually{c}}$\label{fig:ex1}}
\end{figure}
\end{example}
\par
\begin{theorem}[\cite{emerson:1990}]\label{th:ctl}
The problem of testing satisfiability for CTL is complete for
deterministic exponential time.
\end{theorem}

\section{Open Answer Set Programming via Fixed Point Logic}\label{sec:fpl}

In this section, we will show how the external manipulations to define the (open) answer set semantics
can be compiled into fixed point logic, so allowing in the next sections to analyse the decidability of 
OASP via the various decidability results of fragments of FPL.

We assume, without loss of generality, that the predicates in a
program $P$ are differently named than the constants in $P$ and that
each predicate $q$ in $P$ has one associated arity, e.g., $q(x)$ and $q(x,y)$
are not allowed.

\begin{definition}\label{def:p-program}
A program $P$ is a \emph{$p$-program} if the only predicate in $P$
different from the (in)equality predicate is $p$.
\end{definition}
For a program $P$, let $\lit{in(Y)} \equiv \set{Y\neq a
\mid a \in \preds{P} \cup \set{0}}$, i.e., a set of inequalities
between the variable $Y$ and the predicates  in $P$ as well as a new
constant $0$.  For a sequence of variables $\textbf{Y}$, we have
$\lit{in}(\textbf{Y})\equiv \cup_{Y\in \textbf{Y}}\lit{in(Y)}$.
\par
For a predicate name $p$ not appearing in an arbitrary program $P$, we can rewrite $P$ as an equivalent $p$-program
\pprog{P}{p} by replacing every regular $m$-ary atom $q(\textbf{t})$ in $P$ by
$p(\textbf{t},\textbf{0},q)$ where $p$ has arity $n$, with $n$ the maximum of
the arities of predicates in $P$ augmented by $1$, $\textbf{0}$ is a
sequence of new constants $0$ of length $n-m-1$, and $q$ is a new
constant with the same name as the original predicate.   Furthermore,
in order to avoid grounding with the new constants,
we add for every variable $X$ in a non-free rule $r\in P$ and for every newly
added constant $a$ in \pprog{P}{p}, $X \neq a$ to the body.  The rule
in \pprog{P}{p} corresponding to $r:\prule{\alpha}{\beta} \in P$ is
denoted as $\pprog{r}{p}:\prule{\pprog{\alpha}{p}}{\pprog{\beta}{p},
\lit{in(\textbf{X})}}
\in \pprog{P}{p}$ for $\vars{r} = \lit{\textbf{X}}$. 
\begin{example}
Take a program $P$:
\begin{pprogram}
\tsrule{h(a,b)}{q(X)}
\tsrule{q(X) \lor \naf{q(X)}}{}
\tsrule{}{q(a)}
\tsrule{}{q(b)}
\end{pprogram}
For a universe $U = \set{x,a,b}$ of $P$, we have the open answer sets
$M_1 = (U,\emptyset)$ and $M_2 = (U,\set{q(x),h(a,b)})$.  The
translation
\pprog{P}{p} is 
\begin{pprogram}
\tsrule{p(a,b,h)}{p(X,0,q),X \neq 0, X \neq h, X \neq q}
\tsrule{p(X,0,q) \lor \naf{p(X,0,q)}}{}
\tsrule{}{p(a,0,q)}
\tsrule{}{p(b,0,q)}
\end{pprogram}
The open answer sets of this program can then be rewritten as open
answer sets of the original program (by leaving out all ``wrong"
literals $p(q,0,q), p(0,0,q), p(h,0,q)$ that can be generated by the
free rule).
\end{example}
\begin{theorem}\label{prop:p-program}
Let $P$ be a program, $p$ a predicate not in $P$, and $q$ a predicate in $P$. $q$ is satisfiable w.r.t.
$P$ iff there is an open answer set $(U',M')$ of the $p$-program $P_p$ with
$p(\textbf{x},\textbf{0},q)\in M'$.
\end{theorem}
\begin{proof}
\newcommand{\grUa}{\ensuremath{(\pprog{P}{p})_{U'}^{M'}}}
For the ``only if" direction, assume $(U,M)$ is
  an open answer set of $P$ that satisfies $q$, i.e., there is
a $q(\textbf{x})\in M$.  
Let $U' = U\cup \preds{P}\cup \set{0}$ and $M' = \set{p(\textbf{x},\textbf{0},q) \mid
q(\textbf{x}) \in M})$.  Then $(U',M')$ is an open interpretation of
$P_p$ and $p(\textbf{x},\textbf{0},q)\in M'$.  One can show that
$(U',M')$ is an open answer set of $P_p$.
\par
For the ``if" direction, assume $(U',M')$ is an
open answer set of $\pprog{P}{p}$ with $p(\textbf{x},\textbf{0},q)\in M'$.
Define $U \equiv \setmin{U'\;}{(\preds{P}\cup \set{0})}$ and $M \equiv
\set{q(\textbf{x})\mid p(\textbf{x},\textbf{0},q)\in M' \land \textbf{x}\cap
(\preds{P}\cup \set{0}) = \emptyset}$.  
\par
We can assume
that $q$ is a non-free predicate (and we assume this throughout the
rest of this paper).  Then there are no free rules with a $q(\textbf{t})$ in the
head such that 
there are no free rules with a $p(\textbf{t},\textbf{0},q)$ in the
head in $P_p$.  Since there is a $ p(\textbf{x},\textbf{0},q)\in M'$, and
$(U',M')$ is an open answer set, there must be a rule $r[]$ in
$\grUa$\footnote{
For objects $o$ (rules, (sets of) literals, \ldots), we denote with $o[Y_1\vert y_1, \ldots, Y_d\vert y_d]$,
the grounding of $o$ where each variable $Y_i$ is substituted with
$y_i$.  Equivalently, we may write $o[\textbf{Y}\vert\textbf{y}]$ for
$\textbf{Y} = Y_1, \ldots, Y_d$ and $\textbf{y}
= y_1, \ldots, y_d$, or $o[]$ if the grounding substitution is clear
from the context, or if it does not matter what the substitution
exactly looks like.
}
 such that $M'\models \lit{in(\textbf{Y})[]}$ for $\textbf{Y}$ the
variables in the corresponding ungrounded rule $r$. Thus $\textbf{x}
\cap (\preds{P}\cup \set{0}) = \emptyset$, such that $q(\textbf{x})\in
M$, by definition of $M$.  
\par
One can show that $(U,M)$ is an
open answer set of $P$.
\qed
\end{proof}
The translation of a program to a $p$-program does not influence
the complexity of reasoning.
\begin{theorem}\label{prop:compl}
Let $P$ be a program and $p$ a predicate not in $P$.  The size\footnote{
	In the rest of the paper we use $n\times s$ for the size of a program $P$,
	where $n$ is the number of rules in $P$ and $s$ is the maximum size of the rules in $P$.
} of \pprog{P}{p} is 
polynomial in the size of $P$.
\end{theorem}
\begin{proof}
The size of a rule $r\in P$ is of the order $v+k$, with $v$ the number of
variables and $k$ the number of predicate names in $r$.  The
corresponding $r_p$ then contains an extra $v\times n$ inequality
atoms for $n\equiv \card{\preds{P}\cup \set{0}}$, and the size of
$r_p$ is thus in general quadratic in the size of $r$.
\qed
\end{proof}

By Theorems \ref{prop:p-program} and \ref{prop:compl}, we can focus, without loss of
generality, on
$p$-programs only.  Since $p$-programs have open answer sets consisting of one
predicate $p$,
fixed points calculated w.r.t. $p$ yield minimal models of the program as
we will show in Theorem \ref{th:aset-iff-model}.
\par
In \cite{chandra}, a similar motivation drives the reduction of Horn
clauses\footnote{
Horn clauses are rules of the form $a\gets \beta$ where $\beta$ is a
finite set of atoms (i.e., negation as failure is not allowed).
}
to clauses consisting of only one defined predicate.  Their
encoding does not introduce new constants to identify
old predicates and depends entirely on the use of (in)equality.
However, to account for databases consisting of only one element,
\cite{chandra} needs an additional transformation that unfolds bodies of
clauses.
\par
We can reduce a $p$-program $P$ to equivalent formulas \Comp{P} in fixed point
logic.
The \emph{completion} \Comp{P} of a program $P$ consists of 
formulas 
that demand that different constants in $P$ are interpreted as
different elements:
\begin{equation}\label{eq:constants}
a \neq b 
\end{equation}
for every pair of different constants $a$ and $b$ in $P$, and where $a \neq
b \equiv \neg( a = b)$.  
\Comp{P} contains formulas ensuring the existence of at least
one element in the domain of an interpretation:
\begin{equation}\label{eq:oneelem}
\Exists{X}{\true}\; .
\end{equation}
Besides these technical requirements matching FOL interpretations with
open interpretations, \Comp{P} contains the formulas in $\fix{P}
\equiv
\sat{P} \cup \gl{P} \cup \fpf{P}$, which can be intuitively
categorized as follows:
\begin{itemize}
\item \sat{P}
ensures that a model of \fix{P} satisfies all rules in $P$, 
\item \gl{P} is an auxiliary component defining atoms that indicate when a rule in $P$
belongs to the GL-reduct of $P$, and 
\item $\fpf{P}$ ensures that
every model of \fix{P} is a minimal model of the GL-reduct in $P$; it
uses the atoms defined in \gl{P} to select, for the calculation of the
fixed point, only those rules in $P$ that are in the GL-reduct of $P$.
\end{itemize}
We interpret a naf-atom $\naf{a}$ in a FOL formula as the literal
$\neg a$.  Moreover, we assume that, if a set $X$ is empty, $\bigwedge
X = \true$ and $\bigvee X = \false$.  In the following, we assume that the arity of $p$,
the only predicate in a $p$-program is $n$.
\begin{definition}\label{def:comp}
Let $P$ be a $p$-program.  The \emph{fixed point
translation} of $P$ is
$\fix{P} \equiv \sat{P} \cup \gl{P} \cup \fpf{P}$, where
\begin{enumerate}

\item $\sat{P}$ contains formulas
\begin{equation}\label{eq:sat}
\Forall{\textbf{Y}}{\bigwedge \beta \Rightarrow \bigvee \alpha}
\end{equation}
for rules $\prule{\alpha}{\beta} \in P$ with variables $\textbf{Y}$, 

\item $\gl{P}$ contains the formulas
\begin{equation}\label{eq:gl}
\Forall{\textbf{Y}}{\equiva{ r(\textbf{Y}) }{ \bigwedge \nega{\alpha} \land \bigwedge \neg \nega{\beta} }}
\end{equation}
for rules
$r: \prule{\alpha}{\beta} \in P$\footnote{We assume that rules are
uniquely named.} with variables $\textbf{Y}$, 

\item $\fpf{P}$ contains the formula
\begin{equation}\label{eq:fpf}
\Forall{\textbf{X}}{\impl{p(\textbf{X})}{\lfp{W}{\textbf{X}}{\phi(W,\textbf{X})}(\textbf{X})}}
\end{equation}
with
\begin{equation}\label{eq:fpformula}
\phi(W,\textbf{X}) \equiv W(\textbf{X}) \lor \bigvee_{r:\prule{p(\textbf{t}) \lor \alpha}{\beta} \in  P}E(r) 
\end{equation}
and
\begin{equation}\label{eq:Er}
E(r) \equiv \Exists{\textbf{Y}}{X_1 = t_1 \land \ldots \land X_n =
t_n \land \bigwedge \posi{\beta}[p\vert W] \land
r(\textbf{Y})}
\end{equation}
where $\textbf{X} = X_1, \ldots, X_n$ are $n$ new variables, $\textbf{Y}$
are the variables in $r$, $W$ is a new (second-order) variable and $\posi{\beta}[p\vert W]$ is
$\posi{\beta}$ with $p$
replaced by $W$.
\end{enumerate}
The \emph{completion} of $P$ is $\Comp{P} \equiv \fix{P} \cup
\set{(\ref{eq:constants}),(\ref{eq:oneelem})}$.\end{definition}
The predicate $W$ appears only positively in $\phi(W,\textbf{X})$ such that
the fixed point formula in $(\ref{eq:fpf})$ is well-defined.
By the first disjunct in $(\ref{eq:fpformula})$, we have that applying
the operator $\operphi{(U,M)}$ (see pp. \pageref{eq:oper}) to an arbitrary set $S \subseteq U^n$ does not lose
information from $S$.
\begin{theorem}\label{prop:S}
Let $P$ be a $p$-program and $(U,M)$ an interpretation 
with $S \subseteq U^n$. Then 
\[
S \subseteq \operphi{(U,M)}(S) \; .
\]
\end{theorem}
\begin{proof}
Take $\textbf{x} \in S$, then $(U,M),W\to S \models W(\textbf{x})$, such
that, by $(\ref{eq:fpformula})$, we have $(U,M),W\to S \models
\phi(W,\textbf{x})$. Thus, by $(\ref{eq:oper})$, we have that $\textbf{x} \in
\operphi{(U,M)}(S)$.
\qed
\end{proof}

\begin{example}\label{ex:fixpoint}
Take a $p$-program $P$
\begin{pprogramn}
\nrule{r}{p(X)}{p(X)}
\end{pprogramn}
The completion \Comp{P} contains the formulas $\Exists{X}{\true}$, together with
$\fix{P} \equiv \sat{P} \cup \gl{P} \cup \fpf{P}$, where
\[
\sat{P} = \set{\Forall{X}{\impl{ p(X) }{ p(X) }}} \; ,
\]
ensuring that $r$ is satisfied, and
\[
\gl{P} = \set{ \Forall{ X }{ \equiva{ r(X)} { \true } } } \; ,
\]
saying that $r$ belongs to every GL-reduct since there are
no
naf-atoms. Finally,
\[
\fpf{P} = \set{ \Forall{ X_1 }{ \impl{ p(X_1) }{ \lfp{ W }{ X_1 }{ \phi(W,X_1) }(X_1) } } } \; ,
\]
with
\[
\phi(W, X_1) \equiv W(X_1) \lor \Exists{X}{X_1 = X \land W(X) \land r(X) } \; .
\]
The formula $\fpf{P}$ ensures that every atom in a FOL interpretation
is motivated by a fixed point construction, using the available rule
$p(X)\gets p(X)$.
\end{example}
\begin{theorem}\label{th:aset-iff-model}
Let $P$ be a $p$-program. Then, $(U,M)$ is an open answer set of $P$ iff
$(U,M \cup R)$ is a model of $\bigwedge\Comp{P}$, where 
\[
R \equiv \set{r(\textbf{y}) \mid r[\textbf{Y}\mid
\textbf{y}]:\prule{\alpha[]}{\beta[]} \in P_U, M\models \nega{\alpha[]} \cup
\naf{\nega{\beta[]}},
\vars{r} = \textbf{Y}}\; .
\]
\end{theorem}
\begin{proof}
Denote $M \cup R$ as $M'$. \\
\fbox{$\Rightarrow$} For the ``only if" direction, assume $(U,M)$ is
an open answer set of $P$. We show that
$(U,M')$ is a model of $\Comp{P}$. It is not too difficult to show
that $(U,M')$ is a model of 
$(\ref{eq:constants})$,
$(\ref{eq:oneelem})$,
$\sat{P}$, and 
$\gl{P}$. We also have that 
$(U,M')$ is a model of $\fpf{P}$.  Indeed, take 
$\textbf{x}$ for $\textbf{X}$ and assume $p(\textbf{x}) \in M'$.  Thus,
$p(\textbf{x}) \in M$. Since $(U,M)$ is an open answer set we have that
$p(\textbf{x}) \in T^n$ for some $n < \infty$.

\begin{claim}
$\textbf{x} \in \phi^{(U,M')} \uparrow n$, $n < \infty$.
\end{claim}

We prove the claim by induction on $n$.
\begin{description}

\item[$n = 1$ (Base step)]  If $p(\textbf{x}) \in T^1$ there is some
$r': \prule{p(\textbf{x})}{\posi{\beta}[]} \in P_U^M$ originating from $r: \prule{p(\textbf{t})
\lor \alpha}{\beta} \in P$ with variables $\textbf{Y} = Y_1, \ldots, Y_d$ 
such that for $[\textbf{Y}\vert\textbf{y}]$, $r[] = r'$ (and thus $t_i[] =
x_i$ for $1 \leq i \leq n$). Furthermore, we have
	\begin{itemize}
	\item $\emptyset \models \posi{\beta}[]$\footnote{$\posi{\beta}$ may
	contain equalities but no regular atoms.}, 
	\item $M \models \nega{\alpha}[]$, and 
	\item $M \models \naf{\nega{\beta}[]}$.
	\end{itemize}
Thus $\bigwedge \nega{\alpha}[]$ and $\bigwedge \neg \nega{\beta}[]$ are true in $M'$,
such that, by definition of $M'$, $r(\textbf{y}) \in M'$.  It follows
immediately that $E(r)$ is true in $M'$. Since $\emptyset \models \posi{\beta}[]$ 
we do not use $W$ to deduce the latter, such that
$(U,M'),W\to \emptyset \models \phi(W,\textbf{x})$, and thus $\textbf{x} \in
\operphi{(U,M')}(\emptyset) = \operphi{(U,M')} \uparrow 1$.

\item[(Induction)]
Assume for every $p(\textbf{u}) \in T^{n-1}$ that $\textbf{u} \in \phi^{(U,M')}
\uparrow n-1$, $n-1 < \infty$. From $p(\textbf{x}) \in T^n$, we have some
$r': \prule{p(\textbf{x})}{\posi{\beta}[\textbf{Y}\vert\textbf{y}}] \in P_U^M$ originating from $r: \prule{p(\textbf{t})
\lor \alpha}{\beta} \in P$ with variables $\textbf{Y} = Y_1, \ldots, Y_d$ and
such that for $[\textbf{Y}\vert\textbf{y}]$, $r[] = r'$ (and thus $t_i[] =
x_i$ for $1 \leq i \leq n$). Furthermore, we have
	\begin{itemize}
	\item $T^{n-1} \models \posi{\beta}[]$, 
	\item $M \models \nega{\alpha}[]$, and 
	\item $M \models \naf{\nega{\beta}[]}$.
	\end{itemize}
Thus $\bigwedge \nega{\alpha}[]$ and $\bigwedge \neg \nega{\beta}[]$ are true in $M'$,
such that, by definition of $M'$, $r(\textbf{y}) \in M'$.  
Since $P$ is a $p$-program $\beta$ contains only $p$-literals and
(in)equalities.  Furthermore, the equalities in $\posi{\beta}[]$
are true in $M'$. For every regular $p(\textbf{u}) \in
\posi{\beta}[]$, we have that $p(\textbf{u}) \in T^{n-1}$, and thus, by induction, that $\textbf{u} \in \phi^{(U,M')}
\uparrow n-1$. We have that $(U,M'),W\to \phi^{(U,M')}
\uparrow n-1\models E(r)[\textbf{X}\vert\textbf{x}]$, such that $(U,M'),W\to \phi^{(U,M')}
\uparrow n-1\models \phi(W,\textbf{x})$.  Thus $\textbf{x}
\in\phi^{(U,M')}\uparrow n$.

\end{description}

From $\textbf{x} \in\phi^{(U,M')}\uparrow n$, $n < \infty$,
we have that $\textbf{x} \in
\phi^{(U,M')}\uparrow n \subseteq \phi^{(U,M')}\uparrow \alpha$, for a
limit ordinal $\alpha$ such that $\phi^{(U,M')}\uparrow \alpha =
\lfpoint{\operphi{(U,M')}}$. Then, we have that $\textbf{x} \in
\lfpoint{\operphi{(U,M')}}$, and consequently,
$\lfp{W}{\textbf{X}}{\phi(W,\textbf{X})}(\textbf{x})$ is true in $(U,M')$ such
that $(\ref{eq:fpf})$ is satisfied.

\fbox{$\Leftarrow$} For the ``if" direction, assume $(U,M')$ is a
model of $\Comp{P}$.
We show that
$(U,M)$ is an open answer set of $P$.  Denote $\set{\textbf{x} \mid
p(\textbf{x}) \in M}$ as $\overline{M}$.
\begin{enumerate}
\item From $(\ref{eq:constants})$ and $(\ref{eq:oneelem})$, we have
that $U$ is non-empty and interprets different constants as
different elements.  We assume that the elements that interpret the constants in $U$ have the
same name as those constants.
\item $\overline{M} = \lfpoint{\operphi{(U,M')}}$.  
	\begin{itemize}
	\item $\overline{M} = \operphi{(U,M')}(\overline{M})$.
		\begin{itemize}
		\item $\overline{M} \subseteq
		\operphi{(U,M')}(\overline{M})$. Immediate, with
		Theorem \ref{prop:S}.

		\item $\overline{M} \supseteq
		\operphi{(U,M')}(\overline{M})$.  Assume $\textbf{x} \in
		\operphi{(U,M')}(\overline{M})$. Then by
		($\ref{eq:oper})$, we have that $(U,M'),W\to \overline{M} \models \phi(W,\textbf{x})$. 
		Thus, by $(\ref{eq:fpformula})$, we
		have either that $\textbf{x} \in \overline{M}$, which
		means we are done, or there is a $r:\prule{p(\textbf{t}) \lor
		\alpha}{\beta} \in P$ such that $(U,M'),W\to
		\overline{M} \models
		E(r)[\textbf{X}\vert\textbf{x}]$.
		\par
		Then, there exist $[\textbf{Y}\vert \textbf{y}]$ with
		\begin{itemize}
		\item $\textbf{x} = \textbf{t}[]$, 
		\item $(U,M'),W\to \overline{M}\models \posi{\beta}[p\vert W][]$,
		such that $M' \models \posi{\beta}[]$,
		and
		\item $r(\textbf{y}) \in M'$, from which, since $M'$ is a
		model of $\gl{P}$, we have that $M' \models
		\bigwedge \nega{\alpha}[]$ and $M' \models \bigwedge \neg{\nega{\beta}[]}$. 
		\end{itemize}
		Since $M'$ is a model of $\sat{P}$ we then have that
		$p(\textbf{t})[] \in M'$ and thus $p(\textbf{x}) \in M$,
		such that $\textbf{x} \in \overline{M}$.

		\end{itemize}
		
	\item $\overline{M}$ is a least fixed point.
	Assume there is a $Y \subseteq U^n$ such that $Y =
	\operphi{(U,M')}(Y)$. We prove that $\overline{M} \subseteq
	Y$. Take $\textbf{x} \in \overline{M}$, then $p(\textbf{x}) \in M'$.
	Since $M'$ is a model of $\fpf{P}$, we have that $\textbf{x} \in \lfpoint{\operphi{(U,M')}}$. 
	And since $\lfpoint{\operphi{(U,M')}} \subseteq Y$, we have
	that $\textbf{x} \in Y$.

	\end{itemize}

\item $M$ is a model of $P_U^M$. Take a rule $r':
\prule{p(\textbf{x})}{\posi{\beta}[\textbf{Y}\vert{y}]} \in P_U^M$ originating from $r: \prule{p(\textbf{t})
\lor \alpha}{\beta} \in P$ with variables $\textbf{Y} = Y_1, \ldots, Y_d$ and
such that for $[\textbf{Y}\vert\textbf{y}]$, $r[] = r'$ (and thus $t_i[] =
x_i$ for $1 \leq i \leq n$). Furthermore, we have
	\begin{itemize}
	\item $M \models \nega{\alpha}[]$, 
	\item $M \models \naf{\nega{\beta}[]}$.
	\end{itemize}

Assume $M \models \posi{\beta}[]$, we then have that 
	\begin{itemize}
	\item $M' \models \nega{\alpha}[]$, 
	\item $M' \models \naf{\nega{\beta}[]}$,
	\item $M' \models \posi{\beta}[]$.
	\end{itemize}
Since $M'$ is a model of $\sat{P}$, we then have that $p(\textbf{x}) \in
M'$, and thus $p(\textbf{x}) \in M$.

\item $M$ is a minimal model of $P_U^M$. Assume not, then there is a
$N \subset M$, $N$ a model of $P_U^M$.  Take $\overline{N} =
\set{\textbf{x} \mid p(\textbf{x}) \in N}$, one can then show that $\overline{N}$ is a
fixed point of $\operphi{(U,M')}$, i.e., $\overline{N} = \operphi{(U,M')}(\overline{N})$.
Since
$\overline{M}=\lfpoint{\operphi{(U,M')}}$, we have that
$\overline{M}\subseteq \overline{N}$, which is a contradiction with $N
\subset M$, and $M$ is indeed a minimal model of $P_U^M$.
\end{enumerate}
\qed
\end{proof}

\begin{example}
For a universe $U = \set{x}$ we have the unique open answer set
$(U,\emptyset)$ of $P$ in Example \ref{ex:fixpoint}.
Since $U$ is non-empty, every open answer set with a universe $U$
satisfies $\Exists{X}{\true}$.
Both $(U,M_1 = \set{p(x),r(x)})$ and $(U,M_2 = \set{r(x)})$
satisfy $\sat{P} \cup \gl{P}$.  However, $\lfpoint{\operphi{(U,M_1)}} = 
\lfpoint{\operphi{(U,M_2)}} = \emptyset$, such that only $(U,M_2)$
satisfies $\fpf{P}$; $(U,M_2)$ corresponds exactly to the open answer set
$(U,\emptyset)$ of $P$.
\end{example}
The completion in Definition \ref{def:comp} differs from Clark's
completion \cite{Clark87} both in the presence of the fixed point
construct in $(\ref{eq:fpf})$ and atoms representing membership of the
GL-reduct. For $p$-programs $P$ Clark's Completion \CComp{P} does not
contain \gl{P} and \fpf{P} is replaced by a formula that ensures
support for every atom by an applied rule
\[
\Forall{\textbf{X}}{\impl{p(\textbf{X})}{\bigvee_{r:\prule{p(\textbf{t}) \lor \alpha}{\beta} \in  P}D(r)}}
\]
with 
\[
D(r) \equiv \Exists{\textbf{Y}}{X_1 = t_1 \land \ldots \land X_n =
t_n \land \bigwedge {\beta} \land \bigwedge \nega{\alpha}} \; .
\]
Program $P$ in Example \ref{ex:fixpoint} is the open ASP version of
the classical example \prule{p}{p} \cite{lee:loop}. There are
FOL models of \CComp{P} that do not correspond to any open
answer sets: both $(\set{x}, \set{p(x)})$ and $(\set{x},\emptyset)$
are FOL models while only the latter is an open answer set of $P$.
The next example shows the translation to FPL in detail.
\begin{example}\label{ex:fixpoint2}
Take the p-program $P$ corresponding to the program consisting of the rules
$\prule{a}{\naf{b}}$ and $\prule{b}{\naf{a}}$, i.e.
\begin{pprogramn}
\nrule{r_1}{p(X,a)}{\naf{p(X,b)},X\neq a, X\neq b}
\nrule{r_2}{p(X,b)}{\naf{p(X,a)},X\neq a, X\neq b}
\end{pprogramn}
which has, for a universe $U = \set{x,a,b}$, two open answer sets $M_1
= \set{p(x,a)}$ and $M_2
= \set{p(x,b)}$. \sat{P} contains the formulas
\[
\Forall{X}{\impl{ \neg p(X,b) \land X \neq a \land X \neq b }{ p(X,a) } } \; ,
\]
and 
\[
\Forall{X}{\impl{ \neg p(X,a) \land X \neq a \land X \neq b }{ p(X,b) } } \; .
\]
\gl{P} is defined by the formulas $\Forall{X}{\equiva{r_1(X)}{\neg
p(X,b)\land X\neq a \land X \neq b}}$ 
and $\Forall{X}{\equiva{r_2(X)}{\neg p(X,a)\land X\neq a \land X \neq b}}$.  Finally, \fpf{P} is
\[
\Forall{ X_1,X_2 }{ \impl{ p(X_1,X_2) }{ \lfp{ W }{ X_1,X_2 }{ \phi(W, X_1, X_2) }(X_1,X_2) } }
\]
with 
\[
\begin{split}
\phi(W, X_1,X_2) \equiv &~ W(X_1,X_2) \\
		& \lor \Exists{X}{X_1 = X \land X_2 = a \land r_1(X) } \\
		& \lor \Exists{X}{X_1 = X \land X_2 = b \land r_2(X) }  \; .
\end{split}
\]
To satisfy \sat{P} a model must contain $p(x,a)$ or $p(x,b)$.  Taking
into account $\gl{P}$, we then distinguish three different classes of
models, represented by 
\[
\begin{split}
M'_1 & \models \set{p(x,a),\neg p(x,b),r_1(x), \neg r_2(x)}\; ,\\
M'_2 & \models \set{\neg p(x,a),p(x,b), \neg r_1(x), r_2(x)}\; ,\\
M'_3 & \models \set{p(x,a),p(x,b), \neg r_1(x), \neg r_2(x)}\; . 
\end{split}
\]
Now, we have that $\lfpoint{\operphi{(U,M'_3)}}= \emptyset$, such that
$\fpf{P}$ is not satisfied by $M'_3$.  Furthermore,
$\lfpoint{\operphi{(U,M'_1)}} = \set{(x,a)}$ and
$\lfpoint{\operphi{(U,M'_2)}} = \set{(x,b)}$. Thus, in order to
satisfy $\fpf{P}$, we have that $M'_1 = \set{p(x,a),r_1(x)}$
 and $M'_2 =
\set{p(x,b),r_2(x)}$,
which correspond to the open answer sets of $P$.
\par
Note that this example also shows that writing knowledge down in Logic Programming style is easier and more intuitive than the corresponding FPL translation.
\end{example}
\begin{theorem}\label{prop:complinear}
Let $P$ be a $p$-program. The size of $\bigwedge \Comp{P}$ is quadratic in the
size of $P$.
\end{theorem}
\begin{proof}
If the number of constants in a program $P$ is $c$, then the number of
formulas $(\ref{eq:constants})$ is $\frac{1}{2}c(c-1)$, which yields the
quadratic bound.  The size of $\sat{P}$ is linear in the size of $P$,
as is
the size of \gl{P} (with $\card{P}$ new
predicates).  Finally, each $E(r)$ in $\fpf{P}$ is linear in the size
of $r$, such that $\fpf{P}$ is linear in the size of $P$.
\qed
\end{proof}
\begin{theorem}\label{th:sat1}
Let $P$ be a program, $p$ a predicate not appearing in $P$, and $q$ an $n$-ary predicate in $P$. $q$ is satisfiable
w.r.t. $P$ iff $p(\textbf{X},\textbf{0},q) \land \bigwedge\Comp{P_p}$ is satisfiable.
Moreover, this reduction is polynomial in the size of $P$.
\end{theorem}
\begin{proof}
Assume $q$ is satisfiable w.r.t. $P$.  By Theorem
\ref{prop:p-program}, we have that $p(\textbf{x},\textbf{0},q)$ is in an
open answer set of \pprog{P}{p}, such that, with Theorem
\ref{th:aset-iff-model}, $p(\textbf{x},\textbf{0},q)$ is in a model of
\Comp{P_p}.
\par
For the opposite direction, assume $p(\textbf{X},\textbf{0},q)\land
\bigwedge\Comp{P_p}$ is satisfiable.  Then there is a model $(U,M')$ of
$\bigwedge\Comp{P}$ with $p(\textbf{x},\textbf{0},q) \in M'$. We have that $M' = M \cup
R$ as in Theorem \ref{th:aset-iff-model}, such that $(U,M)$ is an open
answer set of $P_p$ and $p(\textbf{x},\textbf{0},q) \in M$.  From
Theorem \ref{prop:p-program}, we then have that $q$ is satisfiable
w.r.t. $P$.
\par
By Theorem \ref{prop:complinear}, the size of $\bigwedge\Comp{\pprog{P}{p}}$
is quadratic in the size of \pprog{P}{p}.  Since the size of the
latter is polynomial in the size of $P$ by Theorem \ref{prop:compl},
the size of $\bigwedge\Comp{\pprog{P}{p}}$ is polynomial in the size of $P$.
\qed
\end{proof}

\section{Guarded Open Answer Set Programming}\label{sec:goasp}

In this section, we will identify a syntactically restricted class of programs such that
the translation to FPL falls within a decidable fragment of FPL and which enables us to devise
some complexity result for satisfiability checking. Intuitively, rules will be equipped with
a guard, i.e. a set of atoms, in the positive body, such that every pair of variables in the 
rule appears together in an atom in that guard.
\par
We repeat the definitions of the \emph{loosely guarded
fragment} \cite{benthem} of
first-order logic as in \cite{gradel}: \textit{
The \emph{loosely guarded fragment LGF} of first-order logic is defined inductively as
follows:}
\begin{enumerate}
\item[(1)] \textit{Every relational atomic formula belongs to LGF.}

\item[(2)] \textit{LGF is closed under propositional connectives $\neg$, $\land$,
$\lor$, $\Rightarrow$, and $\Leftrightarrow$.
}
\item[(3)] \textit{
If $\psi(\textbf{X},\textbf{Y})$\footnote{
Recall that $\psi(\textbf{X},\textbf{Y})$ denotes a formula whose free
variables are all among $\textbf{X}\cup\textbf{Y}$ (\cite{andreka98}, pp.
236).} 
is in LGF, and $\alpha(\textbf{X},\textbf{Y}) =
\alpha_1 \land \ldots \land \alpha_m$ is a conjunction of atoms,
then the formulas
\[
\begin{array}{c}
\Exists{\textbf{Y}}{\alpha(\textbf{X},\textbf{Y}) \land \psi{(\textbf{X},\textbf{Y})}} \\
\Forall{\textbf{Y}}{\alpha(\textbf{X},\textbf{Y}) \Rightarrow \psi{(\textbf{X},\textbf{Y})}}
\end{array}
\]
belong to LGF (and $\alpha(\textbf{X},\textbf{Y})$ is the
\emph{guard} of the
formula), provided that $\mbox{free}(\psi) \subseteq \mbox{free}(\alpha) = \textbf{X} \cup
\textbf{Y}$ and for every quantified variable $Y \in \textbf{Y}$ and every
variable $Z \in \textbf{X} \cup \textbf{Y}$ there is at least one atom
$\alpha_j$ that contains both $Y$ and $Z$
(where $\mbox{free}(\psi)$ are the free variables of $\psi$).
}
\end{enumerate}
The \emph{loosely guarded fixed point logic}
\mulgf{} is
LGF extended with fixed point formulas (\ref{eq:fix}) where
$\psi(W,\textbf{X})$ is a $\mulgf$ 
 formula
such that $W$ does not appear in
guards. The \emph{guarded fragment} GF is defined as LGF but with the
guards $\alpha(\textbf{X},\textbf{Y})$ atoms instead of a conjunction of
atoms. The \emph{guarded fixed point logic} \mugf{} is GF extended with
fixed point formulas where $\psi(W,\textbf{X})$ is a \mugf{} formula such
that $W$ does not appear in guards.
\begin{example}
The infinity axiom in Example \ref{ex:infinite} (pp.
\pageref{ex:infinite}) is a \mugf{} formula where all the formulas are
guarded by $F(X,Y)$.
\end{example}
\begin{example}[\cite{gradel}] Take the formula
\[
\Exists{Y}{X\leq Y \land \varphi(Y) \land \left(\Forall{Z}{(X\leq Z \land Z
< Y)\Rightarrow
\psi(Z)\right)}}\; .
\]
This formula is not guarded as the formula $\Forall{Z}{(X\leq Z \land Z
< Y)\Rightarrow
\psi(Z)}$ has no atom as guard.  It is however loosely guarded.
\end{example}

\begin{definition}\label{def:loosely-guard}
A rule $r:\prule{\alpha}{\beta}$ is \emph{loosely guarded} if there is a
$\gamma_b \subseteq {\posi{\beta}}$ such that
every two variables $X$ and $Y$ from $r$ appear together in an atom
from $\gamma_b$; we call $\gamma_b$ a \emph{body guard} of $r$.  It is
\emph{fully loosely guarded} if it is loosely guarded and there is a $\gamma_h
\subseteq {\nega{\alpha}}$ such that every two
variables $X$ and $Y$ from $r$ appear together in an atom from
$\gamma_h$; $\gamma_h$  is called a \emph{head guard} of $r$. 
\par
A program $P$ is a \emph{(fully) loosely guarded program ((F)LGP)} if every
non-free rule in $P$ is {(fully) loosely guarded}. 
\end{definition}
\begin{example}
The rule in Example \ref{ex:fixpoint} is loosely guarded but not
fully loosely guarded. The program in Example \ref{ex:fixpoint2} is neither
fully loosely guarded nor loosely guarded.  A rule\[
\prule{a(X)\lor\naf{g(X,Y,Z)}}{\naf{b(X,Y)},f(X,Y),f(X,Z),h(Y,Z),\naf{c(Y)}}
\]
has a body
guard $\set{f(X,Y),f(X,Z),h(Y,Z)}$ and a head guard $\set{g(X,Y,Z)}$.
\end{example}
\begin{definition}\label{def:guard}
A rule $r:\prule{\alpha}{\beta}$ is \emph{guarded} if it is loosely
guarded with a singleton body guard.
It is
\emph{fully guarded} if it is fully loosely guarded with body and head guards singleton sets.
\par
A program $P$ is a \emph{(fully) guarded program
((F)GP)} if every
non-free rule in $P$ is {(fully) guarded}. 
\end{definition}
In \cite{gradel2} it is noted that a singleton set $\set{b}\subseteq
U$ for a universe $U$ is always guarded by an atom $b = b$.  With a
similar reasoning one sees that rules with only one variable $X$ can
be made guarded by adding the guard $X = X$ to the body. E.g.,
\prule{a(X)}{\naf{b(X)}} is equivalent to \prule{a(X)}{X =
X,\naf{b(X)}}.
\par
Every F(L)GP is a (L)GP, and we can rewrite every
(L)GP as a F(L)GP.
\begin{example}
The rule \prule{p(X)}{p(X)} can be rewritten as
\prule{p(X)\lor\naf{p(X)}}{p(X)} where the body guard is added
to the negative part of the head to function as the head guard.  Both
programs are equivalent: for a universe $U$, both have the unique
open answer set $(U,\emptyset)$.
\end{example}
Formally, we can rewrite every (L)GP $P$ as
an equivalent F(L)GP $\hbg{P}$, where \hbg{P} is $P$ with every
 $\prule{\alpha}{\beta}$ replaced by $\prule{\alpha \cup
\naf{\posi{\beta}}}{\beta}$.
\par
One can consider the body guard of a rule in a loosely guarded program
$P$ as the head guard such that \hbg{P} is indeed a fully (loosely)
guarded program.
\begin{theorem}\label{prop:hbg2}
Let $P$ a (L)GP. Then, \hbg{P} is a F(L)GP.
\end{theorem}
\begin{proof}
Let $P$ be a (L)GP.  We show that every non-free rule
$r:\prule{\alpha\cup \naf{\posi{\beta}}}{\beta}\in \hbg{P}$ is fully
(loosely) guarded.  Since
$\prule{\alpha}{\beta}$ is a non-free rule of $P$, we have that there
is a body guard $\gamma_b \subseteq \posi{\beta}$, and thus $r$ is
(loosely) guarded.  Furthermore, $\gamma_b \subseteq \nega{(\alpha\cup
\naf{\posi{\beta}})}$ such that $\gamma_b$ is a head guard of $r$ and
$r$ is fully (loosely) guarded.
\qed
\end{proof}
A rule is vacuously satisfied if the body of a rule in \hbg{P} is
false and consequently the head does not matter; if the body is true
then the newly added part in the head becomes false and the rule in
\hbg{P} reduces to its corresponding rule in $P$.
\begin{theorem}\label{prop:hbg}
Let $P$ be any program (not necessarily guarded).
An open interpretation $(U,M)$ of $P$ is an open answer set of $P$ iff
$(U,M)$ is an open answer set of $\hbg{P}$.
\end{theorem}
Since we only copy (a part of) the bodies to the heads, the size of
\hbg{P} only increases linearly in the size of $P$.
\begin{theorem}\label{prop:hbglinear}
Let $P$ be any program (not necessarily guarded).  The size of \hbg{P} is linear in the size of $P$.
\end{theorem}
We have that the construction of a $p$-program retains the 
guardedness properties.
\begin{theorem}\label{prop:hbg3}
Let $P$ be any program (not necessarily guarded). Then, $P$ is a (F)LGP iff $\pprog{P}{p}$ is a
(F)LGP. And similarly for (F)GPs.
\end{theorem}
\begin{proof}
We only prove the LGP case, the cases for FLGPs and (F)GPs are similar.
\par
For the ``only if" direction, take a non-free rule
$\pprog{r}{p}:\prule{\pprog{\alpha}{p}}{\pprog{\beta}{p}, \lit{in(\textbf{X})}} \in \pprog{P}{p}$ and two variables $X$ and $Y$ in $r_p$.  We
have that $r:\prule{\alpha}{\beta}$ is a non-free rule in $P$ by the
construction of $P_p$ and $X$
and $Y$ are two variables in $r$, such that there is a $\gamma
\subseteq \posi{\beta}$ with either a regular atom $q(\textbf{t})$ that contains
$X$ and $Y$ or an equality atom $X = Y$ in $\gamma$.  In the former case, we have
that $p(\textbf{t},\textbf{0},q) \in \gamma_p  \subseteq \posi{\beta_p}$
such that $r_p$ is loosely guarded.  In the latter case, $X = Y \in \gamma_p$
such that again $r_p$ is loosely guarded.
\par
For the ``if" direction, take a non-free $r:\prule{\alpha}{\beta} \in
P$ and two variables $X$ and $Y$ in $r$.  Then
$\pprog{r}{p}:\prule{\pprog{\alpha}{p}}{\pprog{\beta}{p}, \lit{in(\textbf{X})}}$ is non-free in \pprog{P}{p} and $X$ and $Y$ are variables in
$r_p$.  Thus, there is a $\gamma_p \subseteq \posi{(\beta_p\cup \lit{in(\textbf{X})})} = \posi{\beta_p}$ with an atom containing the two variables
$X$ and $Y$.  Then $\gamma \subseteq \posi{\beta}$ with an atom in
$\gamma$ containing $X$ and $Y$.
\qed
\end{proof}
For a fully (loosely) guarded $p$-program $P$, we can rewrite \Comp{P} as the
equivalent \mulgfbrack{} formulas \GComp{P}.  \GComp{P} is \Comp{P}
with the following modifications.
\begin{itemize}
\item Formula $(\ref{eq:oneelem})$ is replaced by
\begin{equation}\label{eq:oneelem1}\Exists{X}{X = X}\; ,\end{equation}
such that it is guarded by $X = X$.

\item Formula $(\ref{eq:sat})$ is removed if $r:\prule{\alpha}{\beta}$ is free or otherwise replaced by
\begin{equation}\label{eq:sat2}
\Forall{\textbf{Y}}{\bigwedge \gamma_b \Rightarrow \bigvee \alpha \lor
\bigvee \neg (\setmin{\posi{\beta}}{\gamma_b})} \lor \bigvee
\nega{\beta}\; ,
\end{equation}
where $\gamma_b$ is a body guard of $r$, thus we have
logically rewritten the formula such that it is (loosely) guarded.
 If
$r$ is a free rule of the form
\prule{q(\textbf{t})\lor\naf{q(\textbf{t})}}{} we have
$\Forall{Y}{\impl{\true}{q(\textbf{t})\lor \neg q(\textbf{t})}}$ which is
always true and can thus be removed from \Comp{P}.  
\item Formula $(\ref{eq:gl})$ is replaced by the formulas
\begin{equation}\label{eq:gl2}
\Forall{\textbf{Y}}{\impl{ r(\textbf{Y}) }{ \bigwedge \nega{\alpha} \land \bigwedge \neg \nega{\beta} }}
\end{equation}
and
\begin{equation}\label{eq:gl3}
\Forall{\textbf{Y}}{\impl{ \bigwedge \gamma_h }{ r(\textbf{Y}) \lor  \bigvee \nega{\beta} \lor \bigvee \neg (\setmin{\nega{\alpha}}{\gamma_h})  }}
\end{equation}
where $\gamma_h$ is a head guard of \prule{\alpha}{\beta}. 
We thus rewrite an
equivalence as two implications where the first implication is guarded
by $r(\textbf{Y})$ and the second one is (loosely) guarded by the head guard of the
rule -- hence the need for a fully (loosely) guarded program, instead of just a
(loosely) guarded one.

\item For every $E(r)$ in $(\ref{eq:fpf})$, replace $E(r)$ by 
\begin{equation}\label{eq:Er2}
E'(r) \equiv \bigwedge_{t_i \not\in \textbf{Y}} X_i = t_i \land \Exists{\textbf{Z}}{(\bigwedge \posi{\beta}[p\vert W] \land
r(\textbf{Y}))[t_i \in \textbf{Y} \vert X_i]}\; ,
\end{equation}
with $\textbf{Z} = \setmin{\textbf{Y}}{\set{t_i\mid t_i\in \textbf{Y}}}$, i.e., move all $X_i = t_i$ where $t_i$ is
constant out of the scope of the quantifier, and remove the others by
substituting each $t_i$ in $\bigwedge \posi{\beta}[p\vert W] \land
r(\textbf{Y})$ by $X_i$. This rewriting makes sure that every variable in
the quantified part of $E'(R)$ is guarded by $r(\textbf{Y})[t_i\in
\textbf{Y}\vert X_i]$.

\end{itemize}
\begin{example}\label{ex:gcomp}
For the fully guarded $p$-program $P$ containing a rule \[
\prule{p(X)\lor\naf{p(X)}}{p(X)} 
\]
with body and
head guard $\set{p(X)}$, 
$\sat{P} = \set{\Forall{X}{\impl{ p(X) }{ p(X)\lor \neg
p(X)}}}$, 
$\gl{P} = \set{ \Forall{ X }{ \equiva{ r(X)} { p(X) } } }$ and
the formula $\phi(W,X_1)$ in \fpf{P} is $\phi(W,X_1) \equiv W(X_1) \lor
\Exists{X}{X_1 = X \land W(X) \land r(X) }$. \GComp{P} translates
$\sat{P}$ identically and rewrites the equivalence of $\gl{P}$ as two
implications resulting in guarded rules.  The rewritten $\phi(W,X_1)$
is $W(X_1) \lor (W(X_1) \land r(X_1))$. There is no quantification
anymore in this
formula since $X$ was substituted by $X_1$. Clearly, for a universe
$\set{x}$, we have that the open answer set of the program is
$(\set{x},\emptyset)$, which corresponds with the unique model of
\GComp{P} for a universe $\set{x}$. 
\end{example}
The translation \GComp{P} is logically equivalent to \Comp{P} and,
moreover,  it contains only formulas in
(loosely) guarded fixed point logic.
\begin{theorem}\label{prop:gcomp}
Let $P$ be a fully (loosely) guarded $p$-program.  $(U,M)$ is a model of
$\bigwedge\Comp{P}$ iff $(U,M)$ is a model of $\bigwedge\GComp{P}$.
\end{theorem}
\begin{proof}
This can be shown by using standard logical equivalences.
\qed
\end{proof}
\begin{theorem}\label{th:gcomp}
Let $P$ be a fully (loosely) guarded $p$-program. Then, the formula $\bigwedge \GComp{P}$ is a
\mulgfbrack{} formula.
\end{theorem}
\begin{proof}
We first show that $\lfp{W}{\textbf{X}}{\phi'(W,\textbf{X})}(\textbf{X})$ is
a valid fixed point formula,
with $\phi'(W,\textbf{X})$ equal to $\phi(W,\textbf{X})$ with $E'(r)$ instead
of $E(r)$. We have that all free variables are still in $\textbf{X}$, 
since only $X_i = t_i$ where $t_i$ is constant is moved out of the
scope of the quantifier in $E(r)$ and all other $t_i$ where
substituted by $X_i$ such that $\textbf{Z}$ in $E(r)$ bounds all other
variables than $\textbf{X}$. Furthermore, $p$ appears only
positively in $\phi'$.
\par
We next show that $\bigwedge\GComp{P}$ is a \mulgf{} formula if $P$ is fully
loosely guarded; the treatment for \mugf{} formulas if $P$ is fully
guarded is similar. 

\begin{itemize}

\item  Formula $(\ref{eq:oneelem1})$ is guarded with guard
$X = X$.

\item Formula $(\ref{eq:sat2})$ corresponds with a non-free rule
\prule{\alpha}{\beta} with a body guard $\gamma_b$; thus
$\vars{\prule{\alpha}{\beta}} \subseteq \vars{\gamma_b}$.
	\begin{itemize}
	
	\item $\mbox{free}({\bigvee \alpha \lor \bigvee \neg
	(\setmin{\posi{\beta}}{\gamma_b}) \lor \bigvee \nega{\beta}})
	\subseteq \textbf{Y} = \vars{\prule{\alpha}{\beta}} =
	\vars{\gamma_b} = \mbox{free}(\bigwedge \gamma_b)$.

	\item Take two variables $Y_i$ and $Y_j$ from $\textbf{Y}$, then $Y_i \in 
	\vars{\prule{\alpha}{\beta}}$ and $Y_j \in 
	\vars{\prule{\alpha}{\beta}}$, such that $Y_i$ and $Y_j$ are
	in an atom from $\gamma_b$.
	
	\end{itemize}

\item Formula $(\ref{eq:gl2})$ is guarded with guard
$r(\textbf{Y})$.

\item Formula $(\ref{eq:gl3})$:
	\begin{itemize}
	\item For a non-free rule \prule{\alpha}{\beta} with a head
	guard $\gamma_h$. Can be done similarly as formula $(\ref{eq:sat2})$.

	\item If \prule{\alpha}{\beta} is free, i.e., of the form
	$\prule{q(\textbf{t})\lor \naf{q(\textbf{t})}}{}$, we have that
	$\gamma_h = \set{q(\textbf{t})}$, and formula $(\ref{eq:gl3})$ is
	of the form $\Forall{\textbf{Y}}{\impl{q(\textbf{t})}{r(\textbf{Y})}}$.

	\begin{itemize}
	
	\item $\mbox{free}({r(\textbf{Y}}))
	= \textbf{Y} = \vars{\prule{\alpha}{\beta}} =
	\vars{q(\textbf{t})} = \mbox{free}(\bigwedge \gamma_h)$.

	\item Take two variables $Y_i$ and $Y_j$ from $\textbf{Y}$, then $Y_i \in 
	\vars{\prule{\alpha}{\beta}}$ and $Y_j \in 
	\vars{\prule{\alpha}{\beta}}$, such that $Y_i$ and $Y_j$ are
	in $\vars{q(\textbf{t})} = \mbox{free}(\gamma_h)$.
	
	\end{itemize}

	\end{itemize}

\item For the last case, we need to show that $\phi'(\textbf{X})$ is a
\mulgf{} formula where $W$ does not appear as a guard.  We show that
for each $r:\prule{\alpha}{\beta}$, $\Exists{\textbf{Z}}{(\bigwedge
\posi{\beta}[p\vert W] \land r(\textbf{Y}))[t_i \in \textbf{Y} \vert X_i]}$ is a 
guarded formula with guard $r(\textbf{Y})[]$. 
Thus $W$ does not appear as a guard.
	\begin{itemize}
	
	\item $\mbox{free}({(\bigwedge \posi{\beta}[p\vert W] \land r(\textbf{Y}))[t_i \in \textbf{Y} \vert X_i]})
	= \setmin{\textbf{Y}}{\set{t_i\mid t_i\in \textbf{Y}}}\cup\set{X_i\mid t_i\in \textbf{Y}} = 
	\mbox{free}(r(\textbf{Y})[])$.

	\item Take a quantified variable $Z\in \setmin{\textbf{Y}}{\set{t_i\mid t_i\in \textbf{Y}}}$
	and $U$ from $\setmin{\textbf{Y}}{\set{t_i\mid t_i\in \textbf{Y}}}\cup\set{X_i\mid t_i\in \textbf{Y}}$,
	then $Z$ and $U$ appear in $r(\textbf{Y})[]$.
	
	\end{itemize}

\end{itemize}
\qed
\end{proof}
Since \GComp{P} is just a logical rewriting of \Comp{P} its size is
linear in the size of \Comp{P}.  
\begin{theorem}\label{prop:gcomplinear}
Let $P$ be a fully (loosely) guarded $p$-program.  The size of the formula \GComp{P} is
linear in the size of $\Comp{P}$. 
\end{theorem}
\begin{proof}
The size of formula $(\ref{eq:oneelem1})$ is linear in the size of
$(\ref{eq:oneelem})$. Formula $(\ref{eq:sat2})$ is just a shuffling of
$(\ref{eq:sat})$. Every  formula $(\ref{eq:gl})$ is replaced by two
shuffled formulas. Finally, $E'(r)$ is $E(r)$ with the movement of some
atoms and applying a substitution, thus the size of $E'(r)$ is linear
in the size of $E(r)$.
\qed
\end{proof}
\begin{theorem}\label{th:sat2}
Let $P$ be a (L)GP and $q$ an $n$-ary predicate in $P$. $q$ is satisfiable
w.r.t. $P$ iff $p(\textbf{X},\textbf{0},q) \land \bigwedge\GComp{\pprog{(\hbg{P})}{p}}$ is satisfiable. 
Moreover, this reduction is polynomial in the size of $P$.
\end{theorem}
\begin{proof}
By Theorem \ref{prop:hbg2} and \ref{prop:hbg3}, we have that
\pprog{(\hbg{P})}{p} is a fully (loosely) guarded $p$-program, thus the formula
$\bigwedge\GComp{\pprog{(\hbg{P})}{p}}$ is defined. By Theorem
\ref{prop:hbg}, we have that $q$ is satisfiable w.r.t. $P$ iff $q$ is
satisfiable w.r.t. \hbg{P}. By Theorem \ref{th:sat1}, we have that
$q$ is satisfiable w.r.t.  \hbg{P} iff $p(\textbf{X},\textbf{0},q) \land
\Comp{\pprog{(\hbg{P})}{p}}$ is satisfiable. Finally, Theorem
\ref{prop:gcomp} yields that $q$ is satisfiable w.r.t. $P$ iff
$p(\textbf{X},\textbf{0},q) \land \bigwedge\GComp{\pprog{(\hbg{P})}{p}}$ is
satisfiable. 
\par
Theorem \ref{prop:hbglinear}, Theorem \ref{th:sat1}, and
Theorem \ref{prop:gcomplinear} yield that this reduction is
polynomial.
\qed
\end{proof}
For a (L)GP $P$, we have, by Theorem
\ref{th:gcomp}, that $\bigwedge\GComp{\pprog{(\hbg{P})}{p}}$ is a \mulgfbrack{}
formula such that the formula $p(\textbf{X},\textbf{0},q) \land
\bigwedge\GComp{\pprog{(\hbg{P})}{p}}$ is as well.
Since satisfiability checking for \mulgfbrack{} is \exptimex{2}-complete
(Theorem [1.1] in \cite{gradel}), satisfiability
checking w.r.t. $P$ is in \exptimex{2}.
\begin{theorem}\label{th:dec}
Satisfiability
checking w.r.t. (L)GPs is in \exptimex{2}.
\end{theorem}
An answer set of a program $P$ (in contrast with an \emph{open} answer
set) is defined as an answer
set of the grounding of $P$ with its constants, i.e., $M$ is an answer set
of $P$ if it is a minimal model of $P_{\cts{P}}^M$.  As is common in
literature, we assume $P$ contains at least one constant.
\par
We can make any program loosely guarded and reduce the answer set semantics
for programs to the open answer set semantics for loosely guarded programs.
For a program $P$, let \gua{P} be the program $P$, such that for each
rule $r$ in $P$ and for each pair of variables $X$ and $Y$ in $r$,
$g(X,Y)$ is added to the body of $r$.  Furthermore, add
\prule{g(a,b)}{} for every $a,b\in \cts{P}$. Note that we assume,
without loss of generality,
that
$P$ does not contain a predicate $g$.
\begin{example}
Take a program $P$
\begin{pprogram}
\tsrule{q(X)}{f(X,Y)}
\tsrule{f(a,Y)\lor\naf{f(a,Y)}}{}
\end{pprogram}
such that $\cts{P} = \set{a}$, and $P$ has answer sets $
\set{f(a,a),q(a)}$ and $\emptyset$.  The loosely guarded program
$\gua{P}$ is 
\begin{pprogram}
\tsrule{q(X)}{g(X,X), g(Y,Y), g(X,Y),f(X,Y)}
\tsrule{f(a,Y)\lor\naf{f(a,Y)}}{g(Y,Y)}
\tsrule{g(a,a)}{}
\end{pprogram}
For a universe $U$, we have the open answer sets
$(U,\set{f(a,a),q(a),g(a,a)}$ and $(U,\set{g(a,a)})$.

\end{example}
The newly added guards in the bodies of rules together with the
definition of those guards for constants only ensure a correspondence
between (normal) answer sets and open answer sets where the universe
of the latter
equals the constants in the program.
\begin{theorem}\label{prop:gua}
Let $P$ be a program. $M$ is an answer set of $P$ iff
$(\cts{P},M\cup\set{g(a,b)\mid a,b \in \cts{P}})$ is
an open answer set of \gua{P}.
\end{theorem}

Note that one can use Theorem \ref{prop:gua} as a definition of \emph{answer
set} of programs with generalized literals, in case one is only interested in
answer sets and not in the open answer sets.

\begin{theorem}\label{prop:gua2}
Let $P$ be a program. The size of \gua{P} is quadratic in the size of
$P$.
\end{theorem}
\begin{proof}
If there are $c$ constants in $P$, we add $c^2$ rules
$\prule{g(a,b)}{}$ to $\gua{P}$.  Furthermore, the size of each rule
grows also grows quadratically, since for a rule with $n$ variables we
add $n^2$ atoms $g(X,Y)$ to the body of $r$.
\qed
\end{proof}
By construction, $\gua{P}$ is loosely guarded.
\begin{theorem}\label{prop:gua3}
Let $P$ be a program. \gua{P} is a LGP.
\end{theorem}
We can reduce checking whether there exists an
answer set containing a literal to satisfiability checking w.r.t. the
open answer set semantics for loosely guarded programs.  
\begin{theorem}\label{cor:gua}
Let $P$ be a program and $q$ an $n$-ary predicate in $P$. There is an answer
set $M$ of $P$ with $q(\textbf{a}) \in M$ iff $q$ is satisfiable w.r.t.
\gua{P}. Moreover, this reduction is quadratic.
\end{theorem}
\begin{theorem}\label{th:gua2}
Satisfiability checking w.r.t. LGPs is \nexptime{}-hard. 
\end{theorem}
\begin{proof}
By \cite{dantsin2001,baralbook} and the disjunction-freeness of the
GL-reduct of the programs we consider, we have that checking whether
there exists an answer set $M$ of $P$ containing a $q(\textbf{a})$ is
\nexptime{}-complete.  Thus, by Theorem \ref{cor:gua},
satisfiability checking w.r.t. a LGP is \nexptime{}-hard.
\qed
\end{proof}

A similar approach to show \nexptime{}-hardness of GPs instead of LGPs does not seem
to be directly applicable.  E.g., a naive approach is to add to the
body of every rule $r$ in a program $P$, an $n$-ary guarding atom
$g(X_1,\ldots,X_k,\ldots X_k)$, $k \leq n$, with $n$ the maximum
number of different variables in rules of $P$ and $X_1, \ldots, X_k$
the pairwise different variables in $r$.  Furthermore, one need to
enforce that for an open answer set and $n$ constants $a_1, \ldots,
a_n$, $g(a_1, \ldots, a_n)$ is in the answer set, and vice versa, if
$g(x_1,\ldots, x_n)$ is in the open answer set then $x_1, \ldots, x_n
\in \cts{P}$.  This amounts to adding $c^n$ rules
$\prule{g(a_1,\ldots, a_n)}{}$ for constants $a_1, \ldots, a_n \in
\cts{P}$ where $c$ is the number of constants in $P$.  Since $n$ is
not bounded, this transformation is, however, not polynomial.
\par
In Section \ref{sec:related}, we improve\footnote{Note that
$\p\subseteq \np \subseteq \exptime \subseteq \nexptime \subseteq
\mbox{\exptimex{2}} \subseteq  \ldots$ where $\p\subset \exptime$, $\exptime
\subset \mbox{\exptimex{2}}$, $\ldots$, and $\np\subset \nexptime$, $\nexptime
\subset \mbox{\xnexptime{2}}$, $\ldots$, see, e.g.,
\cite{papadimitriou,tobies}.}
on Theorem \ref{th:gua2} and
show that both satisfiability checking w.r.t. GPs and w.r.t. LGPs is
\exptimex{2}-hard.

\section{Open Answer Set Programming with Generalized Literals}\label{sec:answergl}

In this section, we extend the language of logic programs with
\emph{generalized literals} and modify the open answer set semantics
to accommodate for those generalized literals. As already argued in the introduction,
generalized literals allow for a more robust representation of knowledge
than is possible without them. 
\par
E.g., in \cite{BalducciniG03} a mapping is given from
an action description into an answer set program. In this mapping, a predicate $prec\_h(D,T)$
needs to be computed, intuitively meaning that all preconditions of $D$ hold at time $T$.
As the authors did not have generalized literals at their disposal, they needed a ternary relation
$pred(D,N,C)$ which encodes that $C$ is the $N$-th precondition of $D$ and a special predicate denoting
the number of preconditions for $D$, i.e. an explicit linear order among the preconditions has to be established.
Next, using this linear order, they had to introduce some additional ternary predicate $all\_h$ that checks if all 
preconditions hold and than use this predicate to compute $prec\_h(D,T)$. However, with generalized literals no linear
order needs to be established to compute $prec\_h(D,T)$, i.e. it suffices to have predicates $prec(D,C)$ encoding
that $D$ is a precondition of $C$. Than, we can use, with $h(C,T)$ meaning that condition $C$ holds at time $T$,
the rule
$$\prule{prec\_h(D,T)}{[\Forall{C}{prec(D,C)\Rightarrow h(C,T)}]}$$
to compute the correct meaning of $prec\_h$.
\par
Formally, a \emph{generalized literal} is a first-order formula of the form
\[
\Forall{\textbf{Y}}{\phi\Rightarrow\psi}\; ,
\]
where $\phi$ is a finite boolean combination of atoms 
(i.e.,  using $\neg$, $\lor$,
and $\land$) and $\psi$ is an atom;  we call
$\phi$ the \emph{antecedent} and $\psi$ the \emph{consequent}.
We refer to literals (i.e., atoms and naf-atoms since we assume the absence of $\neg$) and generalized literals as \emph{g-literals}.
For a set of 
g-literals $\alpha$, 
$\gli{\alpha} \equiv \set{l\mid l \mbox{ generalized literal in }\alpha}$, the 
set of generalized literals in $\alpha$.  We extend $\posi{\alpha}$
and $\nega{\alpha}$ for g-literals as follows: $\posi{\alpha} =
\posi{(\setmin{\alpha}{\gli{\alpha}})}$ and $\nega{\alpha} =
\nega{(\setmin{\alpha}{\gli{\alpha}})}$; thus $\alpha =
\posi{\alpha}\cup \naf{\nega{\alpha}} \cup \gli{\alpha}$.
\par
A \textit {generalized program (gP)} is a countable set of \emph{rules} 
\prule{\alpha}{\beta},
where $\alpha$ is a finite set of literals, 
$\card{\posi{\alpha}} \leq 1$,
$\beta$ is a countable\footnote{Thus the rules may have an infinite body.  }
set of g-literals,
and $\Forall{t,s}{t = s \not\in
{\posi{\alpha}}}$, i.e., $\alpha$ contains at most one positive atom,
and this atom cannot be an equality atom. 
Furthermore,
generalized literals are ground if they do not contain free
variables, and rules and gPs are ground if all g-literals
in it are ground. 
\par
For a g-literal $l$, we
define $\vars{l}$ as the (free) variables in $l$.  For a rule $r$, we define
$\vars{r} \equiv \cup\set{\vars{l} \mid l \mbox{ 
g-literal in }r}$.
For a set of atoms $I$, we extend the $\models$ relation for interpretations $I$, by induction, 
for any boolean formula of ground atoms.   For such ground boolean formulas 
$\phi$ and $\psi$, we have
\begin{enumerate}
\item $I \models \phi \land \psi$ iff $I\models \phi$ and $I\models \psi$,
\item $I \models \phi \lor \psi$ iff $I\models \phi$ or $I\models \psi$, and
\item $I \models \neg\phi$ iff $I\not\models \phi$.  
\end{enumerate}
Similarly as for programs without generalized literals, call a
pair $(U,I)$ where $U$ is a universe for $P$ and $I$ a subset of
$\HBaseU{P}{U}$ an \emph{open interpretation} of $P$. 
For a ground gP $P$ and an open interpretation $(U,I)$ of $P$, we define
the
\textit{GeLi-reduct}
$\glireduct{P}{U,I}$ which removes
the generalized literals from the
program: $\glireduct{P}{U,I}$ contains the
rules
\begin{equation}
\prule{{\alpha}}{{\setmin{\beta}{\gli{\beta}}},(\beta^{\fx})^{\fx(U,I)}
} \; ,
\end{equation}
for \prule{\alpha}{\beta} in $P$, where
\[
(\beta^{\fx})^{\fx(U,I)} \equiv 
\bigcup_{\Forall{\textbf{Y}}{\phi\Rightarrow\psi}\in\gli{\beta}}\set{\psi[\textbf{Y}|\textbf{y}] \mid
\textbf{y}\subseteq U, I\models \phi[\textbf{Y}|\textbf{y}]}\;.
\]
Intuitively, a generalized literal $\Forall{\textbf{Y}}{\phi\Rightarrow
\psi}$ is replaced by those $\psi[\textbf{Y}|\textbf{y}]$ for which $\phi[\textbf{Y}|\textbf{y}]$ is true, such that\footnote{We put square brackets around
generalized literals for clarity.},
e.g., $\prule{p(a)}{[\Forall{\textbf{X}}{q(X)\Rightarrow r(X)}]}$ means
that in order to deduce $p(a)$ one needs to deduce $r(x)$ for all $x$
where $q(x)$ holds.  If only $q(x_1)$ and $q(x_2)$ hold, then the
GeLi-reduct contains $\prule{p(a)}{r(x_1),r(x_2)}$.  With an infinite
universe and a condition $\phi$ that holds for an infinite number of
elements in the universe, one can thus have a rule with an infinite body in
the GeLi-reduct. Note that $\nega{((\beta^{\fx})^{\fx(U,I)})}$ is always
empty by definition of generalized literals: the consequent is always
an atom.
\par
Also note that $\Forall{\textbf{Y}}{\phi\Rightarrow\psi}$ cannot be seen as
$\Forall{\textbf{Y}}{\neg \phi\lor\psi}$, where the forall is an abbreviation 
of the conjunction with respect to a given domain and where we use an extended 
reduction for nested programs \cite{LifschitzTT99}. Consider e.g. the rules
\begin{pprogram}
\tsrule{p(X)}{[\Forall{Y}{\neg b(Y)\land \neg c(Y)\Rightarrow d(Y)}]}
\tsrule{b(a)}{}
\end{pprogram}
and consider the open interpretation $I=(\{a\}, \{b(a)\})$.
The GeLi reduct of $P$ w.r.t. $I$ is (note that $I \models \neg b(a)\land \neg c(a)$)
\begin{pprogram}
\tsrule{p(a)}{d(a)}
\tsrule{b(a)}{}
\end{pprogram}
which will have $I$ as an open answer set according to Definition \ref{def:oasetg} below.
However, if we apply the suggested transformation to $P$, we get
\begin{pprogram}
\tsrule{p(X)}{[\Forall{Y}{b(Y)\lor c(Y) \lor d(Y)}]}
\tsrule{b(a)}{}
\end{pprogram}
which would have the following "GL reduct for nested programs" wrt $I$:
\begin{pprogram}
\tsrule{p(a)}{b(a) \lor c(a) \lor d(a)}
\tsrule{b(a)}{}
\end{pprogram}
But, since $I \models b(a)$, the first rule becomes applicable and
thus any answer set containing $b(a)$ must also contain $p(a)$.
Hence I is not an answer set using this transformation.
\par
\begin{definition}\label{def:oasetg}
An \textit{open answer set} of $P$
is an open interpretation $(U,M)$ of $P$ where $M$ is an answer set
of ${\glireduct{(P_U)}{U,M}}$.
\end{definition}
In the following, a gP is assumed to be a finite set of finite rules;
infinite gPs only appear as byproducts of grounding a finite
program with an infinite universe, or, by taking the GeLi-reduct w.r.t.
an infinite universe. 
Satisfiability checking remains defined as before.
\begin{example}
Take a gP $P$
\begin{pprogram}
\tsrule{p(X)}{[\Forall{Y}{q(Y)\Rightarrow r(Y)}]}
\tsrule{r(X)}{q(X)}
\tsrule{q(X)\lor\naf{q(X)}}{}
\end{pprogram}
Intuitively, the first rule says that $p(X)$ holds if for every $Y$
where $q(Y)$ holds, $r(Y)$ holds (thus $p(X)$ also holds if $q(Y)$
does not hold for any $Y$).
Take an open interpretation $(\set{x,y},\set{p(x),r(x),q(x),p(y)})$.
Then, the GeLi-reduct of $P_{\set{x,y}}$ is  
\begin{pprogram}
\tsrule{p(x)}{r(x)}
\tsrule{p(y)}{r(x)}
\tsrule{r(x)}{q(x)}
\tsrule{r(y)}{q(y)}
\tsrule{q(x)\lor \naf{q(x)}}{}
\tsrule{q(y)\lor \naf{q(y)}}{}
\end{pprogram}
$\set{p(x),r(x),q(x),p(y)}$ is an answer set such that the open
interpretation
is an open answer set.
\end{example}
\begin{example}\label{ex:infiniteg}
Take the following program $P$, i.e., the open answer set variant of the
classical infinity axiom in guarded fixed point logic from \cite{gradel} (see also Example \ref{ex:infinite}, pp. \pageref{ex:infinite}),
where we use $\mathit{well}$ to denote $\mathit{well\_founded}$:
\begin{pprogramn}
\nrule{r_1}{q(X)}{f(X,Y)}
\nrule{r_2}{}{f(X,Y), \naf{q(Y)}}
\nrule{r_3}{}{f(X,Y), \naf{well(Y)}}
\nrule{r_4}{well(Y)}{q(Y), [\Forall{X}{f(X,Y)\Rightarrow well(X)}]}
\nrule{r_5}{f(X,Y)\lor\naf{f(X,Y)}}{}
\end{pprogramn}
Intuitively, in order to satisfy $q$ with some $x$, one needs to apply
$r_1$, which enforces an $f$-successor $y$.  Moreover, the second rule
ensures that also for this $y$ an $f$-successor must exist, etc.  The third
rule makes sure that every $f$-successor is on a well-founded
$f$-chain.  The well-foundedness itself is defined by $r_4$
which says that $y$ is on a well-founded chain of elements where $q$
holds
if all $f$-predecessors of $y$ satisfy the same property.
\par
E.g., take an infinite open interpretation $(U,M)$ with $U =
\set{x_0,x_1, \ldots}$ and $M =
\set{q(x_0),\lit{well(x_0)},f(x_0,x_1),q(x_1),\lit{well(x_1)},f(x_1,x_2), \ldots})$. 
$P_U$ contains the following grounding of $r_4$:
\begin{pprogramn}
\nrule{r_4^0}{well(x_0)}{q(x_0), [\Forall{X}{f(X,x_0)\Rightarrow well(X)}]}
\nrule{r_4^1}{well(x_1)}{q(x_1), [\Forall{X}{f(X,x_1)\Rightarrow well(X)}]}
\vdots \\
\end{pprogramn}
Since, for $r_4^0$, there is no $f(y,x_0)$ in $M$, the body of the
corresponding rule in 
the GeLi-reduct w.r.t. $(U,M)$ contains only $q(x_0)$.  For $r_4^1$, we have that
$f(x_0,x_1)\in M$ such that we include $\lit{well(x_0)}$ in the body:
\begin{pprogram}
\tsrule{well(x_0)}{q(x_0)}
\tsrule{well(x_1)}{q(x_1), well(x_0)}
\vdots \\
\end{pprogram}
One can check that $(U,M)$ is indeed an open answer set of the gP, satisfying $q$.
\par
Moreover, no finite open answer set can satisfy $q$.
First, note that an open answer set $(U,M)$ of $P$ cannot contain loops, i.e.,
$\set{f(x_0,x_1), \ldots, f(x_n,x_0)}\subseteq M$ is not possible.
Assume otherwise.  By rule $r_3$, we need $\lit{well(x_0)}\in M$.
However, the GeLi-reduct of $P_U$ contains rules:
\begin{pprogram}
\tsrule{well(x_0)}{q(x_0), well(x_n),\ldots}
\tsrule{well(x_n)}{q(x_n), well(x_{n-1}),\ldots}
\vdots \\
\tsrule{well(x_1)}{q(x_1), well(x_{0}),\ldots}
\end{pprogram}
such that $\lit{well(x_0)}$ cannot be in any open answer set: we have
a circular dependency and cannot use these rules to motivate
$\lit{well(x_0)}$, i.e., $\lit{well(x_0)}$ is unfounded.  Thus, an open
answer set of $P$ cannot contain loops.
\par
Assume that $q$ is satisfied in an open answer set $(U,M)$ with $q(x_0)\in
M$.  Then, by rule $r_1$, we need some $X$ such that $f(x_0,X)\in M$.
Since $M$ cannot contain loops $X$ must be different from $x_0$ and we
need some new $x_1$.  By rule $r_2$, $q(x_1)\in M$, such that by rule $r_1$,
we again need an $X$ such that $f(x_1,X)$.  Using $x_0$ or $x_1$ for
$X$ results in a loop, such that we need a new $x_2$.  This process
continues infinitely, such that there are only infinite open answer
sets that make $q$ satisfiable w.r.t. $P$.
\end{example}
We defined the open answer set semantics for gPs in function of the answer set
semantics for programs without generalized literals. We can, however,
also define a GL-reduct $P^M$ directly for a ground gP $P$ by treating 
generalized
literals as positive, such that 
$\prule{\posi{\alpha}}{\posi{\beta},\gli{\beta}} \in P^M$ iff
$\prule{\alpha}{\beta} \in P$ and $M \models \nega{\alpha}$ and
$M\models \naf{\nega{\beta}}$ for a ground gP $P$.  Applying the GL-reduct transformation after the GeLi-reduct
transformation (like we defined it), is then equivalent to first
applying the GL-reduct transformation to a gP and subsequently
computing the
GeLi-reduct.
\begin{example}
Take a program $F\cup \set{r}$ with $F\equiv\set{q(x)\gets, b(x)\gets,
b(y)\gets, c(x)\gets}$ and $r: \prule{a(X)}{[\Forall{X}{\neg
q(X)\Rightarrow b(X)}], \naf{c(X)}}$.
For a universe $U = \set{x,y}$, $(F\cup \set{r})_U$ is
$F\cup\set{r_x,r_y}$ where \[r_x: \prule{a(x)}{[\Forall{X}{\neg
q(X)\Rightarrow b(X)}], \naf{c(x)}}\] and 
\[r_y: \prule{a(y)}{[\Forall{X}{\neg
q(X)\Rightarrow b(X)}], \naf{c(y)}}\]
\par
Applying the GeLi-reduct transformation w.r.t. 
\[(U,M=\set{q(x),b(x),b(y),c(x),a(y)})\] yields
\[(F\cup\set{r_x,r_y})^{\fx{(U,M)}}\equiv F\cup
\set{\prule{a(x)}{b(y),\naf{c(x)}}; \prule{a(y)}{b(y),\naf{c(y)}}}\;
.\]
The GL-reduct of the latter is $F\cup
\set{\prule{a(y)}{b(y)}}$, such that $(U,M)$ is a (unique)
open answer set of $F\cup \set{r}$ for $U= \set{x,y}$.
\par
First applying the GL-reduct transformation to $F\cup \set{r_x,r_y}$
yields $F\cup \set{r_y}$,
and, subsequently, the GeLi-reduct again gives
$F\cup \set{\prule{a(y)}{b(y)}}$.  Thus
\[((F\cup\set{r_x,r_y})^{\fx{(U,M)}})^M =
((F\cup\set{r_x,r_y})^M)^{\fx{(U,M)}}\; .
\]
\end{example}
Since the GeLi-reduct transformation never removes rules or naf-atoms
from rules, while the GL-reduct transformation may remove rules (and
thus generalized literals), calculating the GL-reduct before the
GeLi-reduct is likely to be more efficient in practice. We opted,
however, for the ``GeLi-reduct before GL-reduct" transformation as the
standard definition, as it is theoretically more robust against
changes in the definition of generalized literals.  E.g., if naf were
allowed in the consequent of generalized literals, the ``GL-reduct
before GeLi-reduct" approach does not work since the
GeLi-reduct (as currently defined) could introduce naf again in the
program,
making another application of the GL-reduct transformation necessary.
\begin{theorem}\label{prop:gliaftergl}
Let $P$ be a ground gP with an open interpretation $(U,M)$.  Then, 
\[
(P^{\fx{(U,M)}})^M = (P^M)^{\fx{(U,M)}}\; .
\] 
\end{theorem}
%
%
%

We have a similar result as in Theorem \ref{th:finitelyminimal} regarding the finite motivation of
literals in possibly infinite open answer sets.
We again express the motivation of a literal more formally by means of 
the \textit{immediate consequence operator} \cite{emden76} $T$ that
computes the closure of a set of literals w.r.t. a GL-reduct of a
GeLi-reduct.
\par
For a gP $P$ and an open interpretation $(U,M)$ of $P$, $T_P^{(U,M)}:
\HBaseU{P}{U}\to \HBaseU{P}{U}$ is defined as $T(B) = B \cup \set{a
\vert a \gets \beta \in \left(P_U^{\fx{(U,M)}}\right)^M \land B\models \beta 
}$.  Additionally, we have  $T^0(B) = B$\footnote{ We omit the
sub- and superscripts $(U,M)$ and $P$ from $T_P^{(U,M)}$ if they are clear from the context and, furthermore, we will
usually write $T$ instead of $T(\emptyset)$.}, and $T^{n+1}(B) =
T(T^{n}(B))$.
\begin{theorem}\label{th:finitelyminimalgl} 
Let $P$ be a gP and
$(U,M)$ an open answer set of $P$. Then, $\Forall{a \in M}{\Exists{n
< \infty}{a \in T^{n}}}$.  
\end{theorem}
%

Finally, the next example illustrates that their is a difference between our answer set semantics for
generalized literals and the answer set semantics introduced in \cite{LifschitzPV01,OsorioNA04,osorio_ortis} for propositional theories, 
which is based on intuistionistic logic.
\begin{example}
Consider the program
\begin{pprogram}
\tsrule{a(X)}{[\Forall{X}{c(X)\Rightarrow b(X)}]}
\tsrule{a(X)}{b(X)}
\tsrule{b(X)}{c(X)}
\tsrule{c(X)}{a(X)}
\end{pprogram}
and consider the open interpretation $I=(\{a\},\{a(a),b(a),c(a)\})$. Applying the GeLi-reduct on this program
w.r.t. $I$ yields the program
\begin{pprogram}
\tsrule{a(a)}{b(a)}
\tsrule{a(a)}{b(a)}
\tsrule{b(a)}{c(a)}
\tsrule{c(a)}{a(a)}
\end{pprogram}
which only has $\emptyset$ as an answer set, implying that $I$ is not an open answer set for this program.
\par
However, one could expect the programs
\begin{pprogram}
\tsrule{a(a)}{[\Forall{X}{c(a) \Rightarrow b(a)}]}
\tsrule{a(a)}{b(a)}
\tsrule{b(a)}{c(a)}
\tsrule{c(a)}{a(a)}
\end{pprogram}
and
\begin{pprogram}
\tsrule{a(a)}{c(a) \Rightarrow b(a)}
\tsrule{a(a)}{b(a)}
\tsrule{b(a)}{c(a)}
\tsrule{c(a)}{a(a)}
\end{pprogram}
to be equivalent, but, in the context of \cite{LifschitzPV01,OsorioNA04,osorio_ortis}, we have $\{a(a),b(a),c(a)\}$ as the unique answer set
for the last program, as $c(a) \Rightarrow b(a)$ is true because of the third rule in that program. Thus, this example 
illustrates that there is a difference between our semantics and the one in \cite{LifschitzPV01,OsorioNA04,osorio_ortis}.
\end{example}

In \cite{leone}, so-called \emph{parametric connectives} are introduced in the
context of disjunctive logic programs.  The semantics of parametric
connectives, e.g, $\bigwedge\{p(X) : a(X,Y), b(X)\}$, is essentially the same
as the semantics of generalized literals $\Forall{X}{a(X,Y) \land b(X)\Rightarrow p(X)}$.
Note that \cite{leone} also allows for a disjunction in the body (indicated by
a $\bigvee$ instead of $\bigwedge$), however, since we allow for arbitrary
boolean formulas in the antecedent of our generalized literals, the latter are
more flexible.

\section{Open Answer Set Programming with gPs via Fixed Point Logic}\label{sec:fplgl}

We reduce satisfiability checking w.r.t. gPs to satisfiability
checking of FPL formulas.  Note that the exposition in this section is
along the lines of Section \ref{sec:fpl}, such that we will skip the
details of some of the proofs.
\par
First, we rewrite
an arbitrary gP as a gP containing only one designated
predicate $p$ and (in)equality.
A gP $P$ is a \emph{$p$-gP} if $p$ is the only predicate in $P$
different from the (in)equality predicate.  
For a set of g-literals $\alpha$, we construct
$\pprog{\alpha}{p}$ in two stages:
\begin{enumerate}
\item replace every regular $m$-ary atom $q(\textbf{t})$ appearing in $\alpha$
(either in atoms, naf-atoms, or generalized literals) by
$p(\textbf{t},\textbf{0},q)$ where $p$ has arity $n$, with $n$ the maximum
of the arities of predicates in $P$ augmented by $1$, $\textbf{0}$ a
sequence of new constants $0$ of length $n-m-1$, and $q$ a new
constant with the same name as the original predicate,
\item in the set thus obtained, replace every generalized literal
$\Forall{\textbf{Y}}{\phi\Rightarrow\psi}$ by $\Forall{\textbf{Y}}{\phi\land
\bigwedge \lit{in(\textbf{Y})}\Rightarrow\psi}$, where $Y\neq t$ in \lit{in(\textbf{Y})} stands for
$\neg (Y =  t)$ (we defined generalized literals in function of
boolean formulas of atoms).
\end{enumerate}
The $p$-gP $P_p$ is then the program $P$ with all non-free rules
$r:\prule{\alpha}{\beta}$ replaced by
$r_p:\prule{\alpha_p}{\beta_p,\lit{in(\textbf{X})}}$
 where $\vars{r} = \lit{\textbf{X}}$. Note that $P$ and
 $P_p$ have the same free rules.
\begin{example}\label{ex:p-prog}
Let $P$ be the gP:
\begin{pprogram}
\tsrule{q(X)}{[\Forall{Y}{r(Y)\Rightarrow s(X)}]}
\tsrule{r(a)}{}
\tsrule{s(X)\lor\naf{s(X)}}{}
\end{pprogram}
Then $q$ is satisfiable by an open answer set
$(\set{a,x},\set{s(x),r(a),q(x)})$.  The $p$-gP $P_p$ is 
\begin{pprogram}
\tsrule{p(X,q)}{[\Forall{Y\!}{p(Y,r)\land \bigwedge \lit{in(Y)}\!\Rightarrow p(X,s)}],\lit{in(X)}}
\tsrule{p(a,r)}{}
\tsrule{p(X,s)\lor\naf{p(X,s)}}{}
\end{pprogram}
where $\lit{in(X)} = \set{X\neq s, X \neq q, X\neq r, X \neq 0}$.
The corresponding open answer set for this program is
$(\set{a,x,s,r,q},\set{p(x,s),p(a,r),p(x,q)})$.  
\end{example}
\begin{theorem}\label{prop:p-programg}
Let $P$ be a gP, $p$ a predicate not in $P$, and $q$ a predicate in $P$. $q$ is satisfiable w.r.t.
$P$ iff there is an open answer set $(U',M')$ of the $p$-gP $P_p$ with
$p(\textbf{x},\textbf{0},q)\in M'$. Furthermore, the size of $P_p$ is
polynomial in the size of $P$.
\end{theorem}
\begin{proof}
The proof is analogous to the proof of Theorem \ref{prop:p-program}.
\qed
\end{proof}
The
\emph{completion} \Compg{P} of a gP $P$ consists of 
formulas 
that demand that different constants in $P$ are interpreted as
different elements:
\begin{equation}\label{eq:constantsg}
a \neq b \enspace .
\end{equation}
For every pair of different constants $a$ and $b$ in $P$,
\Compg{P} contains formulas ensuring the existence of at least
one element in the domain of an interpretation:
\begin{equation}\label{eq:oneelemg}
\Exists{X}{\true} \enspace .
\end{equation}
Besides these technical requirements matching FOL interpretations with
open interpretations, \Compg{P} contains the formulas in $\fix{P} =
\sat{P} \cup \gl{P} \cup \glit{P} \cup \fpf{P}$, which can be intuitively
categorized as follows:
\begin{itemize}
\item \sat{P}
ensures that a model of \fix{P} satisfies all rules in $P$, 
\item \gl{P} is an auxiliary component defining atoms that indicate when a rule in $P$
belongs to the GL-reduct,
\item $\glit{P}$ indicates when the
antecedents of generalized literals are true, and 
\item $\fpf{P}$ ensures that
every model of \fix{P} is a minimal model of the GL-reduct of the
GeLi-reduct of $P$; it
uses the atoms defined in \gl{P} to select, for the calculation of the
fixed point, only those rules in $P$ that are in the GL-reduct of the
GeLi-reduct of $P$; the atoms defined in $\glit{P}$ ensure that the
generalized literals are interpreted correctly.
\end{itemize}
In the following, we assume that the arity of $p$,
the only predicate in a $p$-gP is $n$.
\begin{definition}\label{def:compg}
Let $P$ be a $p$-gP.  The fixed point translation of $P$ is\linebreak
$\fix{P} \equiv \sat{P}\cup \glit{P} \cup \gl{P} \cup \fpf{P}$, where
\begin{enumerate}

\item $\sat{P}$ contains formulas
\begin{equation}\label{eq:satg}
\Forall{\textbf{Y}}{\bigwedge \beta \Rightarrow \bigvee \alpha}
\end{equation}
for rules $r:\prule{\alpha}{\beta} \in P$ with $\vars{r} = \textbf{Y}$,

\item $\gl{P}$ contains the formulas
\begin{equation}\label{eq:glg}
\Forall{\textbf{Y}}{\equiva{ r(\textbf{Y}) }{ \bigwedge \nega{\alpha} \land \bigwedge \neg \nega{\beta} }}
\end{equation}
for rules
$r: \prule{\alpha}{\beta} \in P$ with $\vars{r} = \textbf{Y}$,

\item $\glit{P}$ contains the formulas
\begin{equation}\label{eq:glit}
\Forall{\textbf{Z}}{\equiva{ g(\textbf{Z}) }{ \phi }}
\end{equation}
for generalized literals 
$g: \Forall{\textbf{Y}}{\phi\Rightarrow\psi} \in P$\footnote{We assume
that generalized literals are named.} where $\phi$ contains the
variables $\textbf{Z}$, 

\item $\fpf{P}$ contains the formula
\begin{equation}\label{eq:fpfg}
\Forall{\textbf{X}}{\impl{p(\textbf{X})}{\lfp{W}{\textbf{X}}{\phi(W,\textbf{X})}(\textbf{X})}}
\end{equation}
with
\begin{equation}\label{eq:fpformulag}
\phi(W,\textbf{X}) \equiv W(\textbf{X}) \lor \bigvee_{r:\prule{p(\textbf{t}) \lor \alpha}{\beta} \in  P}E(r) 
\end{equation}
and
\begin{equation}\label{eq:Erg}
E(r) \equiv \Exists{\textbf{Y}}{X_1 = t_1 \land \ldots \land X_n =
t_n \land \bigwedge \posi{\beta}[p\mid W]\land \bigwedge \gamma \land
r(\textbf{Y})}
\end{equation}
where $\textbf{X} = X_1, \ldots, X_n$ are $n$ new variables, $\vars{r} = \textbf{Y}$,
 $W$ is a new (second-order) variable, $\posi{\beta}[p\mid W]$ is $\posi{\beta}$ with $p$ replaced by $W$,
 and $\gamma$ is $\gli{\beta}$ 
with
\begin{itemize}
\item every generalized literal $g:\Forall{\textbf{Y}}{\phi\Rightarrow\psi}$ replaced by 
$\Forall{\textbf{Y}}{g(\textbf{Z})\Rightarrow\psi}$, $\textbf{Z}$ the
variables of $\phi$, and, subsequently,
\item every $p$ replaced by $W$.
\end{itemize}
\end{enumerate}
The \emph{completion} of $P$ is $\Compg{P} \equiv \fix{P} \cup
\set{(\ref{eq:constantsg}),(\ref{eq:oneelemg})}$.
\end{definition}
The predicate $W$ appears only positively in $\phi(W,\textbf{X})$ such that
the fixed point formula in $(\ref{eq:fpfg})$ is well-defined.  Note
that the predicate $p$ is replaced by the fixed point variable $W$ in
$E(r)$ except in the antecedents of generalized literals, which were
replaced by atoms $g(\textbf{Z})$, and the
negative part of $r$, which were replaced by atoms $r(\textbf{Y})$, thus
respectively encoding the GeLi-reduct and the GL-reduct.\footnote{
Note that we apply the GeLi-reduct and the GL-reduct ``at the same
time", while the open answer set semantics is defined such that first the GeLi-reduct
is constructed and then the GL-reduct.  However, as indicated by
Theorem \ref{prop:gliaftergl}, the order of applying the reducts does
not matter.}
\par
By the first disjunct in $(\ref{eq:fpformulag})$, we have that applying
$\operphi{(U,M)}$ to a set $S \subseteq U^n$ does not lose
information from $S$.
\begin{theorem}\label{prop:Sg}
Let $P$ be a $p$-gP and $(U,M)$ an open interpretation 
with $S \subseteq U^n$. Then 
\[
S \subseteq \operphi{(U,M)}(S) \; .
\]
\end{theorem}
\begin{proof}
Similar to the proof of Theorem \ref{prop:S}.
\qed
\end{proof}
\begin{example}
We rewrite the program from Example \ref{ex:infiniteg} as the
$p$-gP $P$:
{\small
\begin{pprogramn}
\nrule{r_1}{p(X,0,q)}{p(X,Y,f), in(X), in(Y)}
\nrule{r_2}{}{p(X,Y,f), \naf{p(Y,0,q)}, in(X), in(Y)}
\nrule{r_3}{}{p(X,Y,f), \naf{p(Y,0,well), in(X), in(Y)}}
\nrule{r_4}{p(Y,0,well)}{p(Y,0,q), in(Y), }
& & & [\Forall{X}{p(X,Y,f)\land \bigwedge in(X)\Rightarrow p(X,0,well)}] \\
\nrule{r_5}{p(X,Y,f)\lor\naf{p(X,Y,f)}}{}
\end{pprogramn}
}
where $\lit{in(X)}$ and $\lit{in(Y)}$ are shorthand for the
inequalities with the new constants.
$\sat{P}$ consists of the sentences
\begin{itemize}
\item $\Forall{X,Y}{p(X,Y,f)\land \bigwedge \lit{in}(X)\land \bigwedge \lit{in}(Y)\Rightarrow p(X,0,q)}$, 
\item $\Forall{X,Y}{p(X,Y,f)\land \neg p(Y,0,q) \land \bigwedge \lit{in}(X) \land \bigwedge \lit{in}(Y) \Rightarrow \false}$,  
\item $\Forall{X,Y}{p(X,Y,f)\land \neg p(Y,0,\lit{well}) \land \bigwedge \lit{in}(X) \land \bigwedge \lit{in}(Y) \Rightarrow \false}$, 
\item $\Forall{Y}{p(Y,0,q)\land \bigwedge \lit{in}(Y)\land \left(\Forall{X}{p(X,Y,f)\land \bigwedge \lit{in}(X)\Rightarrow p(X,0,\lit{well})\right)}}$ \\
	$\Rightarrow p(Y,0,\lit{well})$, and 
\item $\Forall{X,Y}{\true \Rightarrow p(X,Y,f)\lor \neg p(X,Y,f)}$.
\end{itemize}
$\gl{P}$ contains the sentences
\begin{itemize}
\item $\Forall{X,Y}{\equiva{ r_1(X,Y) }{ \bigwedge\lit{in}(X)\land \bigwedge\lit{in}(Y)}}$,
\item $\Forall{X,Y}{\equiva{ r_2(X,Y) }{ \neg p(Y,0,q) \land \bigwedge\lit{in}(X)\land \bigwedge\lit{in}(Y)}}$,
\item $\Forall{X,Y}{\equiva{ r_3(X,Y) }{ \neg p(Y,0,\lit{well}) \land \bigwedge\lit{in}(X)\land \bigwedge\lit{in}(Y)}}$, 
\item $\Forall{Y}{\equiva{ r_4(Y) }{ \bigwedge\lit{in}(Y)}}$, and
\item $\Forall{X,Y}{\equiva{r_5(X,Y)}{p(X,Y,f)}}$.
\end{itemize}
$\glit{P}$ contains the sentence $\Forall{X,Y}{\equiva{ g(X,Y) }{ p(X,Y,f)\land \bigwedge \lit{in}(X) }}$, and $\fpf{P}$ is
constructed with 
\begin{itemize}
\item $E(r_1) \equiv \Exists{X,Y}{X_1 = X \land X_2 = 0 \land X_3 = q
\land W(X,Y,f) \land r_1(X,Y)}$,
\item $E(r_4)\equiv \Exists{Y}{X_1 = Y \land X_2 = 0 \land X_3 = \mathit{well}  \land W(Y,0,q) \land }$\\
$\left( \Forall{X}{g(X,Y)\Rightarrow W(X,0,\lit{well})}\right) \land r_4(Y)$, and
\item $E(r_5)\equiv \Exists{X,Y}{X_1 = X \land X_2 = Y \land X_3 = f  \land r_5(X,Y)}$.
\end{itemize}
Take an infinite FOL interpretation $(U,M)$ with $U =
\set{q,f,\lit{well},0,x_0,x_1, \ldots}$ and\footnote{We interpret the constants in
\Compg{P} by universe elements of the same name.}
\begin{multline*}
M =
\set{p(x_0,0,q),\lit{p(x_0,0,well)},p(x_0,x_1,f),\\ p(x_1,0,q),\lit{p(x_1,0,well)},p(x_1,x_2,f), \ldots\\
r_1(x_0,x_0), r_1(x_0,x_1), \ldots,r_1(x_1,x_0), \ldots, r_4(x_0),
r_4(x_1), \ldots \\
r_5(x_0,x_1), r_5(x_1,x_2),\ldots,
g(x_0,x_1), g(x_1,x_2),\ldots 
})\; .
\end{multline*}
$\sat{P}, \gl{P}$, and $\glit{P}$ are satisfied.  We check that
$\fpf{P}$ is satisfied by $M$.  We construct the fixed point of
$\phi^{(U,M)}$ where $\phi(W,X_1,X_2,X_3) \equiv W(X_1,X_2,X_3) \lor E(r_1) \lor E(r_4)
\lor E(R_5)$ as in \cite{Gr02}, i.e., in stages starting from $W^0 =
\emptyset$.  We have that 
\begin{itemize}
\item $W^1= \phi^{(U,M)}(W^0) = \set{(x_0,x_1,f),
(x_1,x_2,f), \ldots}$, where the $(x_i,x_{i+1},f)$
are introduced by $E(r_5)$,
\item $W^2= \phi^{(U,M)}(W^1) = W^1 \cup \set{(x_0,0,q), (x_1, 0,q),
\ldots}$, where the $(x_i,0,q)$
are introduced by $E(r_1)$,
\item $W^3= \phi^{(U,M)}(W^2) = W^2 \cup \set{(x_0,0,\lit{well})}$,
where $(x_0,0,\lit{well})$ is introduced by $E(r_4)$,
\item $W^4= \phi^{(U,M)}(W^3) = W^3 \cup \set{(x_1,0,\lit{well})}$,
\item \ldots
\end{itemize}
The least fixed point $\lfpoint{\phi^{(U,M)}}$ is then $\cup_{\alpha <
\infty}W^{\alpha}$ \cite{Gr02}. The sentence $\fpf{P}$ is then satisfied since
every $p$-literal in $M$ is also in this least fixed point.  $(U,M)$ is
thus a model of \Compg{P}, and it corresponds to an open answer
set of $P$.
\end{example}
\begin{theorem}\label{th:aset-iff-modelg}
Let $P$ be a $p$-gP. Then, $(U,M)$ is an open answer set of $P$ iff
$(U,M \cup R \cup G)$ is a model of $\bigwedge \Compg{P}$, where 
\[R \equiv \set{r(\textbf{y}) \mid r[\textbf{Y}\mid
\textbf{y}]:\prule{\alpha[]}{\beta[]} \in P_U, M\models \nega{\alpha[]} \cup
\naf{\nega{\beta[]}},
\vars{r} = \textbf{Y}}\; ,\] 
i.e., the atoms corresponding to rules for which the GeLi-reduct
version will be in the GL-reduct, and 
\[
G \equiv \set{g(\textbf{z}) \mid
g:\Forall{\textbf{Y}}{\phi\Rightarrow \psi}\in P, \vars{\phi}=\textbf{Z},
M\models \phi[\textbf{Z}\mid \textbf{z}] }\; ,\]
i.e., the atoms corresponding to
true antecedents of generalized literals in $P$.
\end{theorem}
\begin{proof} 
Similar to the proof of Theorem
\ref{th:aset-iff-model}.
\qed
\end{proof}
Using Theorems \ref{prop:p-programg} and
\ref{th:aset-iff-modelg}, we can reduce satisfiability checking w.r.t.
gPs
to satisfiability checking in FPL. Moreover, since $\bigwedge\Compg{P}$ contains
only one fixed point predicate, the translation falls in the
alternation-free fragment of FPL.  
\begin{theorem}\label{th:sat1g}
Let $P$ be a gP, $p$ a predicate not appearing in $P$, and $q$ an $n$-ary predicate in $P$. $q$ is satisfiable
w.r.t. $P$ iff $\Exists{\textbf{X}}{p(\textbf{X},\textbf{0},q) \land \bigwedge \Compg{P_p}}$ is satisfiable.
Moreover, this reduction is polynomial.
\end{theorem}
\begin{proof}
Assume $q$ is satisfiable w.r.t. $P$.  By Theorem 
\ref{prop:p-programg}, we have that $p(\textbf{x},\textbf{0},q)$ is in an
open answer set of \pprog{P}{p}, such that with Theorem
\ref{th:aset-iff-modelg}, $p(\textbf{x},\textbf{0},q)$ is in a model of
$\bigwedge\Compg{P_p}$.
\par
For the opposite direction, assume $\Exists{\textbf{X}}{p(\textbf{X},\textbf{0},q)\land
\bigwedge\Compg{P_p}}$ is satisfiable.  Then there is a model $(U,M')$ of
$\bigwedge\Compg{P}$ with $p(\textbf{x},\textbf{0},q) \in M'$. We have that $M' = M \cup
R\cup G$ as in Theorem \ref{th:aset-iff-modelg}, such that $(U,M)$ is an open
answer set of $P_p$ and $p(\textbf{x},\textbf{0},q) \in M$.  From
Theorem \ref{prop:p-programg}, we then have an open answer set of
$P$ satisfying $q$.
\par
The size of $\bigwedge \Compg{\pprog{P}{p}}$ is polynomial in the size of
\pprog{P}{p}.  Since the size of the latter is also polynomial in the
size of $P$, the size of $\bigwedge \Compg{\pprog{P}{p}}$ is polynomial in the
size of $P$.
\qed
\end{proof}

\section{Open Answer Set Programming with Guarded Generalized Programs}\label{sec:goaspg}

As we did in Section \ref{sec:goasp} for programs, we introduce in this
section a notion of guardedness such that the FPL translation of \emph{guarded gPs} falls in
$\mugf$.  We do not, however, consider their \emph{loosely guarded} counterpart
like we did in Section \ref{sec:goasp}, but leave this as an exercise
to the reader.
\begin{definition}\label{def:guardg}
A generalized literal $\Forall{\textbf{Y}}{\phi\Rightarrow\psi}$ is \emph{guarded} if $\phi$ is
of the form $\gamma \land \phi'$ with $\gamma$
an atom, and $\vars{\textbf{Y}}\cup\vars{\phi'}\cup\vars{\psi}\subseteq \vars{\gamma}$; we call $\gamma$ the
\emph{guard} of the generalized literal.
A rule $r:\prule{\alpha}{\beta}$ is \emph{guarded} if every
generalized literal in $r$ is guarded, and there is an atom $\gamma_b
\in {\posi{\beta}}$ such that $\vars{r}\subseteq \vars{\gamma_b}$; 
we call $\gamma_b$ a \emph{body guard} of $r$.  It is
\emph{fully guarded} if it is guarded and there is a $\gamma_h
\subseteq {\nega{\alpha}}$ such that $\vars{r}\subseteq \vars{\gamma_h}$; 
$\gamma_h$  is called a \emph{head guard} of $r$. 
\par
A gP $P$ is a \emph{(fully) guarded gP ((F)GgP)} if every
non-free rule in $P$ is {(fully) guarded}. 
\end{definition}
\begin{example}
Reconsider the gP from Example \ref{ex:infiniteg}.   $r_1, r_2$, and
$r_3$ are guarded with guard $f(X,Y)$.  The generalized literal in
$r_4$ is guarded by $f(X,Y)$, and $r_4$ itself is guarded by $q(Y)$.
Note that $r_5$ does not influence the guardedness as it is a free
rule. 
\end{example}
Every fully guarded gP is guarded.  Vice versa, we can transform every
guarded gP into an equivalent fully guarded one.
For a GgP $P$, 
$\hbg{P}$ is defined as in Section \ref{sec:goasp} (pp.
\pageref{sec:goaspg}), i.e., as $P$ with the rules $\prule{\alpha}{\beta}$
replaced by $\prule{\alpha\cup \naf{\posi{\beta}}}{\beta}$ for the body
guard $\gamma_b$ of $\prule{\alpha}{\beta}$.
For a GgP
$P$, we have that \hbg{P} is a FGgP, where the head guard of each
non-free rule is equal to the body guard.
Moreover, the size of \hbg{P} is linear in the size of $P$.
\begin{theorem}\label{prop:hbgg}
Let $P$ be a GgP.
An open interpretation $(U,M)$ of $P$ is an open answer set of $P$ iff
$(U,M)$ is an open answer set of $\hbg{P}$.
\end{theorem}
\begin{proof}
The proof is analogous to the proof of Theorem \ref{prop:hbg} (pp.
\pageref{prop:hbg}).
\qed
\end{proof}
We have that the construction of a $p$-gP retains the 
guardedness properties.
\begin{theorem}\label{prop:hbg3g}
Let $P$ be a gP. Then, $P$ is a (F)GgP iff $\pprog{P}{p}$ is a
(F)GgP. 
\end{theorem}
\begin{proof}
The proof is analogous to the proof of Theorem \ref{prop:hbg3} (pp.
\pageref{prop:hbg3}).
\qed
\end{proof}
For a fully guarded $p$-gP $P$, we can rewrite \Compg{P} as the
equivalent \mugf{} formulas \GCompg{P}.
For a guarded generalized literal $\xi \equiv \Forall{\textbf{Y}}{\phi\Rightarrow \psi}$, define 
\[
\xi^{g}\equiv
\Forall{\textbf{Y}}{\gamma\Rightarrow \psi\lor \neg \phi'}\; ,
\] where, since the
generalized literal is guarded, $\phi = \gamma \land \phi'$, and
$\vars{\textbf{Y}}\cup \vars{\phi'} \cup \vars{\psi}\subseteq
\vars{\gamma}$, making formula
$\xi^g$ a guarded formula.
The extension of this operator $\cdot^{g}$ for sets (or boolean formulas) of
generalized literals is as usual.
\par 
\GCompg{P} is \Compg{P}
with the following modifications.
\begin{itemize}
\item Formula $\Exists{X}{\true}$ is replaced
by\begin{equation}\label{eq:oneelem1g}\Exists{X}{X = X}\;
,\end{equation}
such that it is guarded by $X = X$.

\item Formula $(\ref{eq:satg})$ is removed if $r:\prule{\alpha}{\beta}$ is free or otherwise replaced by
\begin{equation}\label{eq:sat2g}
\Forall{\textbf{Y}}{\gamma_b \Rightarrow \bigvee \alpha \lor
\bigvee \neg (\setmin{\posi{\beta}}{\set{\gamma_b}})} \lor \bigvee
\nega{\beta}\lor \bigvee
\neg{(\gli{\beta})^g}\; ,
\end{equation}
where $\gamma_b$ is a body guard of $r$, thus we have
logically rewritten the formula such that it is guarded.
 If
$r$ is a free rule of the form
\prule{q(\textbf{t})\lor\naf{q(\textbf{t})}}{} we have
$\Forall{Y}{\impl{\true}{q(\textbf{t})\lor \neg q(\textbf{t})}}$ which is
always true and can thus be removed from \Compg{P}.  

\item Formula $(\ref{eq:glg})$ is replaced by the formulas
\begin{equation}\label{eq:gl2g}
\Forall{\textbf{Y}}{\impl{ r(\textbf{Y}) }{ \bigwedge \nega{\alpha} \land \bigwedge \neg \nega{\beta} }}
\end{equation}
and
\begin{equation}\label{eq:gl3g}
\Forall{\textbf{Y}}{\impl{ \gamma_h }{ r(\textbf{Y}) \lor  \bigvee \nega{\beta} \lor \bigvee \neg (\setmin{\nega{\alpha}}{\set{\gamma_h}})  }}\; ,
\end{equation}
where $\gamma_h$ is a head guard of \prule{\alpha}{\beta}. 
We thus rewrite an
equivalence as two implications where the first implication is guarded
by $r(\textbf{Y})$ and the second one is guarded by the head guard of the
rule.

\item Formula $(\ref{eq:glit})$ is replaced by the formulas
\begin{equation}\label{eq:glit2g}
\Forall{\textbf{Z}}{\impl{ g(\textbf{Z})  }{\phi}}
\end{equation}
and 
\begin{equation}\label{eq:glit3g}
\Forall{\textbf{Z}}{\impl{\gamma  }{g(\textbf{Z}) \lor \neg \phi'}}
\end{equation}
where $\phi = \gamma \land \psi$ by the guardedness of the generalized
literal $\Forall{\textbf{Y}}{\impl{\phi}{\psi}}$. We thus rewrite an
equivalence as two implications where the first one is guarded by
$g(\textbf{Z})$ ($\vars{\phi} = \textbf{Z}$ by definition of $g$), and
the second one is guarded by $\gamma$ ($\vars{g(\textbf{Z}) \lor \neg \phi'} = \vars{\textbf{Z}} = \vars{\gamma}$).

\item For every $E(r)$ in $(\ref{eq:fpfg})$, replace $E(r)$ by 
\begin{equation}\label{eq:Er2g}
E'(r) \equiv \bigwedge_{t_i \not\in \textbf{Y}} X_i = t_i \land \Exists{\textbf{Z}}{(\bigwedge \posi{\beta}[p\vert W] \land \bigwedge \gamma  \land
r(\textbf{Y}))[t_i \in \textbf{Y} \vert X_i]}\; ,
\end{equation}
with $\textbf{Z} = \setmin{\textbf{Y}}{\set{t_i \mid t_i\in \textbf{Y}}}$, i.e., move all $X_i = t_i$ where $t_i$ is
constant out of the scope of the quantifier, and remove the others by
substituting each $t_i$ in $\bigwedge \posi{\beta}[p\vert W] \land
\bigwedge \gamma \land
r(\textbf{Y})$ by $X_i$. This rewriting makes sure that every (free) variable in
the quantified part of $E'(R)$ is guarded by $r(\textbf{Y})[t_i\in
\textbf{Y}\vert X_i]$.

\end{itemize}
\begin{example}
The rule \[
r:\prule{p(X)\lor\naf{p(X)}}{p(X),[\Forall{Y}{p(Y)\land p(b)}\Rightarrow{p(a)}]}
\]
constitutes a
fully guarded $p$-gP $P$.
The generalized literal is guarded by $p(Y)$ and the rule by head and
body guard $p(X)$. $\sat{P}$ contains the formula $\Forall{X}{p(X) \land (\Forall{Y}{p(Y)\land p(b)\Rightarrow p(a)}) \Rightarrow p(X)\lor\neg p(X)}$, 
$\gl{P}$
consists of $\Forall{X}{\equiva{r(X)}{p(X)}}$, $\glit{P}$ is the
formula $\Forall{Y}{\equiva{g(Y)}{p(Y)\land p(b)}}$ and $E(r) \equiv
\Exists{X}{X_1 = X \land W(X) \land (\Forall{Y}{g(Y)\Rightarrow W(a)})\land r(X)}$.
\par
\GCompg{P} consists then of the corresponding guarded formulas:
\begin{itemize}
\item $\Forall{X}{p(X) \Rightarrow p(X)\lor\neg p(X)\lor \neg(\Forall{Y}{p(Y)\Rightarrow p(a)\lor \neg p(b)})}$, 
\item $\Forall{X}{\impl{r(X)}{p(X)}}$,
\item $\Forall{X}{\impl{p(X)}{r(X)}}$,
\item $\Forall{Y}{\impl{g(Y)}{p(Y)\land p(b)}}$,
\item $\Forall{Y}{\impl{p(Y)}{g(Y)\lor \neg p(b)}}$, and
\item $E'(r) \equiv
W(X_1) \land (\Forall{Y}{g(Y)\Rightarrow W(a)})\land r(X_1)$.
\end{itemize}
\end{example}
As $\GCompg{P}$ is basically a linear logical rewriting of $\Compg{P}$, they
are equivalent.  Moreover, $\bigwedge \GCompg{P}$ is an
alternation-free $\mugf{}$ formula.
\begin{theorem}\label{prop:gcompg}
Let $P$ be a fully guarded $p$-gP.  $(U,M)$ is a model of\linebreak
$\bigwedge\Compg{P}$ iff $(U,M)$ is a model of $\bigwedge\GCompg{P}$.
\end{theorem}
\begin{proof}
The only notable difference from the proof of Theorem \ref{prop:gcomp}
is the presence of generalized literals, which are handled by the
observation that 
$(U,M)\models \xi \iff (U,M)\models \xi^g$ for a generalized
literal $\xi$.
\qed
\end{proof}
\begin{theorem}\label{th:gcompg}
Let $P$ be a fully guarded $p$-gP. $\bigwedge \GCompg{P}$ is an
alternation-free \mugf{} formula.
\end{theorem}
\begin{proof}The proof is analogous to the proof of Theorem \ref{th:gcomp}.
\qed
\end{proof}
\begin{theorem}\label{th:sat2g}
Let $P$ be a GgP and $q$ an $n$-ary predicate in $P$. $q$ is satisfiable
w.r.t. $P$ iff $\Exists{\textbf{X}}{p(\textbf{X},\textbf{0},q) \land \bigwedge \GCompg{\pprog{(\hbg{P})}{p}}}$ is satisfiable. 
Moreover, this reduction is polynomial.
\end{theorem}
\begin{proof}
We have that $\hbg{P}$ is a FGgP.  By Theorem \ref{prop:hbg3g}, we have that
\pprog{(\hbg{P})}{p} is a fully guarded $p$-gP, thus the formula
$\bigwedge\GCompg{\pprog{(\hbg{P})}{p}}$ is defined. By Theorem
\ref{prop:hbgg}, we have that $q$ is satisfiable w.r.t. $P$ iff $q$ is
satisfiable w.r.t. \hbg{P}. By Theorem \ref{th:sat1g}, we have that
$q$ is satisfiable w.r.t.  \hbg{P} iff $\Exists{\textbf{X}}{p(\textbf{X},\textbf{0},q) \land
\bigwedge \Compg{\pprog{(\hbg{P})}{p}}}$ is satisfiable. Finally, Theorem
\ref{prop:gcompg} yields that $q$ is satisfiable w.r.t. $P$ iff
$\Exists{\textbf{X}}{p(\textbf{X},\textbf{0},q) \land \bigwedge \GCompg{\pprog{(\hbg{P})}{p}}}$ is
satisfiable. 
\qed
\end{proof}
\begin{corollary}\label{th:dec1g}
Satisfiability checking w.r.t. GgPs can be polynomially reduced to satisfiability
checking of alternation-free $\mugf$-formulas.
\end{corollary}
\begin{proof}
For a GgP $P$, we have, by Theorem
\ref{th:gcompg}, that $\bigwedge \GComp{\pprog{(\hbg{P})}{p}}$ is an
alternation-free \mugf{}, which yields with Theorem \ref{th:sat2g},
the required result.
\qed
\end{proof}
\begin{corollary}\label{th:dec2g}
Satisfiability checking w.r.t. GgPs is in \exptimex{2}.
\end{corollary}
\begin{proof}
Since satisfiability checking of \mugf{} formulas is \exptimex{2}-complete
(Theorem [1.1] in \cite{gradel}), satisfiability
checking w.r.t. GgPs is, by Corollary \ref{th:dec1g}, in \exptimex{2}.
\qed
\end{proof}
Thus, adding generalized literals to guarded programs does not come at the
cost of increased complexity of reasoning, as also for guarded
programs without generalized literals, reasoning is in \exptimex{2},
see Theorem \ref{th:dec}.
\par
In \cite{syrjanen04-jelia}, \emph{$\omega$-restricted}
programs allow for
\emph{cardinality constraints} and \emph{conditional literals}.
Conditional literals have the form $X.L:A$ where $X$ is a set of
variables, $A$ is an atom (the condition) and $L$ is an atom or a
naf-atom.  Intuitively, conditional literals correspond to generalized
literals $\Forall{X}{A\Rightarrow L}$, i.e., the defined reducts add
instantiations of $L$ to the body if the corresponding instantiation
of $A$ is true.  However, conditional literals appear only in
cardinality constraints
$\mathit{Card(b,S)}$\footnote{$\mathit{Card(b,S)}$ is true if at least
$b$ elements from $S$ are true.} where $S$ is a set of literals
(possibly conditional), such that a \emph{for all} effect such as with
generalized literals cannot be obtained with conditional literals.
\par
Take, for example, the rule $q\gets [\Forall{X}{b(X)\Rightarrow a(X)
}]$ and a universe $U=\set{x_1,x_2}$ with an interpretation containing
$b(x_1)$ and $b(x_2)$.  The reduct will contain a rule $q\gets
a(x_1),a(x_2)$ such that, effectively, $q$ holds only if
$a$ holds everywhere where $b$ holds.  The equivalent rule rewritten
with a conditional literal would be something like $q\gets
\mathit{Card}(n,\set{X.a(X):b(X)})$, resulting\footnote{Assume we
again have a universe $\set{x_1,x_2}$.} in a rule $q\gets \mathit{Card}(n,\set{a(x_1),a(x_2)})$. In
order to have the \emph{for all} effect, we have that $n$ must be $2$.
However, we cannot know this $n$ in advance, making it impossible to
express a \emph{for all} restriction.

\section{Relationship with Datalog LITE}\label{sec:related}

We define \emph{\dlite{}} as in \cite{dataloglite}. A \emph{Datalog
rule} is a rule \prule{\alpha}{\beta} where $\alpha = \set{a}$ for
some atom $a$ and $\beta$ does not contain generalized literals. 
A \emph{basic Datalog program} is a finite set of
Datalog rules such that no head predicate appears in negative bodies
of rules.  Predicates that appear only in the body of rules are
\emph{extensional} or \emph{input} predicates. Note that equality is,
by the definition of rules, never a head predicate and thus always
extensional.  The semantics of a basic Datalog program $P$, given a
relational input structure \rel{U} defined over extensional predicates
of $P$\footnote{
	We assume that an input structure always defines equality, and
	that
	it does so as the identity relation.
},
is given by 
 the unique (subset) minimal model of $\Sigma_P$ whose restriction to
 the extensional predicates yields $\rel{U}$ ($\Sigma_P$ are the
 first-order clauses corresponding to $P$, see \cite{abiteboul}).
\par
For a query $(P,q)$, where $P$ is a basic Datalog program and $q$ is
an
$n$-ary predicate, we write $\textbf{a} \in (P,q)(\rel{U})$ if the
minimal
model $M$ of $\Sigma_P$ with input \rel{U} contains $q(\textbf{a})$.
We call $(P,q)$ satisfiable if there exists a \rel{U} and an
$\textbf{a}$ such that  $\textbf{a} \in (P,q)(\rel{U})$. 
\par
A program $P$ is a \emph{stratified Datalog
program} if it can be
written as a union of basic Datalog programs $(P_1,\ldots, P_n)$,
so-called \emph{strata}, such that each of the head predicates in $P$
is a head predicate in exactly one stratum $P_i$.  Furthermore, if a
head predicate in $P_i$ is an extensional predicate in $P_j$, then $i
< j$.  This definition entails that head predicates in the positive
body of rules are head predicates in the same or a lower stratum, and
head predicates in the negative body are head predicates in a lower
stratum.  The semantics of stratified Datalog programs is defined
stratum per stratum, starting from the lowest stratum and defining the
extensional predicates on the way up. For an input structure $\rel{U}$ and a
stratified program $P=(P_1,\ldots, P_n)$, define as in
\cite{abiteboul}:
\[
\begin{array}{rl}
\rel{U}_0 &\equiv \rel{U} \\
\rel{U}_i &\equiv \rel{U}_{i-1} \cup P_i(\rel{U}_{i-1}\vert \edb{P_i})
\end{array}
\]
where $S_i \equiv P_i(\rel{U}_{i-1}\vert \edb{P_i})$ is the minimal model of $\Sigma_{P_i}$
among those models of $\Sigma_{P_i}$ whose restriction to the
extensional predicates of $P_i$ (i.e., $\edb{P_i}$) is equal to $\rel{U}_{i-1}\vert
\edb{P_i}$.  The \emph{least fixed point model} with input $\rel{U}$ of $P$ is per
definition $\rel{U}_n$.
\par
A \emph{\dlite{} generalized literal} is a generalized literal 
$\Forall{\textbf{Y}}{\impl{a}{b}}$ where $a$ and $b$ are atoms and $\vars{b} \subseteq
\vars{a}$.  
Note that \dlite{} generalized literals
$\Forall{\textbf{Y}}{\impl{a}{b}}$ can be replaced by the equivalent
$\Forall{\textbf{Z}}{\impl{a}{b}}$ where $\textbf{Z} \equiv
\setmin{\textbf{Y}}{\set{Y\mid Y\not\in \vars{a}}}$, i.e., with the
variables that are not present in the formula $a\Rightarrow b$ removed from
the quantifier. After such a rewriting, \dlite{} generalized literals
are guarded according to Definition
\ref{def:guard}.\label{remarkguarded}
\par
A \emph{\dlite{}} program is a stratified Datalog program,
possibly containing \dlite{} generalized literals in the positive body, where
each rule is \emph{monadic} or \emph{guarded}.  A rule is
monadic if
each of its (generalized) literals contains only one (free) variable;
it is guarded if there exists an atom in the positive body that
contains all variables (free variables in the case of generalized
literals) of the rule. The definition of stratified is adapted for
generalized literals:  for a $\Forall{\textbf{Y}}{\impl{a}{b}}$
in the body of a rule where the underlying predicate of $a$ is a head
predicate, this head predicate must be a head predicate in a lower
stratum (i.e., $a$ is treated as a naf-atom) and a head predicate
underlying $b$ must be in the same or a lower stratum (i.e., $b$ is
treated as an atom).  The semantics can be adapted accordingly since
$a$ is completely defined in a lower stratum,  as in
\cite{dataloglite}: every generalized literal
$\Forall{\textbf{Y}}{\impl{a}{b}}$ is instantiated (for any $\textbf{x}$
grounding the free variables $\textbf{X}$ in the generalized literal) by
$\bigwedge\set{b[\textbf{X}\mid \textbf{x}][\textbf{Y}\mid \textbf{y}] \mid
a[\textbf{X}\mid \textbf{x}][\textbf{Y}\mid \textbf{y}] \mbox{ is true}}$, which is
well-defined since $a$ is defined in a lower stratum than the rule
where the generalized literal appears.  
\subsection{Reduction from GgPs to Datalog LITE}

In \cite{dataloglite}, Theorem 8.5., a \dlite{}
query
$(\pi_{\varphi},q_{\varphi})$ was defined for an
alternation-free \mugf{} sentence
$\varphi$ such that
\[
(U,M) \models \varphi
\iff
(\pi_{\varphi},q_{\varphi})(M\cup\id{U})\mbox{ evaluates to true }\; ,
\]
where the latter means that $q_{\varphi}$ is in the fixed point model
of $\pi_{\varphi}$ with input $M\cup\id{U}$ and $\id{U} \equiv
\set{x=x\mid x \in U}$.
\begin{example}
Take the \mugf{} sentence $\GComp{P}\equiv \varphi_1 \land \varphi_2
\land\varphi_3 \land \varphi_4$ from Example \ref{ex:gcomp}, i.e., with
\[
\begin{split}
\varphi_1 &\equiv \Forall{X}{\impl{p(X)}{p(X)\lor\neg p(X)}} \\
\varphi_2 &\equiv \Forall{X}{\impl{r(X)}{p(X)}} \\
\varphi_3 &\equiv \Forall{X}{\impl{p(X)}{r(X)}} \\
\varphi_4 &\equiv \Forall{X}{\impl{p(X)}{\lfp{W}{X}{\phi(W,X)}(X)}} 
\end{split}
\]
and $\phi(W,X) \equiv W(X) \lor (W(X) \land r(X))$.  The 
query $(\pi_{\GComp{P}}, q_{\GComp{P}})$ considers atoms and negated
atoms as extensional predicates and introduces rules
\begin{pprogram}
\tsrule{H_{p,\varphi_1}(X)}{p(X)}
\tsrule{H_{\neg p,\varphi_1}(X)}{p(X), \neg p(X)}
\end{pprogram}
for $\varphi_1$ where both rules are guarded by the guard $p(X)$ of $\varphi_1$ (or, in general, the guard in the most closely encompassing
scope).
Disjunction
is defined as usual:
\begin{pprogram}
\tsrule{H_{p\lor\neg p, \varphi_1}(X)}{p(X), H_{p,\varphi_1}(X)}
\tsrule{H_{p\lor\neg p, \varphi_1}(X)}{p(X), H_{\neg p,\varphi_1}(X)}
\end{pprogram}
where $p(X)$ serves again as guard.\footnote{Actually, in this
particular case, the rules
would already be guarded without the guard of
$\varphi_1$, but we include it, as this is not true in general.
}
The sentence $\varphi_1$ itself is translated into
\begin{pprogram}
\tsrule{H_{\varphi_1}}{(\Forall{X}{\impl{p(X)}{H_{p\lor\neg p,\varphi_1}(X)}})}
\end{pprogram}
Formulas $\varphi_2$ and $\varphi_3$ can be translated similarly.
For $\varphi_4$, we translate, as an intermediate step, $\phi(W,X)$ as 
\begin{pprogram}
\tsrule{H_{\phi}(X)}{p(X),H_W(X)}
\tsrule{H_{\phi}(X)}{p(X),H_{W\land r}(X)}
\tsrule{H_{W\land r}(X)}{p(X),H_W(X),H_r(X)}
\tsrule{H_{W}(X)}{p(X),W(X)}
\tsrule{H_{r}(X)}{p(X),r(X)}
\end{pprogram}
from which the translation for $\lfp{W}{X}{\phi(W,X)}(X)$ can be
obtained by replacing $H_{\phi}(X)$ and $W(X)$ by
$H_{\lfp{W}{X}{\phi(W,X)}(X)}$, i.e.,
\begin{pprogram}
\tsrule{H_{\lfp{W}{X}{\phi(W,X)}}(X)}{p(X),H_W(X)}
\tsrule{H_{\lfp{W}{X}{\phi(W,X)}}(X)}{p(X),H_{W\land r}(X)}
\tsrule{H_{W\land r}(X)}{p(X),H_W(X),H_r(X)}
\tsrule{H_{W}(X)}{p(X),H_{\lfp{W}{X}{\phi(W,X)}}(X)}
\tsrule{H_{r}(X)}{p(X),r(X)}
\end{pprogram}
The sentence $\varphi_4$ is translated to 
\begin{pprogram}
\tsrule{H_{\varphi_4}}{(\Forall{X}{\impl{p(X)}{H_{\lfp{W}{X}{\phi(W,X)}}(X)}})}
\end{pprogram}
Finally, we compile the results in the rule
\prule{q_{\GComp{P}}}{H_{\varphi_1},H_{\varphi_2},H_{\varphi_3},H_{\varphi_4}}.
\par
In Example \ref{ex:gcomp}, we had, for a universe $\set{x}$,
the unique model $(\set{x},\emptyset)$ of $\GComp{P}$.  Accordingly,
we have that $\set{x = x}$ is the only relational input structure on
the extensional predicates of $\pi_{\GComp{P}}$, $r$ and $p$, that contains the term
$x$ and results in a
least fixed point model of $\pi_{\GComp{P}}$ containing $q_{\GComp{P}}$.
\end{example}
For the formal details of this reduction, we refer to
\cite{dataloglite}.  Satisfiability checking w.r.t. GgPs can be
polynomially reduced, using the above reduction, to satisfiability checking in \dlite.  
\begin{theorem}\label{th:dlite}
Let $P$ be a GgP, $q$ an $n$-ary predicate in $P$, and $\varphi$ the
\mugf{} sentence $
\Exists{\textbf{X}}{p(\textbf{X},\textbf{0},q) \land \bigwedge\GComp{\pprog{(\hbg{P})}{p}}}$.  $q$ is
satisfiable w.r.t. $P$ iff $(\pi_{\varphi},q_{\varphi})$ is
satisfiable.  Moreover, this reduction is polynomial.
\end{theorem}
\begin{proof}
By Theorem \ref{th:sat2g}, we have that $q$ is satisfiable w.r.t. $P$
iff $\varphi$ is satisfiable.
Since $\varphi$ is a \mugf{} sentence, we have that $\varphi$ is
satisfiable, i.e., there exists a $(U,M)$ such that $(U,M) \models \varphi$,
iff $(\pi_{\varphi},q_{\varphi})(M\cup\id{U})$ evaluates to true, i.e.,
$(\pi_{\varphi},q_{\varphi})$ is satisfiable.  
\par
Since, by Theorem \ref{th:sat2g}, the translation of $P$ to $\varphi$ is
polynomial in the size of $P$ and the query 
$(\pi_{\varphi},q_{\varphi})$ is polynomial in $\varphi$ \cite{dataloglite}, we
have a polynomial reduction.
\qed
\end{proof}

\subsection{Reduction from Datalog LITE to GgPs}

For stratified Datalog programs, possibly with generalized literals, least fixed point models with as input
the identity relation on a universe $U$ coincide with open answer sets
with universe $U$.
\begin{lemma}\label{lemma:pinS}
Let $P = (P_1, \ldots, P_n)$ be a stratified Datalog program, possibly with generalized
literals, and $\rel{U}$ an input structure for $P$. If $p(\textbf{x})
\in S_j$, then $p$ is a head predicate in $P_j$ or $p\in \rel{U}_{j-1}\vert
\edb{P_j}$.
\end{lemma}
\begin{proof}
Either $p\in \edb{P_j}$ or not.  In the former case, we have that
$p(\textbf{x})\in S_j\vert \edb{P_j}$ such that, by the definition of
$S_j$, $p(\textbf{x})\in \rel{U}_{j-1}\vert \edb{P_j}$.  In the latter case, we
have that, since $p$ does not appear in the body of $P_j$, but 
nevertheless $p(\textbf{x})$ is in $S_j$, a minimal model of $P_j$, 
$p$ must be a head predicate in $P_j$.
\qed
\end{proof}
\begin{lemma}\label{lemma:strat}
Let $P = (P_1, \ldots, P_n)$ be a stratified Datalog program, possibly with generalized
literals,  $\rel{U}$ an input structure for $P$. 
If $p$ is a head predicate in some $P_j$, $1\leq j \leq n$, then 
\begin{equation}\label{eq:lemma:strat}
p(\textbf{x}) \in S_j \iff p(\textbf{x})\in {\rel{U}}_n \; .
\end{equation}
If $p\in \edb{P_j}$ and $p(\textbf{x})\not\in \rel{U}_{j-1}$, then
$p(\textbf{x})\not\in {\rel{U}}_n$.
\end{lemma}
\begin{proof}
The ``only if" direction of Equation (\ref{eq:lemma:strat}) is immediate. For the ``if" direction:
assume $p$ is a head predicate in $P_j$ and $p(\textbf{x})\in
\rel{U}_n$. Since
$p(\textbf{x})\in \rel{U}_n$, there must be a $k$, such that
$p(\textbf{x})\in S_k$, $1\leq k \leq n$. 
\par
If $k=j$, we are finished, otherwise, by Lemma \ref{lemma:pinS},
$p(\textbf{x})\in \rel{U}_{k-1}\vert \edb{P_k}$ 
and thus $p(\textbf{x})\in \rel{U}_{k-1}$. Again, we have that there is a $1\leq k_1 \leq k-1$, such that $p(\textbf{x})\in S_{k_1}$.  If $k_1 = j$, we are finished,
otherwise, we continue as before.  After at most $n$ steps, we must
find a $k_n=j$, otherwise we have a contradiction ($p(\textbf{x})\in
\rel{U}$ is not possible since $p$ is a head predicate and input
structures are defined on extensional predicates only).
\par
Take $p$ extensional in $P_j$, $p(\textbf{x})\not\in
\rel{U}_{j-1}$, and $p(\textbf{x})\in \rel{U}_n$.  
We show that this leads to a contradiction.  From 
$p(\textbf{x})\in \rel{U}_n$, we have that $p(\textbf{x})\in \rel{U}_{n-1}$
or $p(\textbf{x})\in S_n$.  For the latter, one would have, with Lemma
\ref{lemma:pinS}, that $p(\textbf{x})\in \rel{U}_{n-1}\vert \edb{P_n}$ or
$p$ is a head predicate in $S_n$. The latter is impossible since $p\in
\edb{P_j}$ and $j\leq n$.  Thus, we have that $p(\textbf{x})\in \rel{U}_{n-1}$.
\par
Continuing this way, we eventually have that $p(\textbf{x})\in
\rel{U}_{j-1}$, a contradiction.
\qed
\end{proof}

\begin{theorem}\label{prop:strat}
Let $P = (P_1, \ldots, P_n)$ be a stratified Datalog program, possibly with
generalized literals,  $U$ a universe for $P$, and $l$ a literal.  For
the least fixed point model $\rel{U}_n$ of $P$ with input $\rel{U}=
\set{\id{U}}$, we have $\rel{U}_n\models l$ iff there exists an open
answer set $(U,M)$ of $P$ such that $M \models l$.
\par
Moreover, for any open answer set $(U,M)$ of $P$, we have that
$M=\setmin{\rel{U}_n}{\id{U}}$.
\end{theorem}
\begin{proof}
For the ``only if" direction, assume $\rel{U}_n\models l$.
 Define 
\[
M \equiv \setmin{\rel{U}_n}{id(U)}  \; .
\]
Clearly, $M\models l$, such that remains to show that $(U,M)$ is
an open answer set of $P$.

\begin{enumerate}

\item $M$ is a model of $R\equiv (P_U^{\fx(U,M)})^M$.

Take a rule $r:a[\textbf{X}\mid \textbf{x}] \gets \posi{\beta[\textbf{X}\mid
\textbf{x}]}, (\gli{\beta[\textbf{X}\mid \textbf{x}]})^{\fx(U,M)} \in R$, thus
$M \models \naf{\nega{\beta[]}}$, originating from $a\gets \beta \in
P$.  Assume $M \models \body{r}$.  We have that
$\Forall{\textbf{X}}{\bigwedge \beta \Rightarrow a}\in \Sigma_{P_i}$ for
some stratum $P_i$.  Take $\textbf{x}$ as in $r$.
\par
We verify that $\rel{U}_n \models \bigwedge \beta[]$.  We have that
$\rel{U}_n \models \bigwedge \posi{\beta[]} \land \bigwedge \neg
\nega{\beta[]}$. Take a generalized literal
$\Forall{\textbf{Y}}{c\Rightarrow b}$ in $\bigwedge \beta[]$ and
$\rel{U}_n \models c[\textbf{Y}\mid \textbf{y}]$.  Then $M \models c[]$ such
that $b[] \in (\gli{\beta[\textbf{X}\mid \textbf{x}]})^{\fx(U,M)}$, and
thus, with $M \models b[]$, that $\rel{U}_n \models b[]$.
\par
With Theorem 15.2.11 in \cite{abiteboul}, we have that $\rel{U}_n$
is a model of $\Sigma_P$, such that $a[]\in \rel{U}_n$, and
thus $M \models a[]$.

\item $M$ is a minimal model of $R\equiv (P_U^{\fx(U,M)})^M$.

Assume not, then there is a $N \subset M$, model of $R$.  Define $N' \equiv N
\cup \id{U}$.  Since $\setmin{M}{N}\neq \emptyset$, we have that
$\setmin{\rel{U}_n}{N'}\neq \emptyset$.  Since $\rel{U}=\id{U}$,
we have $\rel{U}_n = \id{U} \cup S_1 \cup  \ldots \cup
S_n$, such that there is a $1\leq j\leq n$, where
$\setmin{S_j}{N'}\neq \emptyset$ and $\rel{U}_{j-1}
\subseteq N'$.
Define $N_j\equiv 
\setmin{S_j}{(\setmin{S_j}{N'})}$.  One can show that 
$N_j \subset S_j$, $N_j\vert \edb{P_j} =
\rel{U}_{j-1}\vert\edb{P_j}$, and $N_j$ is a model of $\Sigma_{P_j}$, which is a
contradiction with the minimality of $S_j$.
\end{enumerate}

For the ``if" direction, assume $(U,M)$ is an open answer set of $P$
with $M\models l$. 
Assume $\rel{U}_n\not\models l$. Define $M' \equiv
\setmin{\rel{U}_n}{\id{U}}$.  By the previous direction, we know that
$(U,M')$ is an open answer set of $P$ with $M'\not\models l$, such
that $M\models l$ and $M'\not\models l$.
Note that $M\models p(\textbf{x}) \iff M'\models p(\textbf{x})$ for extensional predicates $p$ in $P$.
Indeed, assume $M\models p(\textbf{x})$, then $p(\textbf{x})$ must be in
the head of an applied rule since $M$ is an answer set, contradicting that $p$ is
extensional, unless $p$ is an equality, and then $\emptyset \models
p(\textbf{x})$ such that $M'\models p(\textbf{x})$. The other direction is
similar.
\par
One can show per induction on
$k$, that for a head predicate $p$ in $P_k$, $M\models p(\textbf{x})$ iff
$M'\models p(\textbf{x})$, resulting in $M=M'$, and thus in particular we have a
contradiction for $l$, such that $l \in \rel{U}_n$.

In particular, we have $M = M' = \setmin{\rel{U}_n}{\id{U}}$, which
proves the last part of the Theorem.
\qed
\end{proof}
From Theorem \ref{prop:strat}, we obtain a generalization of
Corollary 2 in \cite{gelfond88stable} (\emph{If $\Pi$ is
stratified, then its unique stable model is identical to its fixed
point model.}) for
stratified Datalog programs with generalized literals and an open answer set
semantics.
\begin{corollary}\label{cor:strat}
Let $P$ be a stratified Datalog program, possibly with
generalized literals,  and $U$ a universe for $P$. The
unique open answer set $(U,M)$ of $P$ is identical to its least fixed point
model (minus the equality atoms) with input structure $\id{U}$.
\end{corollary}
We generalize Theorem \ref{prop:strat}, to take into account
arbitrary input structures $\rel{U}$. For a stratified Datalog program
$P$, possibly with generalized literals, define $F_P\equiv 
\set{q(\textbf{X})\lor \naf{q(\textbf{X})}\gets \mid q \mbox{ extensional
(but not $=$) in } P}$.
\begin{theorem}\label{prop:strat2}
Let $P = (P_1, \ldots, P_n)$ be a stratified Datalog program, possibly with
generalized literals,  and $l$ a literal.  There exists an input
structure $\rel{U}$ for $P$ with least fixed point model $\rel{U}_n$
such that $\rel{U}_n\models l$
iff there exists an open
answer set $(U,M)$ of $P\cup F_P$ such that $M \models l$.
\end{theorem}
\begin{proof}
For the ``only if" direction, assume $\rel{U}_n\models l$. Define
${U}\equiv \cts{P\cup \rel{U}}$ and
\[
M \equiv \setmin{\rel{U}_n}{id(U)}  \; .
\]
Clearly, $M\models l$, and one can show, similarly to the proof of
Theorem \ref{prop:strat}, that $(U,M)$ is
an open answer set of $P$.
\par
For the ``if" direction, assume $(U,M)$ is an open answer set of
$P\cup F_P$
with $M\models l$. Define \[\rel{U}\equiv \id{U}\cup
\set{q(\textbf{x})\mid q(\textbf{x})\in M \land q \mbox{ extensional (but
not equality) in }P}\;.\]  Take $\rel{U}_n$ the least fixed point model
with input $\rel{U}$.
Assume $\rel{U}_n\not\models l$. Define $M' \equiv
\setmin{\rel{U}_n}{\id{U}}$.  By the previous direction, we know that
$(\cts{\rel{U}\cup P} (= U),M')$ is an open answer set of $P$ with $M'\not\models l$, such
that $M\models l$ and $M'\not\models l$. 
The rest of the proof is along the lines of the proof of Theorem
\ref{prop:strat}.
\qed
\end{proof}
The set of free rules $F_P$ ensures a free choice for extensional
predicates, a behavior that corresponds to the free choice of an input
structure for a Datalog program $P$.  Note that $P\cup F_P$ is not a
Datalog program anymore, due to the presence of naf in the heads of
$F_P$.
\par
Define a \dlitem{} program as a \dlite{} program where all rules are
guarded (instead of guarded or monadic). As we will see below this is
not a restriction.   As $F_P$ contains only free rules, $P\cup F_P$ is
a GgP if $P$ is a \dlitem{} program. Furthermore, the size of the GgP
$P\cup F_P$ is linear in the size of $P$.
\begin{theorem}\label{prop:fisgep}
Let $P$ be a \dlitem{} program. Then, $P\cup F_P$ is a GgP whose size
is linear in the size of $P$. 
\end{theorem}
\begin{proof}
Immediate by the Definition of \dlitem{} (note also the remark at pp.
\pageref{remarkguarded}) and the fact that $F_P$ is a set of free
rules and thus has no influence on the guardedness of $P$.
\qed
\end{proof}
Satisfiability checking of \dlitem{} queries can be reduced to
satisfiability checking w.r.t. GgPs.
\begin{theorem}\label{th:reductiontogep}
Let $(P,q)$ be a \dlitem{} query.  Then, $(P,q)$ is satisfiable iff
$q$ is satisfiable w.r.t. the GgP $P\cup F_P$.  Moreover, this reduction is
linear.
\end{theorem}
\begin{proof}
Immediate by Theorems \ref{prop:strat2} and \ref{prop:fisgep}.
\qed
\end{proof}
Theorems \ref{th:dlite} and \ref{th:reductiontogep} lead to the
conclusion that \dlitem{} and open ASP with GgPs are equivalent (i.e.,
satisfiability checking in either one of the formalisms can be
polynomially reduced to satisfiability checking in the
other).\footnote{
Note that $(\pi_\varphi,q_\varphi)$ is a \dlitem{} query
\cite{dataloglite}.
}
Furthermore, since \dlitem{}, \dlite{}, and alternation-free $\mugf$
are equivalent as well \cite{dataloglite}, we have the following
result.
\begin{theorem}\label{th:equivalence}
\dlite{}, alternation-free $\mugf$, and open ASP with GgPs are
equivalent.
\end{theorem}
Satisfiability checking in both GF and LGF is
\exptimex{2}-complete \cite{gradelrestraining}, as are their (alternation-free) extensions
with fixed point predicates \mugf{} and \mulgf{} \cite{gradel}.
Theorem \ref{th:equivalence} gives us then immediately the following
complexity result.
\begin{theorem}\label{th:ggps}
Satisfiability checking w.r.t. GgPs is \exptimex{2}-complete.
\end{theorem}
Some extra terminology is needed to show that satisfiability checking
w.r.t. (L)GPs (i.e.,
without generalized literals) is \exptimex{2}-complete as well.
\par
\emph{Recursion-free} stratified Datalog is stratified Datalog where the head
predicates in the positive bodies of rules must be head predicates in a
lower stratum.  We call recursion-free \dlitem{},
\dliter, where the definition of recursion-free is appropriately extended
to take into account the generalized literals.   
\par
For a \dliter{} program $P$, let $\double{P}$ be the program $P$ with
all generalized  literals replaced by a double negation. E.g.,
\[\prule{q(X)}{f(X),\Forall{Y}{\impl{r(X,Y)}{s(Y)}}}\] is rewritten as the
rules \[\prule{q(X)}{f(X),\naf{q'(X)}}\] and
\[\prule{q'(X)}{r(X,Y),\naf{s(Y)}}\;.\] As indicated in \cite{dataloglite},
this yields an equivalent program $\double{P}$, where the
recursion-freeness ensures that \double{P} is stratified.\footnote{
Note that this translation cannot work for arbitrary generalized programs as
the antecedent of generalized literals can be an arbitrary boolean formula,
which cannot appear in bodies of rules. Replace, e.g., $r(X,Y)$ by $r(X,Y) \lor
d(X,Y)$.
}
\begin{theorem}\label{prop:double}
Let $P$ be a \dliter{} program.  Then $\double{P}\cup F_{\double{P}}$ is a GP.
\end{theorem}
\begin{proof}
Every rule in $P$ is guarded, and thus every rule in \double{P} is too.
Since $\double{P}\cup F_{\double{P}}$ adds but free rules to \double{P}, 
all non-free rules of $\double{P}\cup F_{\double{P}}$ are guarded.
\qed
\end{proof}
Satisfiability checking of
\dliter{} queries can be linearly reduced to satisfiability checking
w.r.t. GPs.
\begin{theorem}\label{th:opendouble}
Let $(P,q)$ be a \dliter{} query.  $(P,q)$ is satisfiable iff $q$ is
satisfiable w.r.t. the GP $\double{P}\cup F_{\double{P}}$. Moreover, this reduction is
linear.
\end{theorem}
\begin{proof}
For a \dliter{} query $(P,q)$, 
$(\double{P},q)$ is an equivalent stratified Datalog query. Hence, by
Theorem \ref{prop:strat2}, $(\double{P},q)$ is satisfiable iff $q$
is satisfiable w.r.t. $\double{P}\cup F_{\double{P}}$.
This reduction is linear since \double{P} is linear in the size of $P$
and so is $\double{P}\cup F_{\double{P}}$. 
\qed
\end{proof}
\begin{theorem}\label{th:gpcomplete}
Satisfiability checking w.r.t. (L)GPs is \exptimex{2}-complete.
\end{theorem}
\begin{proof}
The reduction from alternation-free \mugf{} sentences $\varphi$ to \dlite{}
queries $(\pi_{\varphi},q_{\varphi})$ specializes, as noted in
\cite{dataloglite}, to a reduction from GF sentences to recursion-free
\dlite{} queries. Moreover, the reduction contains only guarded rules such
that GF sentences $\varphi$ are actually translated to \dliter{} queries
$(\pi_{\varphi},q_{\varphi})$.
\par
Satisfiability checking in the guarded fragment GF is
\exptimex{2}-complete \cite{gradelrestraining}, such that, using
Theorem \ref{th:opendouble} and the intermediate \dliter{}
translation, we have that satisfiability checking w.r.t. GPs is
\exptimex{2}-hard.  The \exptimex{2} membership was shown in Theorem
\ref{th:dec}, such that the completeness readily follows.
\par
Every GP is a LGP and satisfiability checking w.r.t. to the
former is \exptimex{2}-complete, thus we have \exptimex{2}-hardness for
satisfiability checking w.r.t. LGPs.  Completeness follows again from
Theorem \ref{th:dec}.
\qed
\end{proof}

\section{CTL Reasoning using
Guarded Generalized Programs}\label{sec:ctloasp}
In this section, we show how to reduce CTL satisfiability checking
to
satisfiability checking w.r.t.  GgPs, i.e., guarded programs with generalized
literals, thus arguing the usability of OASP as a suitable formalism for different kinds of knowledge representation.
\par
In order to keep the treatment simple, we will assume that the only
allowed temporal constructs are \all{\eventually{q}}, \some{(\until{p}{q})},
and \some{{\Next{q}}}, for formulas $p$ and $q$. They are actually adequate in the sense that
other temporal constructs can be equivalently, i.e., preserving
satisfiability, rewritten using only those three \cite{huth}.   
\par
For a CTL formula $p$, let $\clos{p}$ be the \emph{closure} of $p$:
the set of subformulas of $p$.
We construct a GgP $G\cup D_p$ consisting of a generating part $G$ and a defining
part $D_p$.
The guarded program $G$ contains free rules (\ref{g:1}) for every proposition $P \in AP$,
free rules (\ref{g:2}) that allow for state transitions, and rules
(\ref{g:3}) that ensure that the transition relation is total:
\begin{xalignat}{2}
\strule{\pr{P}(S)\lor\naf{\pr{P}(S)}}{} \label{g:1}\tag{$g_1$} \\
\strule{next(S, N)\lor \naf{next(S, N)}}{}   \label{g:2} \tag{$g_2$}  \\
\strule{succ(S)}{next(S, N)} & \strule{}{S = S, \naf{succ(S)}} \label{g:3}\tag{$g_3$} 
\end{xalignat}
where $\pr{P}$ is the predicate corresponding to the proposition $P$.
The $S=S$ is necessary merely for having guarded rules; note that any
rule containing only one (free) variable can be made guarded by adding such
an equality.
\par
The GgP $D_p$ introduces for every non-propositional CTL formula in \clos{p} the
following rules (we write \pr{q} for the predicate corresponding to
the CTL formula $q \in \clos{p}$); as noted before we tacitly assume
that rules containing only one (free) variable $S$ are guarded by
$S=S$:
\begin{itemize}
\item For a formula $\neg q$ in $\clos{p}$, we introduce in $D_p$ the rule
\begin{align}
\strule{\pr{\neg q}(S)}{\naf{\pr{q}(S)}} \label{d:1}\tag{$d_1$}
\end{align}
Thus, the negation of a CTL formula is simulated by negation as
failure.
\item For a formula $q\land r$ in $\clos{p}$, we introduce in $D_p$ the rule
\begin{align}
\strule{\pr{q\land r}(S)}{\pr{q}(S), \pr{r}(S)} \label{d:2}\tag{$d_2$}
\end{align}
Conjunction of CTL formulas thus corresponds to conjunction in the
body.
\item For a formula \all{\eventually{q}} in $\clos{p}$, we introduce in $D_p$ the rules
\begin{align}
\strule{\pr{\all{\eventually{q}}}(S)}{{\pr{q}(S)}} \label{d:32}\tag{$d_3^1$}\\
\strule{\pr{\all{\eventually{q}}}(S)}{\Forall{N}{next(S, N)\Rightarrow \pr{\all{\eventually{q}}}(N)}} \label{d:33}\tag{$d_3^2$}
\end{align}
We define $\all{\eventually{q}}$ corresponding to
the intuition that $\all{\eventually{q}}$ holds if, either $q$ holds at the current state $(d_3^1)$ or for all successors, we
have that $\all{\eventually{q}}$ holds $(d_3^2)$.   Note that
we use generalized literals to express the \emph{for all successors}
part.  Moreover, we explicitly use the minimal model semantics of
(open) answer set programming to ensure that eventually $\pr{q}$ holds
on all paths: one cannot continue to use rule $(d_3^2)$ to motivate
satisfaction of \all{\eventually{q}}, at a certain finite point,
one is obliged to use rule $(d_3^1)$ to obtain a finite motivation.

\item For a formula \some{(\until{q}{r})} in $\clos{p}$, we introduce in $D_p$ the rules
\begin{align}
\strule{\pr{\some{(\until{q}{r}})}(S)}{\pr{r}(S)} \label{d:4}\tag{$d_4$}\\
\strule{\pr{\some{(\until{q}{r})}}(S)}{\pr{q}(S), next(S, N), \pr{\some{(\until{q}{r})}}(N)} \label{d:5}\tag{$d_5$}
\end{align}
based on the intuition that there is a path where $q$ holds until $r$
holds (and $r$ eventually holds) if either $r$ holds at the current
state $(d_4)$, or $q$ holds at the current state and there is some
next state where again \some{(\until{q}{r})} holds $(d_5)$.  The minimality
will again make sure that we eventually must deduce $r$ with rule
$(d_4)$.

\item For a formula \some{\Next{q}} in $\clos{p}$, we introduce in $D_p$ the rule
\begin{align}
\strule{\pr{\some{\Next{q}}}(S)}{next(S,N), \pr{q}(N)} \label{d:6}\tag{$d_6$}
\end{align}
saying that \some{\Next{q}} holds if there is some successor where $q$ holds.
\end{itemize}
Note that replacing the generalized literal in $(d_3^2)$ with a double
negation has not the intended effect:
\begin{pprogram}
\tsrule{\pr{\all{\eventually{q}}}(S)}{\naf{q'(S)}}
\tsrule{q'(S)}{next(S, N), \naf{\pr{\all{\eventually{q}}}(N)}}
\end{pprogram}
A (fragment) of an open answer set could then be 
\begin{multline*}
(\set{s_0,s_1,\ldots
},\{\lit{next}(s_0,s_1),\lit{next}(s_1,s_2),\ldots,
\\
\pr{\all{\eventually{q}}}(s_0),
\pr{\all{\eventually{q}}}(s_1),\ldots\})\; ,
\end{multline*}
such that one would
conclude that \pr{\all{\eventually{q}}} is satisfiable while
there is a path $s_0,s_1,\ldots$ where $q$ never holds.
\begin{example}
Consider the absence of starvation formula $t
\Rightarrow \all{\eventually{c}}$, i.e., if a process tries (\emph{t}) to
access a critical section of code, it must eventually succeed in doing so (\emph{c}). We rewrite this such that it does
not contain $\Rightarrow$, i.e., we consider the equivalent formula
$\neg (t\land \neg \all{\eventually{c}})$. For $AP=\set{c,t}$, the
program $G$ contains the rules \begin{pprogram}
\tsrule{\pr{t}(S)\lor\naf{\pr{t}(S)}}{}
\tsrule{\pr{c}(S)\lor\naf{\pr{c}(S)}}{}
\tsrule{next(S, N)\lor \naf{next(S, N)}}{} 
\tsrule{succ(S)}{next(S, N)} 
\tsrule{}{S = S, \naf{succ(S)}} 
\end{pprogram}
The program $D_p$, with $p \equiv \neg (t\land \neg \all{\eventually{c}})$, contains the rules
\begin{pprogram}
\tsrule{\pr{\neg (t\land \neg \all{\eventually{c}})}(S)}{\naf{\pr{t\land \neg \all{\eventually{c}}}(S)}}
\tsrule{\pr{t\land \neg \all{\eventually{c}}}(S)}{{\pr{t}(S),\pr{\neg \all{\eventually{c}}}}(S)}
\tsrule{\pr{\neg \all{\eventually{c}}}(S)}{\naf{\pr{\all{\eventually{c}}}(S)}}
\tsrule{\pr{\all{\eventually{c}}}(S)}{\pr{c}(S)}
\tsrule{\pr{\all{\eventually{c}}}(S)}{\Forall{N}{next(S, N)\Rightarrow \pr{\all{\eventually{c}}}(N)}}
\end{pprogram}
One can see that $p$ is (CTL) satisfiable iff $\pr{p}$ is satisfiable w.r.t.
$G\cup D_p$.
\end{example}
\begin{theorem}\label{th:modeliffoaset}
Let $p$ be a CTL formula. $p$ is satisfiable iff \pr{p} is
satisfiable w.r.t. the GgP $G \cup D_p$.
\end{theorem}
\begin{proof}
For the ``only if" direction, assume $p$ is satisfiable. Then there
exists a model $K = (S,R,L)$ of $p$ such that $K,s\models p$, for a
state $s\in S$.
Define 
\[
\begin{split}
M &\equiv \set{\mathit{next}(s,t) \mid (s,t) \in R} \cup \set{\lit{succ(s)}
\mid (s,t)\in R}\\
  &\cup \set{\pr{q}(s) \mid K,s \models q \land q \in \clos{p}} \; .
\end{split}
\]

Then $\pr{p}(s) \in M$; one can show that $(S,M)$ is an open answer set of
$G\cup D_p$.
\par
For the ``if" direction, assume $(U,M)$ is an open answer set of $G
\cup D_p$ such that $\pr{p}(s) \in M$ for some $s$, where $p$ is a CTL
formula.  Define the model
$K = (U,R,L)$ with $R = \set{(s,t) \mid \mathit{next}(s,t) \in M}$,
and $L(s) = \set{P\mid \pr{P}(s) \in M \land P \in AP}$. Remains to
show that $K$ is a structure and $K,s \models p$.
\par
The relation $R$ is total, indeed, assume not, then there is a $t\in
U$, which has no successors in $R$.  Then, there is no
$\lit{next(t,t')}\in M$, such that $\lit{succ(t)}\not\in M$, and the
constraint ($g_3$) gives a contradiction.
One can prove per induction on the structure of a CTL formula
$q$, that
\[
K,s \models q \iff \pr{q}(s)\in M \enspace .
\]
\qed
\end{proof}

Since CTL satisfiability checking is \exptime-complete (see Theorem
\ref{th:ctl}, pp. \pageref{th:ctl}) and satisfiability checking w.r.t.
GgPs is \exptimex{2}-complete (see Theorem \ref{th:ggps}, pp.
\pageref{th:ggps}), the reduction from CTL to GgPs does not seem to be
optimal.  However, we can show that the particular GgP $G\cup D_p$ is
a \emph{bound} GgP for which
reasoning is indeed \exptime{}-complete and thus optimal.
\par
The \emph{width} of a formula $\psi$ is the maximal
number of free variables in its subformulas \cite{gradel3}.  We define
\emph{bound} programs by looking at their first-order form and the
arity of its predicates.
\begin{definition}\label{def:bound}
Let $P$ be a gP.  Then, $P$ is \emph{bound} if every formula in 
$\sat{P}$ is of bounded width and the predicates in $P$ have a bounded
arity.
\end{definition}
For a CTL formula $p$, one has that $G\cup D_p$ is a bound GgP.
\begin{theorem}\label{th:bounded}
Let $p$ be a CTL formula.  Then, $G\cup D_p$ is a bound GgP.
\end{theorem}
\begin{proof}
Every subformula of formulas in $\sat{G\cup D_p}$ contains at most $2$
free
variables and the maximum arity of the predicates is $2$ as well.\qed
\end{proof}
\begin{theorem}\label{th:bound}
Satisfiability checking w.r.t. bound GgPs is
\exptime{}-complete.
\end{theorem}
\begin{proof}
Let $P$ be a bound GgP.  We have that $\pprog{(\hbg{P})}{p}$ is bound and one can check that
$\Exists{\textbf{X}}{p(\textbf{X},\textbf{0},q) \land
\bigwedge \GCompg{\pprog{(\hbg{P})}{p}}}$ is of bounded width.  Note
that formula (\ref{eq:fpfg}) on pp. \pageref{eq:fpfg} contains a
$p(\textbf{X})$.  The condition that each formula in $\sat{P}$ is of
bounded width is not enough to guarantee that $p(\textbf{X})$ has bounded
width. Add, e.g., ground rules $r$ to $P$ with increasing arities of
predicates.  Although the width of formulas in $\sat{P}$ remains
constant (no variables are added), the arity of 
$p(\textbf{X})$ in Formula (\ref{eq:fpfg}) increases,  thus increasing the width.  Hence,
the restriction that the arity of predicates in $P$ should be bounded as
well.
\par
By Theorem \ref{th:sat2g} and \ref{th:dec1g}, one can reduce satisfiability checking of a bound
GgP to satisfiability of a $\mugf$-formula with
bounded width.  The latter can be done in \exptime{} by Theorem 1.2 in
\cite{gradel}, such that satisfiability checking w.r.t. bound GgPs is in
\exptime{}.
\par
The \exptime{}-hardness follows from Theorem \ref{th:modeliffoaset}
and the \exptime{}-hardness of CTL satisfiability checking (Theorem
\ref{th:ctl}).
\qed
\end{proof}

As indicated in \cite{dataloglite}, the objects in the database form the states
of the Kripke model.  In the open domain case, one, intuitively, allows, of
extra states in the Kripke model, not explicitly listed in the database.

\section{Conclusions and Directions for Further Research}\label{sec:conclusions}

We embedded OASP in FPL and used this embedding to identify (loosely)
guarded OASP, a decidable fragment of OASP.  Finite ASP was reduced to
loosely guarded OASP.  Satisfiability checking w.r.t. (loosely) guarded
OASP was shown to be \exptimex{2}-complete.
We defined GgPs, guarded programs with generalized
literals, under an open answer set semantics, and showed
\exptimex{2}-completeness of satisfiability checking by a reduction to
$\mugf$.  Furthermore, we translated \dlitem{} programs to GgPs, and
generalized the result that the unique answer set of a stratified
program is identical to its least fixed point. We showed how to
optimally simulate CTL in OASP.
\par
We plan to extend GgPs to loosely guarded gPs, where a guard may be a
set of atoms; a reduction to the loosely guarded fixed point logic
should then provide for decidability.
More liberal generalized literals, with
the consequent a conjunction of atoms and naf-atoms instead
of just an atom, does not affect the definition of the GeLi-reduct, but 
the FPL translation requires modification to ensure no fixed point
variable appears negatively.
\par
We plan to look into the correspondence with Datalog and use
decidability results for Datalog satisfiability checking, as, e.g., in
\cite{halevy}, to search for decidable fragments under an open answer set
semantics.
\par
Although adding generalized literals to guarded programs does not
increase the complexity of reasoning, it does seem to increase
expressivity: one can, for example, express infinity axioms.  Given
the close relation with \dlite{} and the fact that \dlite{} without
generalized literals cannot express well-founded statements, it seems
unlikely that guarded programs without generalized literals can
express infinity axioms; this is subject to further research.
\par
We only considered generalized literals in the positive body.  If the
antecedents in generalized literals are atoms, it seems intuitive to
allow also generalized literals in the negative body.  E.g., take a
rule $\alpha\gets \beta,\naf{[\Forall{X}{b(X)\Rightarrow a(X)}}]$; it
seems natural to treat $\naf{[\Forall{X}{b(X)\Rightarrow a(X)}]}$ as
$\Exists{X}{b(X)\land \neg a(X)}$ such that the rule becomes 
$\alpha\gets \beta,b(X),\naf{a(X)}$.  A rule like ${[\Forall{X}{b(X)\Rightarrow a(X)}}]\lor\alpha\gets
\beta$ is more involved and it seems
that the generalized literal can only be intuitively removed by a
modified GeLi-reduct.
\par
We established the equivalence of open ASP with GgPs, alternation-free
$\mugf{}$, and \dlite{}.  Intuitively, \dlite{} is not expressive
enough to simulate normal $\mugf{}$ since such $\mugf{}$ formulas
could
contain negated fixed point variables, which would result in a
non-stratified program when translating to \dlite{} \cite{dataloglite}.  Open ASP with GgPs
does not seem to be sufficiently expressive either: fixed point
predicates would need to appear under negation as failure, however,
the GL-reduct removes naf-literals, such that, intuitively, there is no real
recursion through naf-literals.  Note that it is unlikely (but still
open) whether alternation-free $\mugf{}$ and normal $\mugf{}$ are
equivalent, i.e., whether the alternation hierarchy can always be
collapsed.
\par
In \cite{dataloglite} one also discusses the data complexity which is simpler
than the combined complexity as studied here.  For the future, we also
investigate the data complexity of reasoning in guarded Open Answer Set
Programming.

\bibliographystyle{acmtrans}
\bibliography{journal_goasp}

\begin{received}
    Received March 2006;
    accepted September 2006
    \end{received}
\end{document}